\pdfoutput=1

\documentclass[11pt]{article}

\usepackage[preprint]{acl}

\usepackage{times}
\usepackage{latexsym}
\usepackage{xspace}
\usepackage{titlesec}
\usepackage{multirow}
\usepackage{graphicx} 
\usepackage{array}
\usepackage{adjustbox}
\usepackage{natbib}
\usepackage[T1]{fontenc}

\usepackage[utf8]{inputenc}

\usepackage[stretch=25, shrink=45, expansion=true, spacing=true, tracking=true, protrusion=true, factor=1100, final=true
]{microtype}
\microtypecontext{spacing=nonfrench}

\usepackage{inconsolata}
\usepackage{amsmath}
\usepackage{booktabs}
\usepackage{multirow}
\usepackage{graphicx}
\usepackage{caption}
\usepackage{subcaption}
\usepackage{amsfonts}
\usepackage{algorithm}
\usepackage{algorithmic}
\usepackage{amssymb}
\usepackage{pifont}
\usepackage{xcolor}
\usepackage{tabularx}
\usepackage{bbding}
\usepackage{utfsym}
\usepackage{hyperref}
\usepackage{enumitem}
\usepackage{setspace}
\usepackage{courier} 
\usepackage{url}            
\usepackage{nicefrac}       
\usepackage[edges]{forest}

\definecolor{hidden-red}{RGB}{205, 44, 36}
\definecolor{hidden-blue}{RGB}{194,232,247}
\definecolor{hidden-orange}{RGB}{243,202,120}
\definecolor{hidden-green}{RGB}{34,139,34}
\definecolor{hidden-pink}{RGB}{255,245,247}
\definecolor{hidden-black}{RGB}{20,68,106}

\pdfoutput=1

\usepackage[utf8]{inputenc}

\usepackage{inconsolata}

\usepackage{graphicx}
\usepackage{booktabs} 
\title{\includegraphics[width=0.5cm]{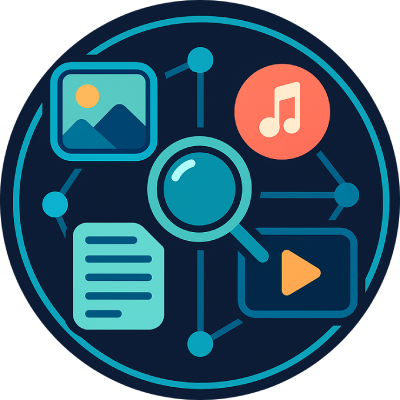} Ask in Any Modality\\ A Comprehensive Survey on Multimodal Retrieval-Augmented Generation}

\author{
\small
  \textbf{Mohammad Mahdi Abootorabi\textsuperscript{\textdagger}},
  \textbf{Amirhosein Zobeiri\textsuperscript{$\diamond$}},
  \textbf{Mahdi Dehghani\textsuperscript{\textparagraph}},
  \textbf{Mohammadali Mohammadkhani\textsuperscript{\textsection}},
\\
  \small
  \textbf{Bardia Mohammadi\textsuperscript{\textsection}},
  \textbf{Omid Ghahroodi\textsuperscript{\textdagger}},
  \textbf{Mahdieh Soleymani Baghshah\textsuperscript{\textsection, \textasteriskcentered}},
  \textbf{Ehsaneddin Asgari\textsuperscript{\textdagger, \textasteriskcentered}}
  \\[0.5em]
  \small
  \textsuperscript{\textsection}Computer Engineering Department, Sharif University of Technology, Tehran, Iran,\\
  \small
  \textsuperscript{$\diamond$}College of Interdisciplinary Science and Technology, University of Tehran, Tehran, Iran,\\
  \small
  \textsuperscript{\textparagraph}Computer Engineering Department, K.N. Toosi University of Technology, Tehran, Iran,\\
  \small
  \textsuperscript{\textdagger}Qatar Computing Research Institute, Doha, Qatar
\\
  \small{
    \textbf{Correspondence:} \href{mailto:soleymani@sharif.edu}{soleymani@sharif.edu} and  
    \href{mailto:easgari@hbku.edu.qa}{easgari@hbku.edu.qa}
  }
}

\titlespacing*{\paragraph}{0pt}{0.3ex plus 0.1ex minus 0.1ex}{0.5em}
\titleformat{\paragraph}[runin]{\normalfont\normalsize\bfseries}{\theparagraph}{0.2em}{}[]

\titlespacing{\section}{0pt}{1.0ex}{1.0ex}
\titlespacing{\subsection}{0pt}{1.0ex}{1.0ex}

\begin{document}
\maketitle
\begin{abstract}
Large Language Models (LLMs) suffer from hallucinations and outdated knowledge due to their reliance on static training data. Retrieval-Augmented Generation (RAG) mitigates these issues by integrating external dynamic information for improved factual grounding. With advances in multimodal learning, Multimodal RAG extends this approach by incorporating multiple modalities such as text, images, audio, and video to enhance the generated outputs. However, cross-modal alignment and reasoning introduce unique challenges beyond those in unimodal RAG. This survey offers a structured and comprehensive analysis of Multimodal RAG systems, covering datasets, benchmarks, metrics, evaluation, methodologies, and innovations in retrieval, fusion, augmentation, and generation. We review training strategies, robustness enhancements, loss functions, and agent-based approaches, while also exploring the diverse Multimodal RAG scenarios. In addition, we outline open challenges and future directions to guide research in this evolving field. This survey lays the foundation for developing more capable and reliable AI systems that effectively leverage multimodal dynamic external knowledge bases. All resources are publicly available~\footnote{\url{https://github.com/llm-lab-org/Multimodal-RAG-Survey}}.
\end{abstract}

\newcommand{\defaultfootnote}{\thefootnote}
\renewcommand{\thefootnote}{\textasteriskcentered}
\footnotetext{These authors contributed equally.}
\renewcommand{\thefootnote}{\defaultfootnote}

\section{Introduction \& Background}

\noindent
Recent advancements in transformer architectures \cite{10.5555/3295222.3295349}, coupled with increased computational resources and the availability of large-scale training datasets \cite{naveed2024comprehensiveoverviewlargelanguage}, have significantly accelerated progress in the development of language models.
The emergence of foundational Large Language Models (LLMs) \cite{NEURIPS2022_b1efde53, grattafiori2024llama3herdmodels, touvron2023llama2openfoundation, qwen2025qwen25technicalreport, anil2023palm2technicalreport}, has revolutionized natural language processing (NLP), excelling in tasks such as instruction following \cite{qin-etal-2024-infobench}, reasoning \cite{10.5555/3600270.3602070}, in-context learning \cite{NEURIPS2020_1457c0d6}, and multilingual translation \cite{zhu-etal-2024-multilingual}. 
Despite these achievements, LLMs face challenges such as hallucinations, outdated knowledge, and a lack of verifiable reasoning \cite{10.1145/3703155, xu2024hallucinationinevitableinnatelimitation}. Their reliance on parametric memory limits access to up-to-date information, reducing their effectiveness in knowledge-intensive tasks. 
\paragraph{Retrieval-Augmented Generation (RAG)}
RAG \cite{lewis2020retrieval} addresses these limitations by enabling LLMs to retrieve and incorporate external knowledge, improving factual accuracy and reducing hallucinations \cite{shuster-etal-2021-retrieval-augmentation, ding2024retrieveneedsadaptiveretrieval}. By dynamically accessing external knowledge sources, RAG enhances knowledge-intensive tasks while grounding responses in verifiable sources \cite{Gao2023RetrievalAugmentedGF}.
In practice, RAG systems follow a retriever-generator pipeline: the retriever uses embedding models 
\cite{chen-etal-2024-m3, rau2024contextembeddingsefficientanswer} to identify relevant passages from external knowledge bases and may apply re-ranking techniques to improve precision \cite{dong2024dontforgetconnectimproving}. The retrieved passages are then provided to the generator, which leverages this contextual information to produce more informed and coherent responses. Recent advancements in RAG frameworks, such as planning-guided retrieval \cite{lee-etal-2024-planrag}, agentic RAG \cite{an2024goldenretrieverhighfidelityagenticretrieval}, and feedback-driven iterative refinement \cite{liu-etal-2024-ra, asai2023selfrag}, have further improved both the retrieval and generation components of these systems.

\begin{figure*}[t!]
\centering
  \includegraphics[width=0.9\textwidth]{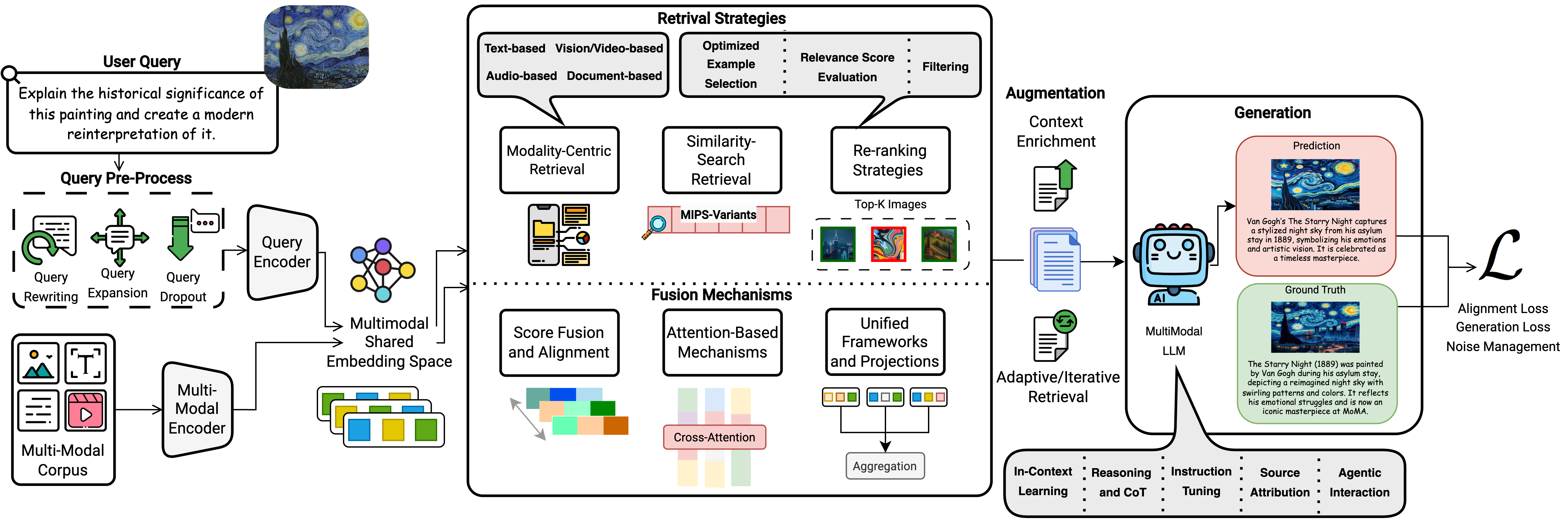}
\caption{\label{fig:overview}
Overview of the multimodal RAG pipeline, illustrating key techniques and recent advancements.
\vspace{-5mm}
}
\end{figure*}
 
\paragraph{Multimodal Learning}
In parallel with advances in language modeling, multimodal learning has emerged as a transformative area in artificial intelligence, enabling systems to integrate and reason over heterogeneous data sources for more comprehensive representations. A pivotal breakthrough was the introduction of CLIP model \cite{radford2021learning}, which aligned visual and textual modalities through contrastive learning and inspired a wave of subsequent models \cite{10.5555/3600270.3601993, wang2023one, pramanick2023volta}. These developments have catalyzed progress across diverse domains, including sentiment analysis \cite{10.1145/3586075} and biomedical applications \cite{hemker2024healnet}, highlighting the effectiveness of multimodal approaches. By facilitating the joint processing of text, images, audio, and video, multimodal learning is increasingly recognized as a critical enabler of progress toward artificial general intelligence (AGI) \cite{song2025bridge}.

\paragraph{Multimodal RAG}
The extension of large language models to multimodal LLMs (MLLMs) has significantly broadened their capabilities, enabling reasoning and generation across multiple data modalities \cite{liu2023llava, geminiteam2024geminifamilyhighlycapable, 10.5555/3618408.3619222}. Notably, GPT-4 \cite{openai2024gpt4technicalreport} demonstrates human-level performance by jointly processing text and images, marking a milestone in multimodal understanding. Building on this progress, multimodal RAG incorporates diverse sources, such as images, audio, and structured data, to enrich contextual grounding and enhance generation quality \cite{Hu_2023_CVPR, chen-etal-2022-murag}. This approach improves reasoning by leveraging cross-modal cues, but also introduces challenges, including modality selection, effective fusion, and managing cross-modal relevance \cite{zhao-etal-2023-retrieving}.
\autoref{fig:overview} illustrates the general pipeline of these systems.

\paragraph{Multimodal RAG Formulation}  
We present a mathematical formulation of multimodal RAG. These systems aim to generate a multimodal response \( r \) given a multimodal query \( q \). Let \( D = \{d_1, d_2, \ldots, d_n\} \) denote a multimodal corpus. For clarity, we assume each document \( d_i \in D \) is associated with a single modality \( M_{d_i} \). In practice, however, documents often span multiple modalities—for example, a scientific article containing both text and images. Such cases are typically addressed by either decomposing the document into modality-specific sub-documents or employing universal encoders capable of jointly processing multiple modalities.

Each document \( d_i \) is encoded using its corresponding modality-specific encoder, yielding \( z_i = \text{Enc}_{M_{d_i}}(d_i) \). The collection of all encoded representations is denoted as \( Z = \{z_1, z_2, \ldots, z_n\} \). These modality-specific encoders project diverse input modalities into a shared semantic space, enabling cross-modal alignment.

A retrieval model \( R \) computes a relevance score \( s(e_q, z_i) \) between the encoded query representation \( e_q \) (obtained by encoding \( q \) using the appropriate encoders) and each document representation \( z_i \). The retrieval-augmented multimodal context \( X \) is constructed by selecting documents whose relevance scores exceed a modality-specific threshold:
\vspace{-0.3cm}
\[
X = \{ d_i \in D \mid s(e_q, z_i) \geq \tau_{M_{d_i}} \},\vspace{-0.1cm}
\] where \( \tau_{M_{d_i}} \) is the relevance threshold for the modality \( M_{d_i} \), and \( s \) is the scoring function that measures semantic relevance. Finally, the generative model \( G \) produces the response conditioned on the original query \( q \) and the retrieved context \( X \), formally defined as
$r = G(q, X)$.


\paragraph{Related Works}  
Multimodal RAG is a rapidly emerging field, yet a comprehensive survey dedicated to its recent advancements remains lacking. While over ten surveys discuss RAG-related topics such as Agentic RAG \cite{singh2025agenticretrievalaugmentedgenerationsurvey}, none specifically focus on the multimodal setting. To our knowledge, the only relevant work \cite{zhao-etal-2023-retrieving} categorizes multimodal RAGs by application and modality. In contrast, our survey adopts a more innovation-driven perspective, offering a detailed taxonomy and addressing recent trends and open challenges. We review over 100 recent papers, primarily from the ACL Anthology, reflecting the growing interest and progress in this domain.\\
\vspace{-0.35cm}


\paragraph{Contributions}  
In this work, \textbf{(i)} we present a comprehensive review of multimodal RAG, covering task formulation, datasets, benchmarks, applications, and key innovations across retrieval, fusion, augmentation, generation, training strategies, loss functions, and agent frameworks. \textbf{(ii)} We propose a structured taxonomy (\autoref{fig:taxonomy_full}) that categorizes state-of-the-art models by their core contributions, highlighting methodological advances and emerging trends. \textbf{(iii)} We provide open-access resources, including datasets, benchmarks, and implementation details, to facilitate future research. \textbf{(iv)} Finally, we identify research gaps and offer insights to guide future directions in this rapidly evolving field.

\section{Datasets, Evaluation, and Applications}
\noindent
We review diverse datasets and benchmarks supporting tasks such as multimodal summarization, visual QA, video understanding, and more. For full details, see Appendix~(\S\ref{sec:app_dataset}) and Tables~\ref{tab:categorized_datasets} and \ref{tab:benchmark_datasets}.
Multimodal RAG has been applied across various domains, including healthcare, software engineering, fashion, entertainment, and emerging fields. An overview of tasks and applications are detailed in Appendix~(\S\ref{sec_app}) and ~\autoref{fig:app_taxonomy}.
Evaluating these systems requires multiple metrics, covering retrieval performance, generation quality, and modality alignment. The complete evaluation methods, metrics, and their definitions and formulations are in Appendix~(\S\ref{sec:evalmetrics}).

\begin{center}

\begin{figure*}[t!]
    \centering
    \resizebox{0.85 \textwidth}{!}{
        \begin{forest}
            forked edges,
            for tree={
                child anchor=west,
                parent anchor=east,
                grow'=east,
                anchor=west,
                base=left,
                font=\normalsize,
                rectangle,
                draw=hidden-black,
                rounded corners,
                minimum height=2em,
                minimum width=4em,
                edge+={darkgray, line width=1pt},
                s sep=3pt,
                inner xsep=0.4em,
                inner ysep=0.6em,
                line width=0.8pt,
                text width=8.5em,
                where level=1{ 
                text width=5em,
                inner xsep=0.3em,
                inner ysep=0.5em,
                font=\small,
                }{},
                where level=2{ 
                text width=10em,
                }{},
                where level=3{ 
                text width=7em,
                }{},
                ver/.style={
                    fill=white!50, 
                    rotate=90,
                    child anchor=north,
                    parent anchor=south,
                    anchor=center,
                    text width=8em
                },
                leaf/.style={
                    text opacity=1,
                    inner sep=2pt,
                    fill opacity=.5,
                    fill=green!20, 
                    text=black,
                    text width=44.5em
                    font=\normalsize,
                    inner xsep=0.4em,
                    inner ysep=0.6em,
                    draw,
                }, 
            },
            [Multimodal RAG, ver
                [
                    Retrieval \\ Strategies\\~(\S\ref{sec_retrieval_strategies}), fill=yellow!20
                    [
                        Efficient-Search and \\ Similarity Retrieval \\~(\S\ref{sec_similarity_search}), fill=red!15
                        [
                            Maximum Inner Product Search (MIPS), fill=blue!15
                            [   
                                TPU-KNN~\cite{chern2022tpuknnknearestneighbor}{,} ScaNN~\cite{guo2020acceleratinglargescaleinferenceanisotropic}{,}
                                MAXSIM score~\cite{chan2008maxsim}{,}
                                ADQ~\cite{li2024adaptivedatasetquantization}{,}
                                \citet{Zhang_Lian_Zhang_Wang_Chen_2023}{,}
                                BanditMIPS~\cite{tiwari2024faster}{,} MUST~\cite{wang2023musteffectivescalableframework}{,}
                                FARGO~\cite{10.14778/3579075.3579084}{,}
                                MuRAG~\cite{chen-etal-2022-murag}{,}
                                RA-CM3~\cite{yasunaga2023retrievalaugmentedmultimodallanguagemodeling}{,}
                                \citet{nguyen2024multimodallearnedsparseretrieval}{,}
                                Graph-based ANNs~\cite{zhang2024efficienteffectiveretrievaldensesparse}{,}
                                \citet{10.1145/3580305.3599897}{,} Deeperimpact~\cite{basnet2024deeperimpactoptimizingsparselearned}{,} RetrievalAttention~\cite{liu2024retrievalattentionacceleratinglongcontextllm}{,}
                                FactMM-RAG~\cite{sun2024factawaremultimodalretrievalaugmentation}
                                , leaf, text width=53em
                            ]
                        ]
                        [
                            Multimodal \\ Encoders, fill=blue!15
                            [
                                CLIP~\cite{radford2021learning}{,}
                                BLIP~\cite{li2022blip}{,}  MARVEL~\cite{zhou2024marvelunlockingmultimodalcapability}{,}
                                ALIGN~\cite{pmlr-v139-jia21b}{,}
                                FLAVA~\cite{singh2022flava}{,}
                                UniVL-DR~\cite{liu2023universal}{,}
                                UniIR~\cite{wei2023uniirtrainingbenchmarkinguniversal}{,} GME~\cite{zhang2024gmeimprovinguniversalmultimodal}{,}
                                VISTA~\cite{zhou-etal-2024-vista}{,} ColPali~\cite{faysse2024colpaliefficientdocumentretrieval}{,} InternVideo~\cite{wang2022internvideogeneralvideofoundation}{,} Ovis~\cite{lu2024ovisstructuralembeddingalignment}{,}
                                LLaVE~\cite{lan2025llave}
                                Mi-RAG~\cite{omar-etal-2024-multi}
                                , leaf, text width=53em
                            ]
                        ]
                    ]
                    [
                        Modality-Centric \\ Retrieval~(\S\ref{sec_modality_based}), fill=red!15
                        [
                            Text-Centric, fill=blue!15
                            [
                                Contriever~\cite{izacard2022unsuperviseddenseinformationretrieval}{,} GTE~\cite{li2023generaltextembeddingsmultistage}{,}
                                Re-Imagen~\cite{chen2022reimagenretrievalaugmentedtexttoimagegenerator}{,}
                                BM25~\cite{INR-019}{,}
                                MiniLM~\cite{wang2020minilmdeepselfattentiondistillation}{,}
                                BGE-M3~\cite{chen2024bgem3embeddingmultilingualmultifunctionality}{,} CapRet~\cite{shohan2024xlheadtagsleveragingmultimodalretrieval}{,} 
                                OMG-QA~\cite{nan-etal-2024-omg}{,}
                                ColBERT~\cite{10.1145/3397271.3401075}{,} PreFLMR~\cite{lin-etal-2024-preflmr}{,}
                                RAFT~\cite{zhang2024raft}{,}
                                CRAG~\cite{yan2024corrective}{,}
                                M2-RAG~\cite{ma2024multimodalretrievalaugmentedmultimodal}
                                , leaf, text width=53em
                            ]
                        ]
                        [
                            Vision-Centric, fill=blue!15
                            [
                                VQA4CIR~\cite{feng2023vqa4cirboostingcomposedimage}{,}
                                Unifashion~\cite{zhao-etal-2024-unifashion}{,}
                                Jang et al.~\cite{jang2024visual}{,} Pic2word~\cite{saito2023pic2wordmappingpictureswords}{,} eClip~\cite{kumar2024improvingmedicalmultimodalcontrastive}{,} RAMM~\cite{yuan2023rammretrievalaugmentedbiomedicalvisual}{,}
                                Joshi et al.~\cite{joshi2024robust}{,} VISA~\cite{ma2024visaretrievalaugmentedgeneration}{,} ImgRet~\cite{shohan2024xlheadtagsleveragingmultimodalretrieval}{,}
                                EchoSight~\cite{Yan_2024}{,}
                                Xue et al.~\cite{xue2024enhancedmultimodalragllmaccurate}
                                , leaf, text width=53em
                            ]
                        ]
                        [
                            Video-Centric, fill=blue!15
                            [
                                iRAG~\cite{Arefeen_2024}{,} VideoRAG~\cite{ren2025videoragretrievalaugmentedgenerationextreme}{,} VideoRAG~\cite{jeong2025videoragretrievalaugmentedgenerationvideo}{,}
                                T-Mass~\cite{wang2024textmassmodelingstochastic}{,}
                                MV-Adapter~\cite{jin2024mv}{,} OmAgent~\cite{zhang-etal-2024-omagent}{,}
                                CM2~\cite{kim2024you}{,}
                                Video-RAG~\cite{luo2024videoragvisuallyalignedretrievalaugmentedlong}{,}
                                CTCH~\cite{ijcai2024p136}{,}
                                RTime~\cite{10.1145/3664647.3680731}{,} VideoMAE~\cite{tong2022videomaemaskedautoencodersdataefficient}{,}  DrVideo~\cite{ma2024drvideodocumentretrievalbased}
                                , leaf, text width=53em
                            ]
                        ]
                        [
                            Audio-Centric, fill=blue!15
                            [
                                CA-CLAP~\cite{xue2024retrievalaugmentedgenerationpromptbased}{,} Recap~\cite{10448030}{,} SpeechRAG~\cite{ 10888900}{,} WavRAG~\cite{chen2025wavragaudiointegratedretrievalaugmented}{,} SEAL~\cite{sun2025sealspeechembeddingalignment}{,}
                                Audiobox TTA-RAG~\cite{yang2024audioboxttaragimprovingzeroshot}{,} DRCap~\cite{10890325}{,} P2PCAP~\cite{changin2024audiocaptioningraggenerative}{,}
                                LA-RAG~\cite{li2024laragenhancingllmbasedasraccuracy}{,}
                                \citet{10890057} 
                                , leaf, text width=53em
                            ]
                        ]                        
                        [
                            Document \\ Retrieval, fill=blue!15
                            [ 
                                ColPali~\cite{faysse2024colpaliefficientdocumentretrieval}{,}
                                ColQwen2~\cite{Qwen2VL}{,} M3DocVQA~\cite{cho2024m3docragmultimodalretrievalneed}{,}
                                ViTLP~\cite{mao-etal-2024-visually}{,}
                                DocLLM~\cite{wang-etal-2024-docllm}{,}
                                CREAM~\cite{10.1145/3664647.3680750}{,}
                                mPLUG-DocOwl 1.5~\cite{hu-etal-2024-mplug}{,}
                                mPLUG-DocOwl 2~\cite{hu2024mplugdocowl2highresolutioncompressingocrfree}{,} VisDom~\cite{suri2024visdommultidocumentqavisually}{,} DSE~\cite{ma2024unifyingmultimodalretrievaldocument}
                                {,}
                                SV-RAG~\cite{chen2025svragloracontextualizingadaptationmllms}
                                , leaf, text width=53em
                            ]
                        ]
                    ]
                    [
                        Re-ranking \\ Strategies~(\S\ref{sec_reranking_strategies}), fill=red!15
                        [
                            Optimized \\ Example \\ Selection, fill=blue!15
                            [ 
                            MSIER~\cite{luo2024doestextualinformationaffect}{,}
                                Hybrid RAG~\cite{su2024hybrid}{,}
                                RULE~\cite{xia-etal-2024-rule}{,} RAMM~\cite{yuan2023rammretrievalaugmentedbiomedicalvisual}{,} M2RAAP~\cite{10.1145/3626772.3657833}
                                , leaf, text width=53em
                            ]
                        ]
                        [
                            Relevance Score \\ Evaluation, fill=blue!15
                            [
                                RAG-Check~\cite{mortaheb2025ragcheckevaluatingmultimodalretrieval, mortaheb2025rerankingcontextmultimodalretrieval}{,}
                                UniRaG~\cite{10535103}{,}        MR2AG~\cite{zhang2024mr2agmultimodalretrievalreflectionaugmentedgeneration}{,}
                                LDRE~\cite{10.1145/3626772.3657740}{,} BM25~\cite{INR-019}{,} RAGTrans~\cite{10.1145/3637528.3672041}{,} OMG-QA~\cite{nan-etal-2024-omg}{,} EchoSight~\cite{Yan_2024}{,} EgoInstructor~\cite{xu2024retrievalaugmentedegocentricvideocaptioning}{,}
                                VR-RAG~\cite{khan2025vr}
                                , leaf, text width=53em
                            ]
                        ]
                        [
                            Filtering \\ Mechanisms, fill=blue!15
                            [
                                MAIN-RAG~\cite{chang2024mainragmultiagentfilteringretrievalaugmented}{,}
                                MM-Embed~\cite{lin2024mmembeduniversalmultimodalretrieval}{,} GME~\cite{zhang2024gmeimprovinguniversalmultimodal}{,}
                                MuRAR~\cite{zhu2024murarsimpleeffectivemultimodal}
                                RAFT~\cite{zhang2024raft}
                                , leaf, text width=53em
                            ]
                        ]
                    ]
                ]
                [
                    Fusion \\ Mechanisms\\~(\S\ref{sec_fusion_mechanisms}), fill=yellow!20
                    [
                        Score Fusion \\ and Alignment~(\S\ref{sec_socre_alignment}), fill=red!15
                        [
                            M3~\cite{cai2025matryoshka}
                            ~\citet{10535103}{,} ~\citet{Sharifymoghaddam2024UniRAGUR}{,} REVEAL~\cite{Hu_2023_CVPR}{,}
                            RAG-Driver~\cite{yuan2024ragdrivergeneralisabledrivingexplanations}{,}
                            C3Net~\cite{zhang2024c3net}{,}
                            LLM-RA~\cite{jian-etal-2024-large}{,}
                            ~\citet{riedler2024textoptimizingragmultimodal}{,} VISA~\cite{ma2024visaretrievalaugmentedgeneration}{,}
                            MA-LMM~\cite{he2024ma}{,}
                            ~\citet{xue2024enhancedmultimodalragllmaccurate}{,}
                            RA-BLIP~\cite{ding2024rablipmultimodaladaptiveretrievalaugmented}{,}
                            Re-IMAGEN~\cite{chen2022reimagenretrievalaugmentedtexttoimagegenerator}{,} MegaPairs~\cite{zhou2024megapairsmassivedatasynthesis}{,}
                            Wiki-LLaVA~\cite{caffagni2024wiki}{,} VISRAG~\cite{yu2024visragvisionbasedretrievalaugmentedgeneration}
                            , leaf, text width=62.0em
                        ]
                    ]
                    [
                        Attention-Based \\ Mechanisms~(\S\ref{sec_attention_based}), fill=red!15
                        [
                            RAMM~\cite{yuan2023rammretrievalaugmentedbiomedicalvisual}{,} EMERGE~\cite{zhu2024emergeintegratingragimproved}{,} MORE~\cite{cui2024moremultimodalretrievalaugmented}{,} RAGTrans~\cite{10.1145/3637528.3672041}{,} AlzheimerRAG~\cite{lahiri2024alzheimerragmultimodalretrievalaugmented}{,}
                            MV-Adapter~\cite{jin2024mv}{,} ~\citet{xu2024retrievalaugmentedegocentricvideocaptioning}{,}
                            ~\citet{kim2024you}{,}
                            M2-RAAP~\cite{10.1145/3626772.3657833}{,}
                            Mu-RAG~\cite{chen-etal-2022-murag}{,}
                            Ou et al.~\cite{10.1007/s00530-024-01649-6}{,}  CADMR~\cite{khalafaoui2024cadmrcrossattentiondisentangledlearning}  
                            , leaf, text width=62.0em
                        ]
                    ]
                    [
                        Unified Frameworks \\and Projections~(\S\ref{sec_unified_frameworks_projections}), fill=red!15
                        [
                            Hybrid-RAG~\cite{su2024hybrid}{,} Dense2Sparse~\cite{nguyen2024multimodallearnedsparseretrieval}{,}
                            IRAMIG~\cite{Liu_2024}{,} M3DocRAG~\cite{cho2024m3docragmultimodalretrievalneed}{,}
                            DQU-CIR~\cite{Wen_2024}{,}
                            PDF-MVQA~\cite{ding2024pdfmvqadatasetmultimodalinformation}{,}
                            SAM-RAG~\cite{zhai2024selfadaptivemultimodalretrievalaugmentedgeneration}{,} UFineBench~\cite{zuo2024ufinebench}{,}
                            ~\citet{li2022blip}
                            , leaf, text width=62.0em
                        ]
                    ]
                ]
                [
                    Augmentation \\ Techniques\\~(\S\ref{sec_augmentation}), fill=yellow!20
                    [
                        Context-Enrichment~(\S\ref{sec_context_enrichment}), fill=red!15
                        [
                            EMERGE~\cite{zhu2024emergeintegratingragimproved}{,}
                            MiRAG~\cite{omar-etal-2024-multi}{,}
                            Wiki-LLaVA~\cite{caffagni2024wiki}{,}
                            Video-RAG~\cite{luo2024videoragvisuallyalignedretrievalaugmentedlong}{,}
                            Img2Loc~\cite{zhou2024img2loc}{,}
                            ~\citet{xue2024enhancedmultimodalragllmaccurate}
                            , leaf, text width=62.0em
                        ]
                    ]
                    [
                        Adaptive and Iterative \\ Retrieval~(\S\ref{sec_adaptive}), fill=red!15
                        [
                              SKURG~\cite{SKURG}{,}
                              IRAMIG~\cite{Liu_2024}{,}
                              OMG-QA~\cite{nan-etal-2024-omg}{,}
                              SAM-RAG~\cite{zhai2024selfadaptivemultimodalretrievalaugmentedgeneration}{,}
                              MMed-RAG~\cite{xia2024mmedragversatilemultimodalrag}{,}
                              OmniSearch~\cite{li2024benchmarking}{,}
                              mR$^2$AG~\cite{Zhang2024mR2AGMR}{,}
                              RAGAR~\cite{khaliq-etal-2024-ragar}{,}
                              UniversalRAG~\cite{yeo2025universalrag}{,}
                              OMGM~\cite{yang2025omgm}
                            , leaf, text width=62.0em
                        ]
                    ]                 
                ]
                [
                    Generation \\ Techniques\\~(\S\ref{sec_generation}), fill=yellow!20 
                    [
                        In-Context Learning ~(\S\ref{sec_incontext_learning}), fill=red!15
                        [
                            RMR~\cite{tan2024retrievalmeetsreasoninghighschool}{,}
                            ~\citet{Sharifymoghaddam2024UniRAGUR}{,}
                            RA-CM3~\cite{yasunaga2023retrievalaugmentedmultimodallanguagemodeling}{,}
                            RAG-Driver~\cite{yuan2024ragdrivergeneralisabledrivingexplanations}{,}
                            MSIER~\cite{luo2024doestextualinformationaffect}{,}
                            Raven~\cite{rao2024ravenmultitaskretrievalaugmented}
                            , leaf, text width=62.0em
                        ]
                    ]
                    [
                        Reasoning~(\S\ref{sec_reasoning}), fill=red!15
                        [
                            RAGAR~\cite{khaliq-etal-2024-ragar}{,}
                            VisDoM~\cite{suri2024visdommultidocumentqavisually}{,}
                            SAM-RAG~\cite{zhai2024selfadaptivemultimodalretrievalaugmentedgeneration}{,}
                            LDRE~\cite{10.1145/3626772.3657740}
                            , leaf, text width=62.0em
                        ]
                    ]
                    [
                        Instruction Tuning ~(\S\ref{sec_instruction_tuning}), fill=red!15
                        [
                            RA-BLIP~\cite{ding2024rablipmultimodaladaptiveretrievalaugmented}{,}
                            RAGPT~\cite{lang2025retrieval}{,}
                            mR$^2$AG~\cite{Zhang2024mR2AGMR}{,}
                            RagVL~\cite{chen2024mllm}{,}
                            \citet{jang2024visual}{,}
                            MMed-RAG~\cite{xia2024mmedragversatilemultimodalrag}{,}
                            MegaPairs~\cite{zhou2024megapairsmassivedatasynthesis}{,}
                            Surf~\cite{sun2024surfteachinglargevisionlanguage}{,}
                            Rule~\cite{xia-etal-2024-rule}
                            , leaf, text width=62.0em
                        ]
                    ]
                    [
                        Source Attribution ~(\S\ref{sec_source_attrib}), fill=red!15
                        [
                            MuRAR~\cite{zhu2024murarsimpleeffectivemultimodal}{,}
                            VISA~\cite{ma2024visaretrievalaugmentedgeneration}{,}
                            OMG-QA~\cite{nan-etal-2024-omg}                          
                            , leaf, text width=62.0em
                        ]
                    ]
                    [
                        Agentic Generation and Interaction ~(\S\ref{sec_agents}), fill=red!15
                        [
                          AppAgent v2 ~\cite{li2024appagentv2advancedagent}{,}
                          USER-LLM R1~\cite{rahimi2025reasoningllmsuserawaremultimodal}{,} MMAD~\cite{jiang2025mmadcomprehensivebenchmarkmultimodal}{,} ~\citet{yi2025multimodalmultiagentframeworkradiology}{,} CollEX~\cite{schneider2025collexmultimodalagentic}{,}
                          HM-RAG~\cite{liu2025hmraghierarchicalmultiagentmultimodal}{,}
                         CogPlanner~\cite{yu2025unveilingpotentialmultimodalretrieval}
                            , leaf, text width=62.0em
                        ]
                    ]
                ]
                [
                    Training \\ Strategies\\~(\S\ref{sec_training_strategies}), fill=yellow!20
                    [
                        Alignment~(\S\ref{sec_alignment}), fill=red!15
                        [
                            VISRAG~\cite{yu2024visragvisionbasedretrievalaugmentedgeneration}{,}
                            MegaPairs~\cite{zhou2024megapairsmassivedatasynthesis}{,}
                            SAM-RAG~\cite{zhai2024selfadaptivemultimodalretrievalaugmentedgeneration}{,}
                            EchoSight~\cite{Yan_2024}{,}
                            HACL~\cite{jiang2024hallucination}{,}
                            ~\citet{10535103}{,}
                            ~\citet{kumar2024improvingmedicalmultimodalcontrastive}{,}
                            Dense2Sparse~\cite{nguyen2024multimodallearnedsparseretrieval}
                            , leaf, text width=62.0em
                        ]
                    ]     
                    [
                        Robustness~(\S\ref{sec:robustness}), fill=red!15
                        [
                            ~\citet{buettner-kovashka-2024-quantifying}{,}
                            MORE~\cite{cui2024moremultimodalretrievalaugmented}{,}
                          AlzheimerRAG~\cite{lahiri2024alzheimerragmultimodalretrievalaugmented}{,}
                          RAGTrans~\cite{10.1145/3637528.3672041}{,}
                          RA-BLIP~\cite{ding2024rablipmultimodaladaptiveretrievalaugmented}{,}
                          RagVL~\cite{chen2024mllm}{,}
                          RA-CM3~\cite{yasunaga2023retrievalaugmentedmultimodallanguagemodeling}
                            , leaf, text width=62.0em
                        ]
                    ]
                    ]
                ]
            ]
        \end{forest}   
    }
    \caption{Taxonomy of recent advances in Multimodal RAG. Refer to Appendix~(\S\ref{sec:taxonomy_details}) for further details.}
    \label{fig:taxonomy_full}
    \vspace{-5mm}
\end{figure*}
\end{center}

\vspace{-8mm}
\section{Key Innovations and Methodologies}
\subsection{Retrieval Strategy}
\label{sec_retrieval_strategies}
\paragraph{Efficient Search and Similarity Retrieval} \label{sec_similarity_search}
Modern multimodal RAG systems encode diverse input modalities into a unified embedding space to enable direct cross-modal retrieval. Early CLIP-based \cite{radford2021learning} methods often struggled to balance retrieval precision and computational cost. BLIP-inspired \cite{li2022blip} approaches addressed some of these trade-offs by integrating cross-modal attention during training, yielding richer alignments between visual and textual features. To reconcile high accuracy with efficiency, contrastive retrieval frameworks such as MARVEL \cite{zhou2024marvelunlockingmultimodalcapability} and Uni-IR \cite{wei2023uniirtrainingbenchmarkinguniversal} improved inter-modal discrimination through hard-negative mining and balanced sampling strategies \cite{zhang2024gmeimprovinguniversalmultimodal, lan2025llave}. Despite these representational gains, direct search over millions of embeddings demands fast similarity computation. Maximum inner product search (MIPS) variants offer sublinear lookup by approximating top-$k$ inner products \cite{tiwari2024faster, wang2023musteffectivescalableframework, 10.14778/3579075.3579084}. However, coarse quantization can degrade recall. To mitigate this, adaptive quantization methods \cite{Zhang_Lian_Zhang_Wang_Chen_2023, li2024adaptivedatasetquantization} dynamically allocate bits where the embedding distribution is dense, resulting in recall improvements over static schemes. Hybrid sparse–dense retrieval \cite{nguyen2024multimodallearnedsparseretrieval, zhang2024efficienteffectiveretrievaldensesparse} further complements dense embeddings with sparse lexical signals. Systems such as MuRAG \cite{chen-etal-2022-murag} and RA-CM3 \cite{yasunaga2023retrievalaugmentedmultimodallanguagemodeling} employ approximate MIPS for efficient top-k candidate retrieval from large collections of image–text embeddings. Large-scale implementations leverage distributed MIPS techniques, such as TPU-KNN \cite{chern2022tpuknnknearestneighbor}, for high-speed retrieval. Other efficient similarity computation methods include ScaNN (Scalable Nearest Neighbors) \cite{guo2020acceleratinglargescaleinferenceanisotropic}, MAXSIM score \cite{chan2008maxsim, cho2024m3docragmultimodalretrievalneed}, and approximate KNN methods \cite{caffagni2024wiki}. Emerging approaches explore learned index structures \cite{10.1145/3580305.3599897, basnet2024deeperimpactoptimizingsparselearned}, which embed the search tree itself in neural parameters, aiming to adapt retrieval paths to the data distribution and reduce both latency and storage overhead. 

\vspace{-1pt}
\paragraph{Modality-Based Retrieval} \label{sec_modality_based}
Modality-aware retrieval techniques optimize efficiency by leveraging the unique characteristics of each modality.
\textit{\textbf{(i) Text-centric retrieval}} remains foundational in multimodal RAG systems, with both traditional methods like BM25 \cite{INR-019} and dense retrievers such as MiniLM \cite{wang2020minilmdeepselfattentiondistillation} and BGE-M3 \cite{chen2024bgem3embeddingmultilingualmultifunctionality} dominating text-based evidence retrieval \cite{chen2022reimagenretrievalaugmentedtexttoimagegenerator, suri2024visdommultidocumentqavisually, nan-etal-2024-omg}. Novel approaches also address the need for fine-grained semantic matching and domain specificity: For instance, ColBERT \cite{10.1145/3397271.3401075} and PreFLMR \cite{lin-etal-2024-preflmr} employ token-level interaction mechanisms that preserve nuanced textual details to improve precision for multimodal queries, while RAFT \cite{zhang2024raft} and CRAG \cite{yan2024corrective} enhance retrieval by ensuring accurate citation of text spans. \textit{\textbf{(ii) Vision-centric retrieval}} leverages image representations for knowledge extraction \cite{kumar2024improvingmedicalmultimodalcontrastive, yuan2023rammretrievalaugmentedbiomedicalvisual}. Systems such as EchoSight \cite{Yan_2024} and ImgRet \cite{shohan2024xlheadtagsleveragingmultimodalretrieval} retrieve visually similar content by using reference images as queries. In addition, composed image retrieval methods \cite{feng2023vqa4cirboostingcomposedimage, zhao-etal-2024-unifashion, jang2024visual, saito2023pic2wordmappingpictureswords} integrate multiple image features into unified query representations, enabling zero-shot image retrieval. \textit{\textbf{(iii) Video-centric retrieval}} extends vision-based techniques by incorporating temporal dynamics and large video-language models. For instance, iRAG \cite{Arefeen_2024} enables incremental retrieval for sequential video understanding, addressing the need for temporal coherence, while T-Mass \cite{wang2024textmassmodelingstochastic} uses stochastic text embeddings to improve robustness in text-video alignment. Tackling long-context processing, Video-RAG \cite{luo2024videoragvisuallyalignedretrievalaugmentedlong} avoids reliance on proprietary models by using auxiliary OCR/ASR texts, whereas VideoRAG \cite{ren2025videoragretrievalaugmentedgenerationextreme} employs dual-channel architectures and graph-based knowledge grounding for extreme-length videos. To capture temporal reasoning, CTCH \cite{ijcai2024p136} applies contrastive transformer hashing for long-term dependencies, which RTime \cite{10.1145/3664647.3680731} further refines by introducing reversed-video hard negatives for more robust causality benchmarking. Finally, OmAgent \cite{zhang-etal-2024-omagent} addresses the challenge of complex video understanding with a divide-and-conquer framework, while DRVideo \cite{ma2024drvideodocumentretrievalbased} takes a complementary document-centric approach to enhance narrative preservation.
\textit{\textbf{(iv) Audio-centric retrieval}} aims to bypass traditional ASR pipelines while improving contextual alignment and real-time processing \cite{xue2024retrievalaugmentedgenerationpromptbased, 10448030, 10888900}. Pioneering frameworks like WavRAG \cite{chen2025wavragaudiointegratedretrievalaugmented} and SEAL \cite{sun2025sealspeechembeddingalignment} introduce unified embedding architectures, directly mapping raw audio into a shared latent space to enable retrieval from hybrid knowledge bases. Audiobox TTA-RAG \cite{yang2024audioboxttaragimprovingzeroshot} conditions text-to-audio synthesis on retrieved acoustic samples, thereby enhancing zero-shot performance by enriching prompts with unlabeled audio context. 
For audio captioning, DRCap \cite{10890325} bridges the audio-text latent space of CLAP \cite{10095969} via text-only training for domain-adaptable descriptions without paired data. In parallel, P2PCAP \cite{changin2024audiocaptioningraggenerative} improves retrieval precision by regenerating captions as dynamic queries.
Further innovations address error correction and efficiency. 
LA-RAG \cite{li2024laragenhancingllmbasedasraccuracy} utilizes fine-grained speech-to-speech retrieval and forced alignment to enhance ASR accuracy through LLM in-context learning.
Meanwhile, hybrid systems, such as \citet{10890057}, integrate LLMs to correct errors in noisy environments using retrieved text/audio context.

\paragraph{Document Retrieval and Layout Understanding} 
Recent research has moved beyond traditional unimodal retrieval, developing models that process entire documents by integrating textual, visual, and layout information.
ColPali \cite{faysse2024colpaliefficientdocumentretrieval} pioneers end-to-end document image retrieval by embedding page patches with a vision-language backbone, bypassing OCR entirely. Models like ColQwen2 \cite{Qwen2VL, faysse2024colpaliefficientdocumentretrieval} and M3DocVQA \cite{cho2024m3docragmultimodalretrievalneed} extend this paradigm with dynamic resolution handling and holistic multi-page reasoning. Newer frameworks refine efficiency and layout understanding: ViTLP \cite{mao-etal-2024-visually} and DocLLM \cite{wang-etal-2024-docllm} pre-train generative models to align spatial layouts with text, while CREAM \cite{10.1145/3664647.3680750} employs coarse-to-fine retrieval with multimodal efficient tuning to balance accuracy and computational costs. Finally, mPLUG-DocOwl 1.5 \cite{hu-etal-2024-mplug} and 2 \cite{hu2024mplugdocowl2highresolutioncompressingocrfree} unify structure learning across formats (e.g., invoices, forms) without OCR dependencies, while SV-RAG \cite{chen2025svragloracontextualizingadaptationmllms} leverages MLLMs' intrinsic retrieval capabilities via dual LoRA adapters: one for evidence page retrieval and the other for question answering.

\paragraph{Re-ranking and Selection Strategies} \label{sec_reranking_strategies} 
Effective retrieval in multimodal RAG systems requires not only identifying relevant information but also prioritizing retrieved candidates. Re-ranking and selection strategies improve retrieval quality through optimized example selection, refined relevance scoring, and filtering mechanisms.
\textit{\textbf{(i) Optimized example selection}} techniques often employ multi-step retrieval, integrating both supervised and unsupervised selection approaches \cite{luo2024doestextualinformationaffect, yuan2023rammretrievalaugmentedbiomedicalvisual}. Supervised methods like \citet{su2024hybrid} enhance multimodal inputs using probabilistic control keywords, whereas RULE \cite{xia-etal-2024-rule} calibrates retrieved context via statistical methods like the Bonferroni correction \cite{Haynes2013} to mitigate factuality risks. Clustering-based key-frame selection ensures diversity in video-based retrieval \cite{10.1145/3626772.3657833}. Advanced \textit{\textbf{(ii) scoring mechanisms}} are employed by several methods to improve retrieval relevance \cite{mortaheb2025rerankingcontextmultimodalretrieval,mortaheb2025ragcheckevaluatingmultimodalretrieval, 10535103}. Multimodal similarity measures, including structural similarity index measure (SSIM) \cite{wang2020deeplearningimagesuperresolution}, normalized cross-correlation (NCC), and BERTScore \cite{zhang2020bertscoreevaluatingtextgeneration}, aid in re-ranking documents.
Some frameworks combine similarity scores derived from various modalities for more robust re-ranking. For example, VR-RAG \cite{khan2025vr} proposes a visual re-ranking framework that combines cross-modal text-image similarity with intra-modal visual similarity using DINOv2 \cite{oquab2023dinov2}, demonstrating significant improvements in open-vocabulary recognition tasks.
Hierarchical post-processing integrates passage-level and answer confidence scores for improved ranking \cite{zhang2024mr2agmultimodalretrievalreflectionaugmentedgeneration, Yan_2024, xu2024retrievalaugmentedegocentricvideocaptioning}. LDRE \cite{10.1145/3626772.3657740} employs semantic ensemble methods to adaptively weigh multiple caption features, while RAGTrans \cite{10.1145/3637528.3672041} and OMG-QA \cite{nan-etal-2024-omg} incorporate traditional ranking functions like BM25 \cite{INR-019}. \textit{\textbf{(iii) Filtering methods}} ensure high-quality retrieval by eliminating irrelevant data. Hard negative mining, as used in GME \cite{zhang2024gmeimprovinguniversalmultimodal} and MM-Embed \cite{lin2024mmembeduniversalmultimodalretrieval}, mitigates modality bias through modality-aware sampling and synthesized negatives. Similarly, consensus-based filtering, seen in MuRAR \cite{zhu2024murarsimpleeffectivemultimodal} and ColPali \cite{faysse2024colpaliefficientdocumentretrieval}, employs source attribution and multi-vector mapping to filter out low-similarity candidates. Dynamic modality filtering methods, such as RAFT \cite{zhang2024raft} and MAIN-RAG \cite{chang2024mainragmultiagentfilteringretrievalaugmented}, train retrievers to disregard confusing data, improving multimodal retrieval robustness.

\subsection{Fusion Mechanisms}
\label{sec_fusion_mechanisms}
\paragraph{Score Fusion and Alignment} \label{sec_socre_alignment}
Models in this category utilize distinct strategies to align multimodal representations. \citet{10535103} convert text, tables, and images into a single textual format using a cross-encoder trained for relevance scoring. \citet{Sharifymoghaddam2024UniRAGUR} introduce interleaved image–text pairs that vertically merge multiple few-shot images (as in LLaVA \cite{liu2023llava}), while aligning modalities via CLIP score fusion \cite{hessel-etal-2021-clipscore} and BLIP feature fusion \cite{li2022blip}. Wiki-LLaVA \cite{caffagni2024wiki}, C3Net \cite{zhang2024c3net}, \citet{riedler2024textoptimizingragmultimodal}, and MegaPairs \cite{zhou2024megapairsmassivedatasynthesis} embed images and queries into a shared CLIP space. In particular, MegaPairs \cite{zhou2024megapairsmassivedatasynthesis} scales this approach by integrating both CLIP-based and MLLM-based retrieval, fusing their scores to leverage complementary strengths, but at the cost of increased inference complexity. VISA \cite{ma2024visaretrievalaugmentedgeneration} employs the Document Screenshot Embedding (DSE) model to align textual queries with visual document representations by encoding both into a shared embedding space. REVEAL \cite{Hu_2023_CVPR} injects retrieval scores into attention layers to minimize L2-norm differences between query and knowledge embeddings, and MA-LMM \cite{he2024ma} aligns video-text embeddings via a BLIP-inspired Query Transformer \cite{li2022blip}. LLM-RA \cite{jian-etal-2024-large} concatenates text and visual embeddings into joint queries to reduce retrieval noise, while RA-BLIP \cite{ding2024rablipmultimodaladaptiveretrievalaugmented} employs a 3-layer BERT-based adaptive fusion module to unify visual–textual semantics. \citet{xue2024enhancedmultimodalragllmaccurate} use a prototype-based embedding network \cite{10204116} to map object-predicate pairs into a shared semantic space, aligning visual features with textual prototypes.
Re-IMAGEN \cite{chen2022reimagenretrievalaugmentedtexttoimagegenerator} balances creativity and entity fidelity in text-to-image synthesis via interleaved classifier-free guidance during diffusion sampling.
To improve multimodal alignment, VISRAG \cite{yu2024visragvisionbasedretrievalaugmentedgeneration} applies position-weighted mean pooling over VLM hidden states, giving higher relevance to later tokens. In contrast, RAG-Driver \cite{yuan2024ragdrivergeneralisabledrivingexplanations} aligns visual and language embeddings through visual instruction tuning and an MLP projector.

\vspace{-1mm}
\paragraph{Attention-Based Mechanisms} \label{sec_attention_based}
Attention-based methods dynamically modulate cross-modal interactions to enable fine-tuned reasoning across tasks, balancing specificity and interpretability. Cross-attention is frequently used to integrate heterogeneous modalities, as in EMERGE \cite{zhu2024emergeintegratingragimproved}, MORE \cite{cui2024moremultimodalretrievalaugmented}, and AlzheimerRAG \cite{lahiri2024alzheimerragmultimodalretrievalaugmented}, though often requiring task-specific attention heads.
RAMM \cite{yuan2023rammretrievalaugmentedbiomedicalvisual} employs a dual-stream co-attention transformer, combining self-attention and cross-attention to fuse retrieved biomedical images/texts with input data. RAGTrans \cite{10.1145/3637528.3672041} applies user-aware attention to social media features. MV-Adapter \cite{jin2024mv} introduces Cross-Modality Tying to align video-text embeddings by sharing latent factors, improving robustness but reducing granularity of modality-specific features. M2-RAAP \cite{10.1145/3626772.3657833} enhances fusion through an auxiliary caption-guided strategy that re-weights frames and text captions based on intra-modal similarity, then uses a mutual-guided alignment head to filter misaligned features via dot-product similarity and frame-to-token attention; however, this method is computationally intensive. \citet{xu2024retrievalaugmentedegocentricvideocaptioning} condition text generation on visual features using gated cross-attention, optimizing controllability but requiring aligned supervision, and Mu-RAG \cite{chen-etal-2022-murag} employs intermediate cross-attention for open-domain QA. \citet{kim2024you} leverage cross-modal memory retrieval with pre-trained CLIP ViT-L/14 to map video-text pairs into a shared space, enabling dense captioning through the attention-based fusion of retrieved memories.

\vspace{-1mm}
\paragraph{Unified Frameworks and Projections}\label{sec_unified_frameworks_projections}
Unified frameworks and projection methods consolidate multimodal inputs into coherent representations. \citet{su2024hybrid} employ hierarchical cross-chains and late fusion for healthcare data, while IRAMIG \cite{Liu_2024} iteratively integrates multimodal results into unified knowledge representations, enhancing consistency but requiring multiple reasoning passes. M3DocRAG \cite{cho2024m3docragmultimodalretrievalneed} flattens multi-page documents into a single embedding tensor, and PDF-MVQA \cite{ding2024pdfmvqadatasetmultimodalinformation} proposes a joint-grained retriever that fuses coarse-grained semantic entity representations with their fine-grained token-level textual content, creating a richer, unified representation. DQU-CIR \cite{Wen_2024} unifies raw data by converting images into text captions for complex queries and overlaying text onto images for simple ones, then fusing embeddings via MLP-learned weights. SAM-RAG \cite{zhai2024selfadaptivemultimodalretrievalaugmentedgeneration} aligns image-text modalities by generating captions for images, converting the multimodal input to unimodal text for subsequent processing. UFineBench \cite{zuo2024ufinebench} uses a shared granularity decoder for ultra-fine text–person retrieval. \citet{nguyen2024multimodallearnedsparseretrieval} introduce Dense2Sparse projection, converting dense embeddings from models like BLIP/ALBEF \cite{li2022blip} into sparse lexical vectors using layer normalization and probabilistic expansion control to optimize storage and interpretability.

\subsection{Augmentation Techniques} 
\label{sec_augmentation}
\noindent
Basic RAG systems typically retrieve content in a single step, directly passing it to generation, often leading to inefficiencies and suboptimal outputs. Augmentation techniques refine retrieved data beforehand, improving multimodal interpretation, structuring, and integration \cite{Gao2023RetrievalAugmentedGF}.

\paragraph{Context Enrichment} \label{sec_context_enrichment}
This focuses on enhancing the relevance of retrieved knowledge by refining or expanding retrieved data. General approaches incorporate additional contextual elements (e.g., text chunks, image tokens, structured data) to provide a richer grounding for generation \cite{caffagni2024wiki, xue2024enhancedmultimodalragllmaccurate}.
EMERGE \cite{zhu2024emergeintegratingragimproved} enriches context by integrating entity relationships and semantic descriptions.
MiRAG \cite{omar-etal-2024-multi} expands initial queries through entity retrieval and reformulation, enhancing subsequent stages for the visual question-answering.
Video-RAG \cite{luo2024videoragvisuallyalignedretrievalaugmentedlong} enhances long-video understanding through Query Decoupling, which reformulates user queries into structured retrieval requests to extract auxiliary multimodal context.
Img2Loc \cite{zhou2024img2loc} boosts accuracy by including both similar and dissimilar points in prompts, helping rule out implausible locations.

\paragraph{Adaptive and Iterative Retrieval} \label{sec_adaptive} 
For more complex queries, dynamic retrieval mechanisms have proven effective. Adaptive retrieval approaches optimize relevance by adjusting retrieval dynamically. 
For instance, UniversalRAG \cite{yeo2025universalrag} introduces a framework that adapts retrieval by dynamically routing queries to the most suitable corpus based on both the required modality and granularity (e.g., paragraph vs. document, clip vs. full video), thereby addressing the specific knowledge type and scope demanded by the query.
SKURG \cite{SKURG} determines the number of retrieval hops based on query complexity. SAM-RAG \cite{zhai2024selfadaptivemultimodalretrievalaugmentedgeneration} and mR$^2$AG \cite{Zhang2024mR2AGMR} dynamically assess the need for external knowledge and filter irrelevant content using MLLMs to retain only task-critical information. MMed-RAG \cite{xia2024mmedragversatilemultimodalrag} further improves retrieval precision by discarding low-relevance results, while OmniSearch \cite{li2024benchmarking} decomposes multimodal queries into structured sub-questions, planning retrieval actions in real time.
Iterative approaches refine results over multiple steps by incorporating feedback from prior iterations. For example, OMGM \cite{yang2025omgm} orchestrates a multi-step, coarse-to-fine retrieval process for knowledge-based visual question answering, starting with a broad entity search and progressively refining the selection through multimodal reranking and fine-grained textual filtering to pinpoint the most relevant knowledge, achieving superior retrieval performance in comparison to prior methods.
IRAMIG \cite{Liu_2024} improves multimodal retrieval by dynamically updating queries based on retrieved content. OMG-QA \cite{nan-etal-2024-omg} integrates episodic memory to refine retrieval across multiple rounds, ensuring continuity in reasoning. RAGAR \cite{khaliq-etal-2024-ragar} further enhances contextual consistency by iteratively adjusting retrieval based on prior responses and multimodal analysis.

\vspace{-1mm}
\subsection{Generation Techniques}
\label{sec_generation}
\noindent
\vspace{-1.8em}
\paragraph{In-Context Learning (ICL)} \label{sec_incontext_learning}
ICL with retrieval augmentation enhances reasoning in multimodal RAGs by leveraging retrieved content as few-shot examples without requiring retraining. Models such as RMR \cite{tan2024retrievalmeetsreasoninghighschool}, \citet{Sharifymoghaddam2024UniRAGUR}, and RA-CM3 \cite{yasunaga2023retrievalaugmentedmultimodallanguagemodeling}, extend this paradigm to multimodal RAG settings. RAG-Driver \cite{yuan2024ragdrivergeneralisabledrivingexplanations} refines ICL by retrieving relevant driving experiences from a memory database. MSIER \cite{luo2024doestextualinformationaffect} improves example selection with a multimodal supervised in-context examples retrieval framework, using an MLLM scorer to assess textual and visual relevance. Raven \cite{rao2024ravenmultitaskretrievalaugmented} introduces Fusion-in-Context Learning, integrating diverse in-context examples for superior performance over standard ICL.
\vspace{-0.2em}
\paragraph{Reasoning} \label{sec_reasoning}
Reasoning methods, like chain of thought (CoT), decompose complex reasoning into sequential steps, improving coherence and robustness in multimodal RAG. RAGAR \cite{khaliq-etal-2024-ragar} refines fact-checking queries and explores branching reasoning paths by introducing Chain of RAG and Tree of RAG, while VisDoM \cite{suri2024visdommultidocumentqavisually} and SAM-RAG \cite{zhai2024selfadaptivemultimodalretrievalaugmentedgeneration} integrate CoT with evidence curation and multi-stage verification to enhance accuracy and support. Notably, VisDoM performs well in scenarios where key information is distributed across modalities. LDRE \cite{10.1145/3626772.3657740} applies LLMs for divergent compositional reasoning by refining captions using dense descriptions and textual modifications, achieving superior zero-shot composed image retrieval results.

\paragraph{Instruction Tuning} \label{sec_instruction_tuning} Several works have fine-tuned or instruct-tuned generation components for specific applications. RA-BLIP \cite{ding2024rablipmultimodaladaptiveretrievalaugmented} leverages the Q-Former architecture from InstructBLIP \cite{dai2023instructblipgeneralpurposevisionlanguagemodels} to extract visual features based on question instructions, while RAGPT \cite{lang2025retrieval} employs a context-aware prompter to generate dynamic prompts from relevant instances. MR$^2$AG \cite{Zhang2024mR2AGMR} and RagVL \cite{chen2024mllm} train MLLMs to invoke retrieval adaptively, identify relevant evidence, and enhance ranking capabilities for improved response accuracy. \citet{jang2024visual} focus on distinguishing image differences to generate descriptive textual responses. MMed-RAG \cite{xia2024mmedragversatilemultimodalrag} applies preference fine-tuning to help models balance retrieved knowledge with internal reasoning. To improve generation quality, MegaPairs \cite{zhou2024megapairsmassivedatasynthesis} and Surf \cite{sun2024surfteachinglargevisionlanguage} construct multimodal instruction-tuning datasets from prior LLM errors, while Rule \cite{xia-etal-2024-rule} refines a medical large vision language model through direct preference optimization to mitigate overreliance on retrieved contexts.

\paragraph{Source Attribution and Evidence Transparency} \label{sec_source_attrib} Ensuring source attribution in multimodal RAG systems is a significant research focus. OMG-QA \cite{nan-etal-2024-omg} prompts LLMs for explicit evidence citation in generated responses. MuRAR \cite{zhu2024murarsimpleeffectivemultimodal} refines an LLM's initial response by integrating multimodal information from a source-based retriever to improve informativeness. However, its recall is constrained, as the retriever may miss evidence spanning different sections or web documents. Similarly, VISA \cite{ma2024visaretrievalaugmentedgeneration} employs vision-language models to generate answers with visual source attribution by highlighting evidence in retrieved screenshots. Nevertheless, its attribution accuracy degrades when evidence spans multiple sections or requires cross-modal integration.

\vspace{-0.2em}
\paragraph{Agentic Generation and Interaction}
\label{sec_agents}
Agent-driven multimodal RAG uses versatile autonomous/semi-autonomous systems across diverse interaction paradigms and specialized domains, often generating complex outputs. For user interaction, AppAgent v2 \cite{li2024appagentv2advancedagent} enables mobile GUI navigation while USER-LLM R1 \cite{rahimi2025reasoningllmsuserawaremultimodal} creates personalized conversational agents via dynamic profiling, particularly for elderly users.
In specialized applications, MMAD \cite{jiang2025mmadcomprehensivebenchmarkmultimodal} addresses industrial anomaly detection with training-free enhancement strategies, \citet{yi2025multimodalmultiagentframeworkradiology} improve clinical report generation while reducing hallucination, and CollEX \cite{schneider2025collexmultimodalagentic} facilitates scientific collection exploration for researchers and learners.
For complex reasoning, HM-RAG \cite{liu2025hmraghierarchicalmultiagentmultimodal} coordinates hierarchical multi-agent collaboration across multimodal data streams, while CogPlanner \cite{yu2025unveilingpotentialmultimodalretrieval} introduces a cognitively inspired planning framework that iteratively refines queries and selects retrieval strategies adaptively.

\subsection{Training Strategies}
\label{sec_training_strategies}
\noindent
Training multimodal RAG models follows a multi-stage process to effectively capture cross-modal interactions \cite{chen-etal-2022-murag}. Pretraining on large paired datasets establishes cross-modal relationships, while fine-tuning adapts models to task-specific objectives by aligning outputs with task requirements \cite{ye2019cross}.
For example, REVEAL \cite{Hu_2023_CVPR} integrates multiple training objectives. Its pretraining phase optimizes Prefix Language Modeling Loss (\(L_{\text{PrefixLM}}\)), where text is predicted from a given prefix and an associated image. Supporting losses include Contrastive Loss (\(L_{\text{contra}}\)) which aligns queries with pseudo-ground-truth knowledge, Disentangled Regularization Loss (\(L_{\text{decor}}\)) to enhance embedding expressiveness, and Alignment Regularization Loss (\(L_{\text{align}}\)) to refine query-knowledge alignment. Fine-tuning employs a cross-entropy objective for downstream tasks like visual question answering or image captioning. Details on robustness advancements and loss formulations are in Appendix~(\S\ref{sec:training}).

\paragraph{Alignment} \label{sec_alignment}
Contrastive learning improves representation quality by pulling positive pairs closer and pushing negative pairs apart in the embedding space. The InfoNCE loss \cite{oord2019representationlearningcontrastivepredictive} is widely employed in multimodal RAG models, including VISRAG \cite{yu2024visragvisionbasedretrievalaugmentedgeneration}, MegaPairs \cite{zhou2024megapairsmassivedatasynthesis}, and SAM-RAG \cite{zhai2024selfadaptivemultimodalretrievalaugmentedgeneration}, to improve retrieval-augmented generation.
Several models introduce refinements to contrastive training. EchoSight \cite{Yan_2024} enhances retrieval accuracy by selecting visually similar yet semantically distinct negatives, while HACL \cite{jiang2024hallucination} mitigates hallucinations by incorporating adversarial captions as distractors. Similarly, UniRaG \cite{10535103} improves retrieval robustness by leveraging hard negative documents to help the model discriminate between relevant and irrelevant contexts.
The eCLIP loss \cite{kumar2024improvingmedicalmultimodalcontrastive} extends contrastive learning by integrating expert-annotated data and an auxiliary MSE loss to refine embedding quality. Mixup strategies further improve generalization by generating synthetic positive pairs \cite{kumar2024improvingmedicalmultimodalcontrastive}.
Dense2Sparse \cite{nguyen2024multimodallearnedsparseretrieval} employs image-to-caption \(\ell(I \to C)\) and caption-to-image \(\ell(C \to I)\) losses, while enforcing sparsity through \(\ell1\) regularization, optimizing retrieval precision by balancing dense and sparse representations.

\section{Open Problems and Future Directions}
\noindent
Additional challenges and future directions about long-context processing, scalability, efficiency, and personalization are discussed in Appendix~(\S\ref{sec:future}).

\paragraph{Generalization, Explainability, and Robustness}
Multimodal RAG systems often struggle with domain adaptation and exhibit modality biases, frequently over-relying on text for both retrieval and generation \cite{winterbottom2020modalitybiastvqadataset}. Explainability remains a major challenge, as these systems often attribute responses to broad sources, citing entire documents or large visual regions instead of pinpointing exact contributing elements across modalities \cite{ma2024visaretrievalaugmentedgeneration, Hu_2023_CVPR}.
Moreover, the interplay between modalities affects the outcome quality; for example, answers derived solely from text sources may differ in quality compared to those requiring a combination of text and image inputs \cite{10.1109/TPAMI.2018.2798607}.
They are also vulnerable to adversarial perturbations, such as misleading images influencing textual outputs, and their performance degrades when relying on low-quality or outdated sources \cite{chen2022reimagenretrievalaugmentedtexttoimagegenerator}. MM-PoisonRAG \cite{ha2025mmpoisonragdisruptingmultimodalrag} and Poisoned-MRAG \cite{liu2025poisoned} demonstrate that even a few adversarial knowledge injections can hijack cross-modal retrieval and derail generation, underscoring the imperative for robust defense mechanisms against knowledge poisoning in multimodal RAG systems.
While the trustworthiness of unimodal RAGs has been studied \cite{zhou2024TrustworthyRAG}, ensuring robustness in multimodal RAGs remains an open challenge and a crucial research direction.

\paragraph{Reasoning, Alignment, and Retrieval Enhancement}
Multimodal RAGs struggle with compositional reasoning, requiring logical integration of information across modalities for coherent, context-rich outputs. While cross-modal techniques like Multimodal-CoT \cite{zhang2023multicot} have emerged, further advancements are needed to enhance coherence and contextual relevance. Improving modality alignment and entity-aware retrieval is crucial. Moreover, despite the potential of knowledge graphs to enrich cross-modal reasoning, they remain underexplored in multimodal RAGs compared to text-based RAGs \cite{Zhang2024mR2AGMR, peng2024graphragsurvey}.
Retrieval biases such as position sensitivity \cite{hu2024mragbench}, redundancy \cite{nan-etal-2024-omg}, and biases from training data or retrieved content \cite{zhai2024selfadaptivemultimodalretrievalaugmentedgeneration}, pose significant challenges. A promising direction is a unified embedding space for all modalities, enabling direct multimodal search without intermediary models (e.g., ASRs). Despite progress, mapping multimodal knowledge into a unified space remains an open challenge with substantial potential.

\paragraph{Agent-Based and Self-Guided Systems}
Recent trends indicate a shift towards agent-based multimodal RAGs that integrate retrieval, reasoning, and generation across diverse domains.
Unlike static RAGs, future systems should incorporate interactive feedback and self-guided decision-making to iteratively refine outputs. Existing feedback mechanisms often fail to determine whether errors stem from retrieval, generation, or other stages \cite{10.1145/3626772.3657833}.
The incorporation of reinforcement learning and end-to-end human-aligned feedback remains largely overlooked but holds significant potential for 
assessing whether retrieval is necessary, evaluating the relevance of retrieved content, and dynamically determining the most suitable modalities for response generation. Robust support for any-to-any modality is crucial for open-ended tasks \cite{wu24next}. Future multimodal RAGs should incorporate data from diverse real-world sources, such as environmental sensors, alongside traditional modalities to enhance situational awareness.
This progression aligns with the trend toward embodied AI, where models integrate knowledge with physical interaction, enabling applications in robotics, navigation, and physics-informed reasoning.
Bridging retrieval-based reasoning with real-world agency brings these systems closer to AGI.

\section{Conclusion}
\noindent
This study provides a comprehensive review of multimodal RAG, categorizing key advancements in retrieval, multimodal fusion, augmentation, generation, training strategies, and agents. We also examine task-specific applications, datasets, benchmarks, and evaluation methods while highlighting open challenges and promising future directions.
We hope this work inspires future research, particularly in enhancing cross-modal reasoning and retrieval, developing agent-based interactive systems, and advancing unified multimodal embedding spaces.

\section{Limitations}
\noindent
This study offers a comprehensive examination of multimodal RAG systems. Extended discussions, details of datasets and benchmarks, and additional relevant work are available in the Appendices. While we have made our maximum effort; however, some limits may persist. First, due to space constraints, our descriptions of individual methodologies are necessarily concise. Second, although we curate studies from major venues (e.g., ACL, EMNLP, NeurIPS, CVPR, ICLR, ICML, ACM Multimedia) and arXiv, our selection may inadvertently overlook emerging or domain-specific research, with a primary focus on recent advancements. Additionally, this work does not include a comparative performance evaluation of the various models, as task definitions, evaluation metrics, and implementation details vary significantly across studies, and executing these models requires substantial computational resources.

\noindent
Furthermore, multimodal RAG is a rapidly evolving field with many open questions, such as optimizing fusion strategies for diverse modalities and addressing scalability challenges. As new paradigms emerge, our taxonomy and conclusions will inevitably evolve. To address these gaps, we plan to continuously monitor developments and update this survey and the corresponding repository to incorporate overlooked contributions and refine our perspectives.

\section{Ethical Statement}
\noindent
This survey provides a comprehensive review of research on multimodal RAG systems, offering insights that we believe will be valuable to researchers in this evolving field. All the studies, datasets, and benchmarks analyzed in this work are publicly available, with only a very small number of papers requiring institutional access. Additionally, this survey does not involve personal data or user interactions, and we adhere to ethical guidelines throughout.

\noindent
Since this work is purely a survey of existing literature and does not introduce new models, datasets, or experimental methodologies, it presents no potential risks. However, we acknowledge that multimodal RAG systems inherently raise ethical concerns, including bias, misinformation, privacy, and intellectual property issues. Bias can emerge from both retrieval and generation processes, potentially leading to skewed or unfair outputs. Additionally, these models may hallucinate or propagate misinformation, particularly when retrieval mechanisms fail or rely on unreliable sources. The handling of sensitive multimodal data also poses privacy risks, while content generation raises concerns about proper attribution and copyright compliance. Addressing these challenges requires careful dataset curation, bias mitigation strategies, and transparent evaluation of retrieval and generation mechanisms.

\bibliography{acl_latex}

\begin{thebibliography}{311}
\providecommand{\natexlab}[1]{#1}

\bibitem[{Abootorabi and Asgari(2024)}]{abootorabi2024claspcontrastivelanguagespeechpretraining}
Mohammad~Mahdi Abootorabi and Ehsaneddin Asgari. 2024.
\newblock \href {https://arxiv.org/abs/2412.13071} {Clasp: Contrastive language-speech pretraining for multilingual multimodal information retrieval}.
\newblock \emph{Preprint}, arXiv:2412.13071.

\bibitem[{Adjali et~al.(2024)Adjali, Ferret, Ghannay, and Le~Borgne}]{omar-etal-2024-multi}
Omar Adjali, Olivier Ferret, Sahar Ghannay, and Herv{\'e} Le~Borgne. 2024.
\newblock \href {https://doi.org/10.18653/v1/2024.emnlp-main.922} {Multi-level information retrieval augmented generation for knowledge-based visual question answering}.
\newblock In \emph{Proceedings of the 2024 Conference on Empirical Methods in Natural Language Processing}, pages 16499--16513, Miami, Florida, USA. Association for Computational Linguistics.

\bibitem[{Agostinelli et~al.(2023)Agostinelli, Denk, Borsos, Engel, Verzetti, Caillon, Huang, Jansen, Roberts, Tagliasacchi et~al.}]{agostinelli2023musiclm}
Andrea Agostinelli, Timo~I Denk, Zal{\'a}n Borsos, Jesse Engel, Mauro Verzetti, Antoine Caillon, Qingqing Huang, Aren Jansen, Adam Roberts, Marco Tagliasacchi, et~al. 2023.
\newblock Musiclm: Generating music from text.
\newblock \emph{arXiv preprint arXiv:2301.11325}.

\bibitem[{Agrawal et~al.(2019)Agrawal, Desai, Wang, Chen, Jain et~al.}]{nocaps}
Harsh Agrawal, Karan Desai, Yufei Wang, Xinlei Chen, Rishabh Jain, et~al. 2019.
\newblock Nocaps: Novel object captioning at scale.
\newblock \emph{Proceedings of the IEEE/CVF International Conference on Computer Vision (ICCV)}, pages 1--10.

\bibitem[{Alayrac et~al.(2024)Alayrac, Donahue, Luc, Miech, Barr, Hasson, Lenc, Mensch, Millicah, Reynolds, Ring, Rutherford, Cabi, Han, Gong, Samangooei, Monteiro, Menick, Borgeaud, Brock, Nematzadeh, Sharifzadeh, Binkowski, Barreira, Vinyals, Zisserman, and Simonyan}]{10.5555/3600270.3601993}
Jean-Baptiste Alayrac, Jeff Donahue, Pauline Luc, Antoine Miech, Iain Barr, Yana Hasson, Karel Lenc, Arthur Mensch, Katie Millicah, Malcolm Reynolds, Roman Ring, Eliza Rutherford, Serkan Cabi, Tengda Han, Zhitao Gong, Sina Samangooei, Marianne Monteiro, Jacob Menick, Sebastian Borgeaud, Andrew Brock, Aida Nematzadeh, Sahand Sharifzadeh, Mikolaj Binkowski, Ricardo Barreira, Oriol Vinyals, Andrew Zisserman, and Karen Simonyan. 2024.
\newblock Flamingo: a visual language model for few-shot learning.
\newblock In \emph{Proceedings of the 36th International Conference on Neural Information Processing Systems}, NIPS '22, Red Hook, NY, USA. Curran Associates Inc.

\bibitem[{An et~al.(2024)An, Ding, Fu, Chu, Li, and Du}]{an2024goldenretrieverhighfidelityagenticretrieval}
Zhiyu An, Xianzhong Ding, Yen-Chun Fu, Cheng-Chung Chu, Yan Li, and Wan Du. 2024.
\newblock \href {https://arxiv.org/abs/2408.00798} {Golden-retriever: High-fidelity agentic retrieval augmented generation for industrial knowledge base}.
\newblock \emph{Preprint}, arXiv:2408.00798.

\bibitem[{Anderson et~al.(2016)Anderson, Fernando, Johnson, and Gould}]{anderson2016spicesemanticpropositionalimage}
Peter Anderson, Basura Fernando, Mark Johnson, and Stephen Gould. 2016.
\newblock Spice: Semantic propositional image caption evaluation.
\newblock In \emph{Computer Vision--ECCV 2016: 14th European Conference, Amsterdam, The Netherlands, October 11-14, 2016, Proceedings, Part V 14}, pages 382--398. Springer.

\bibitem[{Anil et~al.(2023)Anil, Dai, Firat, Johnson, Lepikhin, Passos, Shakeri, Taropa, Bailey, Chen, Chu, Clark, Shafey, Huang, Meier-Hellstern, Mishra, Moreira, Omernick, Robinson, Ruder, Tay, Xiao, Xu, Zhang, Abrego, Ahn, Austin, Barham, Botha, Bradbury, Brahma, Brooks, Catasta, Cheng, Cherry, Choquette-Choo, Chowdhery, Crepy, Dave, Dehghani, Dev, Devlin, Díaz, Du, Dyer, Feinberg, Feng, Fienber, Freitag, Garcia, Gehrmann, Gonzalez, and et~al.}]{anil2023palm2technicalreport}
Rohan Anil, Andrew~M. Dai, Orhan Firat, Melvin Johnson, Dmitry Lepikhin, Alexandre Passos, Siamak Shakeri, Emanuel Taropa, Paige Bailey, Zhifeng Chen, Eric Chu, Jonathan~H. Clark, Laurent~El Shafey, Yanping Huang, Kathy Meier-Hellstern, Gaurav Mishra, Erica Moreira, Mark Omernick, Kevin Robinson, Sebastian Ruder, Yi~Tay, Kefan Xiao, Yuanzhong Xu, Yujing Zhang, Gustavo~Hernandez Abrego, Junwhan Ahn, Jacob Austin, Paul Barham, Jan Botha, James Bradbury, Siddhartha Brahma, Kevin Brooks, Michele Catasta, Yong Cheng, Colin Cherry, Christopher~A. Choquette-Choo, Aakanksha Chowdhery, Clément Crepy, Shachi Dave, Mostafa Dehghani, Sunipa Dev, Jacob Devlin, Mark Díaz, Nan Du, Ethan Dyer, Vlad Feinberg, Fangxiaoyu Feng, Vlad Fienber, Markus Freitag, Xavier Garcia, Sebastian Gehrmann, Lucas Gonzalez, and et~al. 2023.
\newblock \href {https://arxiv.org/abs/2305.10403} {Palm 2 technical report}.
\newblock \emph{Preprint}, arXiv:2305.10403.

\bibitem[{Anne~Hendricks et~al.(2017)Anne~Hendricks, Wang, Shechtman, Sivic, Darrell, and Russell}]{Hendricks_2017_ICCV}
Lisa Anne~Hendricks, Oliver Wang, Eli Shechtman, Josef Sivic, Trevor Darrell, and Bryan Russell. 2017.
\newblock Localizing moments in video with natural language.
\newblock In \emph{Proceedings of the IEEE International Conference on Computer Vision (ICCV)}.

\bibitem[{Antol et~al.(2015)Antol, Agrawal, Lu, Mitchell, Batra, Zitnick, and Parikh}]{antol2015vqa}
Stanislaw Antol, Aishwarya Agrawal, Jiasen Lu, Margaret Mitchell, Dhruv Batra, C~Lawrence Zitnick, and Devi Parikh. 2015.
\newblock Vqa: Visual question answering.
\newblock \emph{Proceedings of the IEEE International Conference on Computer Vision}, pages 2425--2433.

\bibitem[{Arefeen et~al.(2024)Arefeen, Debnath, Uddin, and Chakradhar}]{Arefeen_2024}
Md~Adnan Arefeen, Biplob Debnath, Md~Yusuf~Sarwar Uddin, and Srimat Chakradhar. 2024.
\newblock \href {https://doi.org/10.1145/3627673.3680088} {irag: Advancing rag for videos with an incremental approach}.
\newblock In \emph{Proceedings of the 33rd ACM International Conference on Information and Knowledge Management}, CIKM ’24, page 4341–4348. ACM.

\bibitem[{Asai et~al.(2023)Asai, Wu, Wang, Sil, and Hajishirzi}]{asai2023selfrag}
Akari Asai, Zeqiu Wu, Yizhong Wang, Avirup Sil, and Hannaneh Hajishirzi. 2023.
\newblock \href {https://arxiv.org/abs/2310.11511} {{Self-RAG}: Learning to retrieve, generate, and critique through self-reflection}.
\newblock \emph{arXiv preprint arXiv:2310.11511}.

\bibitem[{Awadalla et~al.(2024)Awadalla, Xue, Lo, Shu, Lee, Guha, Shen, Awadalla, Savarese, Xiong et~al.}]{awadalla2024mint}
Anas Awadalla, Le~Xue, Oscar Lo, Manli Shu, Hannah Lee, Etash Guha, Sheng Shen, Mohamed Awadalla, Silvio Savarese, Caiming Xiong, et~al. 2024.
\newblock Mint-1t: Scaling open-source multimodal data by 10x: A multimodal dataset with one trillion tokens.
\newblock \emph{Advances in Neural Information Processing Systems}, 37:36805--36828.

\bibitem[{Bahaj and Ghogho(2024)}]{bahaj2024asthmabot}
Adil Bahaj and Mounir Ghogho. 2024.
\newblock Asthmabot: Multi-modal, multi-lingual retrieval augmented generation for asthma patient support.
\newblock \emph{arXiv preprint arXiv:2409.15815}.

\bibitem[{Bain et~al.(2021)Bain, Nagrani, Varol, and Zisserman}]{bain2021frozen}
Max Bain, Arsha Nagrani, Gul Varol, and Andrew Zisserman. 2021.
\newblock Frozen in time: A joint video and image encoder for end-to-end retrieval.
\newblock \emph{Proceedings of the IEEE/CVF International Conference on Computer Vision (ICCV)}, pages 1--10.

\bibitem[{Baldrati et~al.(2023)Baldrati, Bertini, and Del~Bimbo}]{circo}
Alberto Baldrati, Marco Bertini, and Alberto Del~Bimbo. 2023.
\newblock Zero-shot composed image retrieval with textual inversion.
\newblock \emph{Proceedings of the IEEE/CVF International Conference on Computer Vision (ICCV)}, pages 1--10.

\bibitem[{Baltrusaitis et~al.(2019)Baltrusaitis, Ahuja, and Morency}]{10.1109/TPAMI.2018.2798607}
Tadas Baltrusaitis, Chaitanya Ahuja, and Louis-Philippe Morency. 2019.
\newblock \href {https://doi.org/10.1109/TPAMI.2018.2798607} {Multimodal machine learning: A survey and taxonomy}.
\newblock \emph{IEEE Trans. Pattern Anal. Mach. Intell.}, 41(2):423–443.

\bibitem[{Banerjee and Lavie(2005)}]{banerjee-lavie-2005-meteor}
Satanjeev Banerjee and Alon Lavie. 2005.
\newblock \href {https://aclanthology.org/W05-0909/} {{METEOR}: An automatic metric for {MT} evaluation with improved correlation with human judgments}.
\newblock In \emph{Proceedings of the {ACL} Workshop on Intrinsic and Extrinsic Evaluation Measures for Machine Translation and/or Summarization}, pages 65--72, Ann Arbor, Michigan. Association for Computational Linguistics.

\bibitem[{Basnet et~al.(2024)Basnet, Gou, Mallia, and Suel}]{basnet2024deeperimpactoptimizingsparselearned}
Soyuj Basnet, Jerry Gou, Antonio Mallia, and Torsten Suel. 2024.
\newblock \href {https://arxiv.org/abs/2405.17093} {Deeperimpact: Optimizing sparse learned index structures}.
\newblock \emph{Preprint}, arXiv:2405.17093.

\bibitem[{Bińkowski et~al.(2018)Bińkowski, Sutherland, Arbel, and Gretton}]{bińkowski2018demystifying}
Mikołaj Bińkowski, Dougal~J. Sutherland, Michael Arbel, and Arthur Gretton. 2018.
\newblock \href {https://openreview.net/forum?id=r1lUOzWCW} {Demystifying {MMD} {GAN}s}.
\newblock In \emph{International Conference on Learning Representations}.

\bibitem[{Brown et~al.(2020)Brown, Mann, Ryder, Subbiah, Kaplan, Dhariwal, Neelakantan, Shyam, Sastry, Askell, Agarwal, Herbert-Voss, Krueger, Henighan, Child, Ramesh, Ziegler, Wu, Winter, Hesse, Chen, Sigler, Litwin, Gray, Chess, Clark, Berner, McCandlish, Radford, Sutskever, and Amodei}]{NEURIPS2020_1457c0d6}
Tom Brown, Benjamin Mann, Nick Ryder, Melanie Subbiah, Jared~D Kaplan, Prafulla Dhariwal, Arvind Neelakantan, Pranav Shyam, Girish Sastry, Amanda Askell, Sandhini Agarwal, Ariel Herbert-Voss, Gretchen Krueger, Tom Henighan, Rewon Child, Aditya Ramesh, Daniel Ziegler, Jeffrey Wu, Clemens Winter, Chris Hesse, Mark Chen, Eric Sigler, Mateusz Litwin, Scott Gray, Benjamin Chess, Jack Clark, Christopher Berner, Sam McCandlish, Alec Radford, Ilya Sutskever, and Dario Amodei. 2020.
\newblock \href {https://proceedings.neurips.cc/paper_files/paper/2020/file/1457c0d6bfcb4967418bfb8ac142f64a-Paper.pdf} {Language models are few-shot learners}.
\newblock In \emph{Advances in Neural Information Processing Systems}, volume~33, pages 1877--1901. Curran Associates, Inc.

\bibitem[{Buettner and Kovashka(2024)}]{buettner-kovashka-2024-quantifying}
Kyle Buettner and Adriana Kovashka. 2024.
\newblock \href {https://doi.org/10.18653/v1/2024.emnlp-main.335} {Quantifying the gaps between translation and native perception in training for multimodal, multilingual retrieval}.
\newblock In \emph{Proceedings of the 2024 Conference on Empirical Methods in Natural Language Processing}, pages 5863--5870, Miami, Florida, USA. Association for Computational Linguistics.

\bibitem[{Caba~Heilbron et~al.(2015)Caba~Heilbron, Escorcia, Ghanem, and Carlos~Niebles}]{caba2015activitynet}
Fabian Caba~Heilbron, Victor Escorcia, Bernard Ghanem, and Juan Carlos~Niebles. 2015.
\newblock Activitynet: A large-scale video benchmark for human activity understanding.
\newblock \emph{Proceedings of the IEEE Conference on Computer Vision and Pattern Recognition (CVPR)}, pages 961--970.

\bibitem[{Caffagni et~al.(2024)Caffagni, Cocchi, Moratelli, Sarto, Cornia, Baraldi, and Cucchiara}]{caffagni2024wiki}
Davide Caffagni, Federico Cocchi, Nicholas Moratelli, Sara Sarto, Marcella Cornia, Lorenzo Baraldi, and Rita Cucchiara. 2024.
\newblock Wiki-llava: Hierarchical retrieval-augmented generation for multimodal llms.
\newblock In \emph{Proceedings of the IEEE/CVF Conference on Computer Vision and Pattern Recognition}, pages 1818--1826.

\bibitem[{Cai et~al.(2025)Cai, Yang, Gao, and Lee}]{cai2025matryoshka}
Mu~Cai, Jianwei Yang, Jianfeng Gao, and Yong~Jae Lee. 2025.
\newblock \href {https://openreview.net/forum?id=Uhj5OxAz7I} {Matryoshka multimodal models}.
\newblock In \emph{The Thirteenth International Conference on Learning Representations}.

\bibitem[{Chan and Ng(2008)}]{chan2008maxsim}
Yee~Seng Chan and Hwee~Tou Ng. 2008.
\newblock Maxsim: A maximum similarity metric for machine translation evaluation.
\newblock In \emph{Proceedings of ACL-08: HLT}, pages 55--62.

\bibitem[{Chang et~al.(2015)Chang, Funkhouser, Guibas, Hanrahan, Huang et~al.}]{chang2015shapenet}
Angel~X Chang, Thomas Funkhouser, Leonidas Guibas, Pat Hanrahan, Qixing Huang, et~al. 2015.
\newblock Shapenet: An information-rich 3d model repository.
\newblock \emph{arXiv preprint arXiv:1512.03012}.

\bibitem[{Chang et~al.(2024)Chang, Jiang, Rakesh, Pan, Yeh, Wang, Hu, Xu, Zheng, Das, and Zou}]{chang2024mainragmultiagentfilteringretrievalaugmented}
Chia-Yuan Chang, Zhimeng Jiang, Vineeth Rakesh, Menghai Pan, Chin-Chia~Michael Yeh, Guanchu Wang, Mingzhi Hu, Zhichao Xu, Yan Zheng, Mahashweta Das, and Na~Zou. 2024.
\newblock \href {https://arxiv.org/abs/2501.00332} {Main-rag: Multi-agent filtering retrieval-augmented generation}.
\newblock \emph{Preprint}, arXiv:2501.00332.

\bibitem[{Chang et~al.(2022)Chang, Narang, Suzuki, Cao, Gao, and Bisk}]{Chang_2022_CVPR}
Yingshan Chang, Mridu Narang, Hisami Suzuki, Guihong Cao, Jianfeng Gao, and Yonatan Bisk. 2022.
\newblock Webqa: Multihop and multimodal qa.
\newblock In \emph{Proceedings of the IEEE/CVF Conference on Computer Vision and Pattern Recognition (CVPR)}, pages 16495--16504.

\bibitem[{Changin et~al.(2024)Changin, Sungjun, and Wonjong}]{changin2024audiocaptioningraggenerative}
Choi Changin, Lim Sungjun, and Rhee Wonjong. 2024.
\newblock \href {https://arxiv.org/abs/2410.10913} {Audio captioning rag via generative pair-to-pair retrieval with refined knowledge base}.
\newblock \emph{Preprint}, arXiv:2410.10913.

\bibitem[{Chen and Dolan(2011)}]{chen2011collecting}
David~L Chen and William~B Dolan. 2011.
\newblock Collecting highly parallel data for paraphrase evaluation.
\newblock \emph{Proceedings of the 49th Annual Meeting of the Association for Computational Linguistics: Human Language Technologies}, pages 190--200.

\bibitem[{Chen et~al.(2025{\natexlab{a}})Chen, Zhang, Zhou, Yu, Dernoncourt, Gu, Rossi, Chen, and Sun}]{chen2025svragloracontextualizingadaptationmllms}
Jian Chen, Ruiyi Zhang, Yufan Zhou, Tong Yu, Franck Dernoncourt, Jiuxiang Gu, Ryan~A. Rossi, Changyou Chen, and Tong Sun. 2025{\natexlab{a}}.
\newblock \href {https://arxiv.org/abs/2411.01106} {Sv-rag: Lora-contextualizing adaptation of mllms for long document understanding}.
\newblock \emph{Preprint}, arXiv:2411.01106.

\bibitem[{Chen et~al.(2024{\natexlab{a}})Chen, Xiao, Zhang, Luo, Lian, and Liu}]{chen-etal-2024-m3}
Jianlyu Chen, Shitao Xiao, Peitian Zhang, Kun Luo, Defu Lian, and Zheng Liu. 2024{\natexlab{a}}.
\newblock \href {https://doi.org/10.18653/v1/2024.findings-acl.137} {{M}3-embedding: Multi-linguality, multi-functionality, multi-granularity text embeddings through self-knowledge distillation}.
\newblock In \emph{Findings of the Association for Computational Linguistics: ACL 2024}, pages 2318--2335, Bangkok, Thailand. Association for Computational Linguistics.

\bibitem[{Chen et~al.(2024{\natexlab{b}})Chen, Xiao, Zhang, Luo, Lian, and Liu}]{chen2024bgem3embeddingmultilingualmultifunctionality}
Jianlyu Chen, Shitao Xiao, Peitian Zhang, Kun Luo, Defu Lian, and Zheng Liu. 2024{\natexlab{b}}.
\newblock M3-embedding: Multi-linguality, multi-functionality, multi-granularity text embeddings through self-knowledge distillation.
\newblock In \emph{Findings of the Association for Computational Linguistics ACL 2024}, pages 2318--2335.

\bibitem[{Chen et~al.(2024{\natexlab{c}})Chen, Li, Dong, Zhang, He, Wang, Zhao, and Lin}]{chen2024sharegpt4v}
Lin Chen, Jinsong Li, Xiaoyi Dong, Pan Zhang, Conghui He, Jiaqi Wang, Feng Zhao, and Dahua Lin. 2024{\natexlab{c}}.
\newblock Sharegpt4v: Improving large multi-modal models with better captions.
\newblock In \emph{European Conference on Computer Vision}, pages 370--387. Springer.

\bibitem[{Chen et~al.(2024{\natexlab{d}})Chen, Yao, and Jiang}]{chen2024llm4design}
Ran Chen, Xueqi Yao, and Xuhui Jiang. 2024{\natexlab{d}}.
\newblock Llm4design: An automated multi-modal system for architectural and environmental design.
\newblock \emph{arXiv preprint arXiv:2407.12025}.

\bibitem[{Chen et~al.(2022{\natexlab{a}})Chen, Hu, Chen, Verga, and Cohen}]{chen-etal-2022-murag}
Wenhu Chen, Hexiang Hu, Xi~Chen, Pat Verga, and William Cohen. 2022{\natexlab{a}}.
\newblock \href {https://doi.org/10.18653/v1/2022.emnlp-main.375} {Murag: Multimodal retrieval-augmented generator for open question answering over images and text}.
\newblock In \emph{Proceedings of the 2022 Conference on Empirical Methods in Natural Language Processing}, pages 5558--5570, Abu Dhabi, United Arab Emirates. Association for Computational Linguistics.

\bibitem[{Chen et~al.(2022{\natexlab{b}})Chen, Hu, Saharia, and Cohen}]{chen2022reimagenretrievalaugmentedtexttoimagegenerator}
Wenhu Chen, Hexiang Hu, Chitwan Saharia, and William~W. Cohen. 2022{\natexlab{b}}.
\newblock \href {https://arxiv.org/abs/2209.14491} {Re-imagen: Retrieval-augmented text-to-image generator}.
\newblock \emph{Preprint}, arXiv:2209.14491.

\bibitem[{Chen et~al.(2023)Chen, Hu, Luan, Sun, Changpinyo, Ritter, and Chang}]{infoseek}
Yang Chen, Hexiang Hu, Yi~Luan, Haitian Sun, Soravit Changpinyo, Alan Ritter, and Ming-Wei Chang. 2023.
\newblock \href {https://doi.org/10.18653/v1/2023.emnlp-main.925} {Can pre-trained vision and language models answer visual information-seeking questions?}
\newblock In \emph{Proceedings of the 2023 Conference on Empirical Methods in Natural Language Processing}, pages 14948--14968, Singapore. Association for Computational Linguistics.

\bibitem[{Chen et~al.(2025{\natexlab{b}})Chen, Ji, Wang, Wang, Chen, He, Xu, and Zhao}]{chen2025wavragaudiointegratedretrievalaugmented}
Yifu Chen, Shengpeng Ji, Haoxiao Wang, Ziqing Wang, Siyu Chen, Jinzheng He, Jin Xu, and Zhou Zhao. 2025{\natexlab{b}}.
\newblock \href {https://arxiv.org/abs/2502.14727} {Wavrag: Audio-integrated retrieval augmented generation for spoken dialogue models}.
\newblock \emph{Preprint}, arXiv:2502.14727.

\bibitem[{Chen et~al.(2024{\natexlab{e}})Chen, Xu, Qi, and Guo}]{chen2024mllm}
Zhanpeng Chen, Chengjin Xu, Yiyan Qi, and Jian Guo. 2024{\natexlab{e}}.
\newblock Mllm is a strong reranker: Advancing multimodal retrieval-augmented generation via knowledge-enhanced reranking and noise-injected training.
\newblock \emph{arXiv preprint arXiv:2407.21439}.

\bibitem[{Cheng et~al.(2024)Cheng, Zhang, Xu, Trajcevski, Zhong, and Zhou}]{10.1145/3637528.3672041}
Zhangtao Cheng, Jienan Zhang, Xovee Xu, Goce Trajcevski, Ting Zhong, and Fan Zhou. 2024.
\newblock \href {https://doi.org/10.1145/3637528.3672041} {Retrieval-augmented hypergraph for multimodal social media popularity prediction}.
\newblock In \emph{Proceedings of the 30th ACM SIGKDD Conference on Knowledge Discovery and Data Mining}, KDD '24, page 445–455, New York, NY, USA. Association for Computing Machinery.

\bibitem[{Chern et~al.(2022)Chern, Hechtman, Davis, Guo, Majnemer, and Kumar}]{chern2022tpuknnknearestneighbor}
Felix Chern, Blake Hechtman, Andy Davis, Ruiqi Guo, David Majnemer, and Sanjiv Kumar. 2022.
\newblock Tpu-knn: K nearest neighbor search at peak flop/s.
\newblock \emph{Advances in Neural Information Processing Systems}, 35:15489--15501.

\bibitem[{Cho et~al.(2024)Cho, Mahata, Irsoy, He, and Bansal}]{cho2024m3docragmultimodalretrievalneed}
Jaemin Cho, Debanjan Mahata, Ozan Irsoy, Yujie He, and Mohit Bansal. 2024.
\newblock \href {https://arxiv.org/abs/2411.04952} {M3docrag: Multi-modal retrieval is what you need for multi-page multi-document understanding}.
\newblock \emph{Preprint}, arXiv:2411.04952.

\bibitem[{Choi et~al.(2025)Choi, Yoon, Kim, and Park}]{choi2025leveraging}
Kyoyun Choi, Byungmu Yoon, Soobum Kim, and Jonggwon Park. 2025.
\newblock Leveraging llms for multimodal retrieval-augmented radiology report generation via key phrase extraction.
\newblock \emph{arXiv preprint arXiv:2504.07415}.

\bibitem[{Choi et~al.(2021)Choi, Choi, Kim, Ha, Kim, and Choo}]{vitonhd}
Yunjey Choi, Minje Choi, Munyoung Kim, Jung-Woo Ha, Sunghun Kim, and Jaegul Choo. 2021.
\newblock Viton-hd: High-resolution virtual try-on via image translation.
\newblock \emph{Proceedings of the IEEE/CVF Conference on Computer Vision and Pattern Recognition (CVPR)}, pages 14131--14140.

\bibitem[{Cui et~al.(2024)Cui, Bi, Guo, and Cheng}]{cui2024moremultimodalretrievalaugmented}
Wanqing Cui, Keping Bi, Jiafeng Guo, and Xueqi Cheng. 2024.
\newblock \href {https://arxiv.org/abs/2402.13625} {More: Multi-modal retrieval augmented generative commonsense reasoning}.
\newblock \emph{Preprint}, arXiv:2402.13625.

\bibitem[{Dai et~al.(2023)Dai, Li, Li, Tiong, Zhao, Wang, Li, Fung, and Hoi}]{dai2023instructblipgeneralpurposevisionlanguagemodels}
Wenliang Dai, Junnan Li, Dongxu Li, Anthony Meng~Huat Tiong, Junqi Zhao, Weisheng Wang, Boyang Li, Pascale Fung, and Steven Hoi. 2023.
\newblock Instructblip: towards general-purpose vision-language models with instruction tuning.
\newblock In \emph{Proceedings of the 37th International Conference on Neural Information Processing Systems}, NIPS '23, Red Hook, NY, USA. Curran Associates Inc.

\bibitem[{Damen et~al.(2022)Damen, Doughty, Farinella, Furnari, Kazakos, Ma, Moltisanti, Munro, Perrett, Price, and Wray}]{dima2020rescaling}
Dima Damen, Hazel Doughty, Giovanni~Maria Farinella, Antonino Furnari, Evangelos Kazakos, Jian Ma, Davide Moltisanti, Jonathan Munro, Toby Perrett, Will Price, and Michael Wray. 2022.
\newblock \href {https://doi.org/10.1007/s11263-021-01531-2} {Rescaling egocentric vision: Collection, pipeline and challenges for epic-kitchens-100}.
\newblock \emph{Int. J. Comput. Vision}, 130(1):33–55.

\bibitem[{Dang(2024)}]{dang2024multi}
Quang-Vinh Dang. 2024.
\newblock Multi-modal retrieval augmented generation for product query.
\newblock \emph{Library of Progress-Library Science, Information Technology \& Computer}, 44(3).

\bibitem[{Das and Singh(2023)}]{10.1145/3586075}
Ringki Das and Thoudam~Doren Singh. 2023.
\newblock \href {https://doi.org/10.1145/3586075} {Multimodal sentiment analysis: A survey of methods, trends, and challenges}.
\newblock \emph{ACM Comput. Surv.}, 55(13s).

\bibitem[{Ding et~al.(2024{\natexlab{a}})Ding, Pang, Wei, Shen, and Cheng}]{ding2024retrieveneedsadaptiveretrieval}
Hanxing Ding, Liang Pang, Zihao Wei, Huawei Shen, and Xueqi Cheng. 2024{\natexlab{a}}.
\newblock \href {https://arxiv.org/abs/2402.10612} {Retrieve only when it needs: Adaptive retrieval augmentation for hallucination mitigation in large language models}.
\newblock \emph{Preprint}, arXiv:2402.10612.

\bibitem[{Ding et~al.(2024{\natexlab{b}})Ding, Ma, Qin, Wu, Li, and Nie}]{ding2024rablipmultimodaladaptiveretrievalaugmented}
Muhe Ding, Yang Ma, Pengda Qin, Jianlong Wu, Yuhong Li, and Liqiang Nie. 2024{\natexlab{b}}.
\newblock \href {https://arxiv.org/abs/2410.14154} {Ra-blip: Multimodal adaptive retrieval-augmented bootstrapping language-image pre-training}.
\newblock \emph{Preprint}, arXiv:2410.14154.

\bibitem[{Ding et~al.(2024{\natexlab{c}})Ding, Ren, Huang, Luo, and Han}]{ding2024mmvqa}
Yihao Ding, Kaixuan Ren, Jiabin Huang, Siwen Luo, and Soyeon~Caren Han. 2024{\natexlab{c}}.
\newblock Mmvqa: A comprehensive dataset for investigating multipage multimodal information retrieval in pdf-based visual question answering.
\newblock In \emph{Proceedings of the Thirty-Third International Joint Conference on Artificial Intelligence, IJCAI}, pages 3--9.

\bibitem[{Ding et~al.(2024{\natexlab{d}})Ding, Ren, Huang, Luo, and Han}]{ding2024pdfmvqadatasetmultimodalinformation}
Yihao Ding, Kaixuan Ren, Jiabin Huang, Siwen Luo, and Soyeon~Caren Han. 2024{\natexlab{d}}.
\newblock \href {https://arxiv.org/abs/2404.12720} {Pdf-mvqa: A dataset for multimodal information retrieval in pdf-based visual question answering}.
\newblock \emph{Preprint}, arXiv:2404.12720.

\bibitem[{Dong et~al.(2024{\natexlab{a}})Dong, Fatemi, Perozzi, Yang, and Tsitsulin}]{dong2024dontforgetconnectimproving}
Jialin Dong, Bahare Fatemi, Bryan Perozzi, Lin~F. Yang, and Anton Tsitsulin. 2024{\natexlab{a}}.
\newblock \href {https://arxiv.org/abs/2405.18414} {Don't forget to connect! improving rag with graph-based reranking}.
\newblock \emph{Preprint}, arXiv:2405.18414.

\bibitem[{Dong et~al.(2024{\natexlab{b}})Dong, Feng, Zhou, Yu, Yang, and Guo}]{10.1145/3626772.3657833}
Xingning Dong, Zipeng Feng, Chunluan Zhou, Xuzheng Yu, Ming Yang, and Qingpei Guo. 2024{\natexlab{b}}.
\newblock \href {https://doi.org/10.1145/3626772.3657833} {M2-raap: A multi-modal recipe for advancing adaptation-based pre-training towards effective and efficient zero-shot video-text retrieval}.
\newblock In \emph{Proceedings of the 47th International ACM SIGIR Conference on Research and Development in Information Retrieval}, SIGIR '24, page 2156–2166, New York, NY, USA. Association for Computing Machinery.

\bibitem[{Drossos et~al.(2020)Drossos, Lipping, and Virtanen}]{drossos2020clotho}
Konstantinos Drossos, Samuel Lipping, and Tuomas Virtanen. 2020.
\newblock Clotho: An audio captioning dataset.
\newblock In \emph{ICASSP 2020-2020 IEEE International Conference on Acoustics, Speech and Signal Processing (ICASSP)}, pages 736--740. IEEE.

\bibitem[{Du et~al.(2024)Du, Liu, and Jin}]{10.1145/3664647.3680731}
Yang Du, Yuqi Liu, and Qin Jin. 2024.
\newblock \href {https://doi.org/10.1145/3664647.3680731} {Reversed in time: A novel temporal-emphasized benchmark for cross-modal video-text retrieval}.
\newblock In \emph{Proceedings of the 32nd ACM International Conference on Multimedia}, MM '24, page 5260–5269, New York, NY, USA. Association for Computing Machinery.

\bibitem[{Elliott et~al.(2016)Elliott, Frank, Sima'an, and Specia}]{elliott2016multi30k}
Desmond Elliott, Stella Frank, Khalil Sima'an, and Lucia Specia. 2016.
\newblock Multi30k: Multilingual english-german image descriptions.
\newblock \emph{Proceedings of the 5th Workshop on Vision and Language}, pages 70--74.

\bibitem[{Fan et~al.(2019)Fan, Gardent, Braud, and Bordes}]{fan2019eli5}
Angela Fan, Claire Gardent, Chloé Braud, and Antoine Bordes. 2019.
\newblock Eli5: Long form question answering.
\newblock \emph{Proceedings of the 57th Annual Meeting of the Association for Computational Linguistics}, pages 3558--3567.

\bibitem[{Faysse et~al.(2024)Faysse, Sibille, Wu, Omrani, Viaud, Hudelot, and Colombo}]{faysse2024colpaliefficientdocumentretrieval}
Manuel Faysse, Hugues Sibille, Tony Wu, Bilel Omrani, Gautier Viaud, Céline Hudelot, and Pierre Colombo. 2024.
\newblock \href {https://arxiv.org/abs/2407.01449} {Colpali: Efficient document retrieval with vision language models}.
\newblock \emph{Preprint}, arXiv:2407.01449.

\bibitem[{Feng et~al.(2023)Feng, Bai, Luo, Li, Khan, Zuo, Xu, Goh, and Liu}]{feng2023vqa4cirboostingcomposedimage}
Chun-Mei Feng, Yang Bai, Tao Luo, Zhen Li, Salman Khan, Wangmeng Zuo, Xinxing Xu, Rick Siow~Mong Goh, and Yong Liu. 2023.
\newblock \href {https://arxiv.org/abs/2312.12273} {Vqa4cir: Boosting composed image retrieval with visual question answering}.
\newblock \emph{Preprint}, arXiv:2312.12273.

\bibitem[{Gao et~al.(2023)Gao, Xiong, Gao, Jia, Pan, Bi, Dai, Sun, Guo, Wang, and Wang}]{Gao2023RetrievalAugmentedGF}
Yunfan Gao, Yun Xiong, Xinyu Gao, Kangxiang Jia, Jinliu Pan, Yuxi Bi, Yi~Dai, Jiawei Sun, Qianyu Guo, Meng Wang, and Haofen Wang. 2023.
\newblock \href {https://api.semanticscholar.org/CorpusID:266359151} {Retrieval-augmented generation for large language models: A survey}.
\newblock \emph{ArXiv}, abs/2312.10997.

\bibitem[{Gemmeke et~al.(2017)Gemmeke, Ellis, Freedman, Jansen, Lawrence et~al.}]{gemmeke2017audioset}
Jort~F Gemmeke, Daniel~PW Ellis, Dylan Freedman, Aren Jansen, Wade Lawrence, et~al. 2017.
\newblock Audioset: An ontology and human-labeled dataset for audio events.
\newblock \emph{Proceedings of the IEEE International Conference on Acoustics, Speech and Signal Processing (ICASSP)}, pages 776--780.

\bibitem[{Ghosh et~al.(2024)Ghosh, Kumar, Reddy~Evuru, Duraiswami, and Manocha}]{10448030}
Sreyan Ghosh, Sonal Kumar, Chandra~Kiran Reddy~Evuru, Ramani Duraiswami, and Dinesh Manocha. 2024.
\newblock \href {https://doi.org/10.1109/ICASSP48485.2024.10448030} {Recap: Retrieval-augmented audio captioning}.
\newblock In \emph{ICASSP 2024 - 2024 IEEE International Conference on Acoustics, Speech and Signal Processing (ICASSP)}, pages 1161--1165.

\bibitem[{Giancola et~al.(2018)Giancola, Amine, Dghaily, and Ghanem}]{giancola2018soccernet}
Silvio Giancola, Mohieddine Amine, Tarek Dghaily, and Bernard Ghanem. 2018.
\newblock Soccernet: A scalable dataset for action spotting in soccer videos.
\newblock In \emph{Proceedings of the IEEE conference on computer vision and pattern recognition workshops}, pages 1711--1721.

\bibitem[{Goyal et~al.(2017{\natexlab{a}})Goyal, Khot, Summers{-}Stay, Batra, and Parikh}]{balanced_vqa_v2}
Yash Goyal, Tejas Khot, Douglas Summers{-}Stay, Dhruv Batra, and Devi Parikh. 2017{\natexlab{a}}.
\newblock Making the {V} in {VQA} matter: Elevating the role of image understanding in {V}isual {Q}uestion {A}nswering.
\newblock In \emph{Conference on Computer Vision and Pattern Recognition (CVPR)}.

\bibitem[{Goyal et~al.(2017{\natexlab{b}})Goyal, Khot, Summers-Stay, Batra, and Parikh}]{goyal2017making}
Yash Goyal, Tejas Khot, Douglas Summers-Stay, Dhruv Batra, and Devi Parikh. 2017{\natexlab{b}}.
\newblock Making the v in vqa matter: Elevating the role of image understanding in visual question answering.
\newblock In \emph{Proceedings of the IEEE conference on computer vision and pattern recognition}, pages 6904--6913.

\bibitem[{Grattafiori et~al.(2024)Grattafiori, Dubey, Jauhri, Pandey, Kadian, Al-Dahle, Letman, Mathur, Schelten, Vaughan, Yang, Fan, Goyal, Hartshorn, Yang, Mitra, Sravankumar, Korenev, Hinsvark, Rao, Zhang, Rodriguez, Gregerson, Spataru, Roziere, Biron, Tang, Chern, Caucheteux, Nayak, Bi, Marra, McConnell, Keller, Touret, Wu, Wong, Ferrer, Nikolaidis, Allonsius, Song, Pintz, Livshits, Wyatt, Esiobu, Choudhary, Mahajan, Garcia-Olano, Perino, Hupkes, Lakomkin, AlBadawy, Lobanova, Dinan, Smith, and et~al.}]{grattafiori2024llama3herdmodels}
Aaron Grattafiori, Abhimanyu Dubey, Abhinav Jauhri, Abhinav Pandey, Abhishek Kadian, Ahmad Al-Dahle, Aiesha Letman, Akhil Mathur, Alan Schelten, Alex Vaughan, Amy Yang, Angela Fan, Anirudh Goyal, Anthony Hartshorn, Aobo Yang, Archi Mitra, Archie Sravankumar, Artem Korenev, Arthur Hinsvark, Arun Rao, Aston Zhang, Aurelien Rodriguez, Austen Gregerson, Ava Spataru, Baptiste Roziere, Bethany Biron, Binh Tang, Bobbie Chern, Charlotte Caucheteux, Chaya Nayak, Chloe Bi, Chris Marra, Chris McConnell, Christian Keller, Christophe Touret, Chunyang Wu, Corinne Wong, Cristian~Canton Ferrer, Cyrus Nikolaidis, Damien Allonsius, Daniel Song, Danielle Pintz, Danny Livshits, Danny Wyatt, David Esiobu, Dhruv Choudhary, Dhruv Mahajan, Diego Garcia-Olano, Diego Perino, Dieuwke Hupkes, Egor Lakomkin, Ehab AlBadawy, Elina Lobanova, Emily Dinan, Eric~Michael Smith, and et~al. 2024.
\newblock \href {https://arxiv.org/abs/2407.21783} {The llama 3 herd of models}.
\newblock \emph{Preprint}, arXiv:2407.21783.

\bibitem[{Grauman et~al.(2022)Grauman, Westbury, Byrne, Chavis, Furnari, Girdhar, Hamburger, Jiang, Liu, Liu et~al.}]{grauman2022ego4d}
Kristen Grauman, Andrew Westbury, Eugene Byrne, Zachary Chavis, Antonino Furnari, Rohit Girdhar, Jackson Hamburger, Hao Jiang, Miao Liu, Xingyu Liu, et~al. 2022.
\newblock Ego4d: Around the world in 3,000 hours of egocentric video.
\newblock In \emph{Proceedings of the IEEE/CVF Conference on Computer Vision and Pattern Recognition}, pages 18995--19012.

\bibitem[{Guo et~al.(2020)Guo, Sun, Lindgren, Geng, Simcha, Chern, and Kumar}]{guo2020acceleratinglargescaleinferenceanisotropic}
Ruiqi Guo, Philip Sun, Erik Lindgren, Quan Geng, David Simcha, Felix Chern, and Sanjiv Kumar. 2020.
\newblock Accelerating large-scale inference with anisotropic vector quantization.
\newblock In \emph{International Conference on Machine Learning}, pages 3887--3896. PMLR.

\bibitem[{Ha et~al.(2025)Ha, Zhan, Kim, Bralios, Sanniboina, Peng, Chang, Kang, and Ji}]{ha2025mmpoisonragdisruptingmultimodalrag}
Hyeonjeong Ha, Qiusi Zhan, Jeonghwan Kim, Dimitrios Bralios, Saikrishna Sanniboina, Nanyun Peng, Kai-Wei Chang, Daniel Kang, and Heng Ji. 2025.
\newblock \href {https://arxiv.org/abs/2502.17832} {Mm-poisonrag: Disrupting multimodal rag with local and global poisoning attacks}.
\newblock \emph{Preprint}, arXiv:2502.17832.

\bibitem[{Haynes(2013)}]{Haynes2013}
Winston Haynes. 2013.
\newblock \href {https://doi.org/10.1007/978-1-4419-9863-7_1213} {\emph{Bonferroni Correction}}, pages 154--154.
\newblock Springer New York, New York, NY.

\bibitem[{He et~al.(2024)He, Li, Jang, Jia, Cao, Shah, Shrivastava, and Lim}]{he2024ma}
Bo~He, Hengduo Li, Young~Kyun Jang, Menglin Jia, Xuefei Cao, Ashish Shah, Abhinav Shrivastava, and Ser-Nam Lim. 2024.
\newblock Ma-lmm: Memory-augmented large multimodal model for long-term video understanding.
\newblock In \emph{Proceedings of the IEEE/CVF Conference on Computer Vision and Pattern Recognition}, pages 13504--13514.

\bibitem[{Hemker et~al.(2024)Hemker, Simidjievski, and Jamnik}]{hemker2024healnet}
Konstantin Hemker, Nikola Simidjievski, and Mateja Jamnik. 2024.
\newblock \href {https://openreview.net/forum?id=HUxtJcQpDS} {{HEALN}et: Multimodal fusion for heterogeneous biomedical data}.
\newblock In \emph{The Thirty-eighth Annual Conference on Neural Information Processing Systems}.

\bibitem[{Hessel et~al.(2021)Hessel, Holtzman, Forbes, Le~Bras, and Choi}]{hessel-etal-2021-clipscore}
Jack Hessel, Ari Holtzman, Maxwell Forbes, Ronan Le~Bras, and Yejin Choi. 2021.
\newblock \href {https://doi.org/10.18653/v1/2021.emnlp-main.595} {{CLIPS}core: A reference-free evaluation metric for image captioning}.
\newblock In \emph{Proceedings of the 2021 Conference on Empirical Methods in Natural Language Processing}, pages 7514--7528, Online and Punta Cana, Dominican Republic. Association for Computational Linguistics.

\bibitem[{Heusel et~al.(2017)Heusel, Ramsauer, Unterthiner, Nessler, and Hochreiter}]{heusel2017gans}
Martin Heusel, Hubert Ramsauer, Thomas Unterthiner, Bernhard Nessler, and Sepp Hochreiter. 2017.
\newblock Gans trained by a two time-scale update rule converge to a local nash equilibrium.
\newblock \emph{Advances in neural information processing systems}, 30.

\bibitem[{Hu et~al.(2024{\natexlab{a}})Hu, Xu, Ye, Yan, Zhang, Zhang, Zhang, Jin, Huang, and Zhou}]{hu-etal-2024-mplug}
Anwen Hu, Haiyang Xu, Jiabo Ye, Ming Yan, Liang Zhang, Bo~Zhang, Ji~Zhang, Qin Jin, Fei Huang, and Jingren Zhou. 2024{\natexlab{a}}.
\newblock \href {https://doi.org/10.18653/v1/2024.findings-emnlp.175} {m{PLUG}-{D}oc{O}wl 1.5: Unified structure learning for {OCR}-free document understanding}.
\newblock In \emph{Findings of the Association for Computational Linguistics: EMNLP 2024}, pages 3096--3120, Miami, Florida, USA. Association for Computational Linguistics.

\bibitem[{Hu et~al.(2024{\natexlab{b}})Hu, Xu, Zhang, Ye, Yan, Zhang, Jin, Huang, and Zhou}]{hu2024mplugdocowl2highresolutioncompressingocrfree}
Anwen Hu, Haiyang Xu, Liang Zhang, Jiabo Ye, Ming Yan, Ji~Zhang, Qin Jin, Fei Huang, and Jingren Zhou. 2024{\natexlab{b}}.
\newblock \href {https://arxiv.org/abs/2409.03420} {mplug-docowl2: High-resolution compressing for ocr-free multi-page document understanding}.
\newblock \emph{Preprint}, arXiv:2409.03420.

\bibitem[{Hu et~al.(2024{\natexlab{c}})Hu, Gu, Dou, Fayyaz, Lu, Chang, and Peng}]{hu2024mragbench}
Wenbo Hu, Jia-Chen Gu, Zi-Yi Dou, Mohsen Fayyaz, Pan Lu, Kai-Wei Chang, and Nanyun Peng. 2024{\natexlab{c}}.
\newblock Mrag-bench: Vision-centric evaluation for retrieval-augmented multimodal models.
\newblock \emph{arXiv preprint arXiv:2410.08182}.

\bibitem[{Hu et~al.(2023)Hu, Iscen, Sun, Wang, Chang, Sun, Schmid, Ross, and Fathi}]{Hu_2023_CVPR}
Ziniu Hu, Ahmet Iscen, Chen Sun, Zirui Wang, Kai-Wei Chang, Yizhou Sun, Cordelia Schmid, David~A. Ross, and Alireza Fathi. 2023.
\newblock Reveal: Retrieval-augmented visual-language pre-training with multi-source multimodal knowledge memory.
\newblock In \emph{Proceedings of the IEEE/CVF Conference on Computer Vision and Pattern Recognition (CVPR)}, pages 23369--23379.

\bibitem[{Huang et~al.(2024)Huang, Yu, Ma, Zhong, Feng, Wang, Chen, Peng, Feng, Qin, and Liu}]{10.1145/3703155}
Lei Huang, Weijiang Yu, Weitao Ma, Weihong Zhong, Zhangyin Feng, Haotian Wang, Qianglong Chen, Weihua Peng, Xiaocheng Feng, Bing Qin, and Ting Liu. 2024.
\newblock \href {https://doi.org/10.1145/3703155} {A survey on hallucination in large language models: Principles, taxonomy, challenges, and open questions}.
\newblock \emph{ACM Trans. Inf. Syst.}
\newblock Just Accepted.

\bibitem[{Huang et~al.(2025)Huang, Zhang, Zhang, Zhou, and Wang}]{huang2025ru}
Liting Huang, Zhihao Zhang, Yiran Zhang, Xiyue Zhou, and Shoujin Wang. 2025.
\newblock Ru-ai: A large multimodal dataset for machine-generated content detection.
\newblock In \emph{Companion Proceedings of the ACM on Web Conference 2025}, pages 733--736.

\bibitem[{Ikezogwo et~al.(2023)Ikezogwo, Seyfioglu, Ghezloo, Geva, Mohammed, Anand, Krishna, and Shapiro}]{ikezogwo2023quilt}
Wisdom~Oluchi Ikezogwo, Mehmet~Saygin Seyfioglu, Fatemeh Ghezloo, Dylan Stefan~Chan Geva, Fatwir~Sheikh Mohammed, Pavan~Kumar Anand, Ranjay Krishna, and Linda Shapiro. 2023.
\newblock Quilt-1m: One million image-text pairs for histopathology.
\newblock \emph{arXiv preprint arXiv:2306.11207}.

\bibitem[{Irvin et~al.(2019)Irvin, Rajpurkar, Ko, Yu, Ciurea-Ilcus, Chute, Marklund, Haghgoo, Ball, Shpanskaya et~al.}]{irvin2019chexpert}
Jeremy Irvin, Pranav Rajpurkar, Michael Ko, Yifan Yu, Silviana Ciurea-Ilcus, Chris Chute, Henrik Marklund, Behzad Haghgoo, Robyn Ball, Katie Shpanskaya, et~al. 2019.
\newblock Chexpert: A large chest radiograph dataset with uncertainty labels and expert comparison.
\newblock In \emph{Proceedings of the AAAI conference on artificial intelligence}, volume~33, pages 590--597.

\bibitem[{Izacard et~al.(2022)Izacard, Caron, Hosseini, Riedel, Bojanowski, Joulin, and Grave}]{izacard2022unsuperviseddenseinformationretrieval}
Gautier Izacard, Mathilde Caron, Lucas Hosseini, Sebastian Riedel, Piotr Bojanowski, Armand Joulin, and Edouard Grave. 2022.
\newblock \href {https://arxiv.org/abs/2112.09118} {Unsupervised dense information retrieval with contrastive learning}.
\newblock \emph{Preprint}, arXiv:2112.09118.

\bibitem[{Jang et~al.(2024)Jang, Kim, Meng, Huynh, and Lim}]{jang2024visual}
Young~Kyun Jang, Donghyun Kim, Zihang Meng, Dat Huynh, and Ser-Nam Lim. 2024.
\newblock Visual delta generator with large multi-modal models for semi-supervised composed image retrieval.
\newblock In \emph{Proceedings of the IEEE/CVF Conference on Computer Vision and Pattern Recognition}, pages 16805--16814.

\bibitem[{Jeong et~al.(2025)Jeong, Kim, Baek, and Hwang}]{jeong2025videoragretrievalaugmentedgenerationvideo}
Soyeong Jeong, Kangsan Kim, Jinheon Baek, and Sung~Ju Hwang. 2025.
\newblock \href {https://arxiv.org/abs/2501.05874} {Videorag: Retrieval-augmented generation over video corpus}.
\newblock \emph{Preprint}, arXiv:2501.05874.

\bibitem[{Jia et~al.(2021)Jia, Yang, Xia, Chen, Parekh, Pham, Le, Sung, Li, and Duerig}]{pmlr-v139-jia21b}
Chao Jia, Yinfei Yang, Ye~Xia, Yi-Ting Chen, Zarana Parekh, Hieu Pham, Quoc Le, Yun-Hsuan Sung, Zhen Li, and Tom Duerig. 2021.
\newblock \href {https://proceedings.mlr.press/v139/jia21b.html} {Scaling up visual and vision-language representation learning with noisy text supervision}.
\newblock In \emph{Proceedings of the 38th International Conference on Machine Learning}, volume 139 of \emph{Proceedings of Machine Learning Research}, pages 4904--4916. PMLR.

\bibitem[{Jia et~al.(2020)Jia, Shi, Sirotenko, Cui, Cardie, Hariharan, Adam, and Belongie}]{fashionpedia}
Menglin Jia, Mengyun Shi, Mikhail Sirotenko, Yin Cui, Claire Cardie, Bharath Hariharan, Hartwig Adam, and Serge Belongie. 2020.
\newblock \href {https://arxiv.org/abs/2004.12276} {Fashionpedia: Ontology, segmentation, and an attribute localization dataset}.
\newblock \emph{Preprint}, arXiv:2004.12276.

\bibitem[{Jian et~al.(2024)Jian, Yu, and Zhang}]{jian-etal-2024-large}
Pu~Jian, Donglei Yu, and Jiajun Zhang. 2024.
\newblock \href {https://doi.org/10.18653/v1/2024.emnlp-main.613} {Large language models know what is key visual entity: An {LLM}-assisted multimodal retrieval for {VQA}}.
\newblock In \emph{Proceedings of the 2024 Conference on Empirical Methods in Natural Language Processing}, pages 10939--10956, Miami, Florida, USA. Association for Computational Linguistics.

\bibitem[{Jiang et~al.(2024)Jiang, Xu, Dong, Chen, Ye, Yan, Ye, Zhang, Huang, and Zhang}]{jiang2024hallucination}
Chaoya Jiang, Haiyang Xu, Mengfan Dong, Jiaxing Chen, Wei Ye, Ming Yan, Qinghao Ye, Ji~Zhang, Fei Huang, and Shikun Zhang. 2024.
\newblock Hallucination augmented contrastive learning for multimodal large language model.
\newblock In \emph{Proceedings of the IEEE/CVF Conference on Computer Vision and Pattern Recognition}, pages 27036--27046.

\bibitem[{Jiang et~al.(2025)Jiang, Li, Deng, Liu, Gao, Zhou, Li, Wang, and Zheng}]{jiang2025mmadcomprehensivebenchmarkmultimodal}
Xi~Jiang, Jian Li, Hanqiu Deng, Yong Liu, Bin-Bin Gao, Yifeng Zhou, Jialin Li, Chengjie Wang, and Feng Zheng. 2025.
\newblock \href {https://arxiv.org/abs/2410.09453} {Mmad: A comprehensive benchmark for multimodal large language models in industrial anomaly detection}.
\newblock \emph{Preprint}, arXiv:2410.09453.

\bibitem[{Jin et~al.(2024)Jin, Zhang, Gong, Xu, Deng, Wang, Zhang, Shen, and Feng}]{jin2024mv}
Xiaojie Jin, Bowen Zhang, Weibo Gong, Kai Xu, Xueqing Deng, Peng Wang, Zhao Zhang, Xiaohui Shen, and Jiashi Feng. 2024.
\newblock Mv-adapter: Multimodal video transfer learning for video text retrieval.
\newblock In \emph{Proceedings of the IEEE/CVF Conference on Computer Vision and Pattern Recognition}, pages 27144--27153.

\bibitem[{Johnson et~al.(2019)Johnson, Pollard, Greenbaum, Lungren, Deng, Peng et~al.}]{johnson2019mimic}
Alistair~EW Johnson, Tom~J Pollard, Nathaniel~R Greenbaum, Matthew~P Lungren, Chih-ying Deng, Yifan Peng, et~al. 2019.
\newblock Mimic-cxr-jpg, a large publicly available database of labeled chest radiographs.
\newblock \emph{arXiv preprint arXiv:1901.07042}.

\bibitem[{Johnson et~al.(2016)Johnson, Pollard, Shen, Li-wei, Feng et~al.}]{johnson2016mimic}
Alistair~EW Johnson, Tom~J Pollard, Lu~Shen, H~Lehman Li-wei, Mengling Feng, et~al. 2016.
\newblock Mimic-iii, a freely accessible critical care database.
\newblock \emph{Scientific Data}, 3:160035.

\bibitem[{Joshi et~al.(2017)Joshi, Choi, Weld, and Zettlemoyer}]{joshi2017triviaqa}
Mandar Joshi, Eunsol Choi, Daniel Weld, and Luke Zettlemoyer. 2017.
\newblock \href {https://doi.org/10.18653/v1/P17-1147} {{T}rivia{QA}: A large scale distantly supervised challenge dataset for reading comprehension}.
\newblock In \emph{Proceedings of the 55th Annual Meeting of the Association for Computational Linguistics (Volume 1: Long Papers)}, pages 1601--1611, Vancouver, Canada. Association for Computational Linguistics.

\bibitem[{Joshi et~al.(2024)Joshi, Gupta, Kumar, and Sisodia}]{joshi2024robust}
Pankaj Joshi, Aditya Gupta, Pankaj Kumar, and Manas Sisodia. 2024.
\newblock Robust multi model rag pipeline for documents containing text, table \& images.
\newblock In \emph{2024 3rd International Conference on Applied Artificial Intelligence and Computing (ICAAIC)}, pages 993--999. IEEE.

\bibitem[{Kafle et~al.(2018)Kafle, Price, Cohen, and Kanan}]{kafle2018dvqa}
Kushal Kafle, Brian Price, Scott Cohen, and Christopher Kanan. 2018.
\newblock Dvqa: Understanding data visualizations via question answering.
\newblock In \emph{Proceedings of the IEEE conference on computer vision and pattern recognition}, pages 5648--5656.

\bibitem[{Kandhare and Gisselbrecht(2024)}]{kandhare2024empiricalcomparisonvideoframe}
Mahesh Kandhare and Thibault Gisselbrecht. 2024.
\newblock \href {https://arxiv.org/abs/2408.03340} {An empirical comparison of video frame sampling methods for multi-modal rag retrieval}.
\newblock \emph{Preprint}, arXiv:2408.03340.

\bibitem[{Khalafaoui et~al.(2024)Khalafaoui, Lovisetto, Matei, and Grozavu}]{khalafaoui2024cadmrcrossattentiondisentangledlearning}
Yasser Khalafaoui, Martino Lovisetto, Basarab Matei, and Nistor Grozavu. 2024.
\newblock \href {https://arxiv.org/abs/2412.02295} {Cadmr: Cross-attention and disentangled learning for multimodal recommender systems}.
\newblock \emph{Preprint}, arXiv:2412.02295.

\bibitem[{Khaliq et~al.(2024)Khaliq, Chang, Ma, Pflugfelder, and Mileti{\'c}}]{khaliq-etal-2024-ragar}
Mohammed~Abdul Khaliq, Paul Yu-Chun Chang, Mingyang Ma, Bernhard Pflugfelder, and Filip Mileti{\'c}. 2024.
\newblock \href {https://doi.org/10.18653/v1/2024.fever-1.29} {{RAGAR}, your falsehood radar: {RAG}-augmented reasoning for political fact-checking using multimodal large language models}.
\newblock In \emph{Proceedings of the Seventh Fact Extraction and VERification Workshop (FEVER)}, pages 280--296, Miami, Florida, USA. Association for Computational Linguistics.

\bibitem[{Khan et~al.(2025)Khan, Chen, Mohamed, Feng, and Elhoseiny}]{khan2025vr}
Faizan~Farooq Khan, Jun Chen, Youssef Mohamed, Chun-Mei Feng, and Mohamed Elhoseiny. 2025.
\newblock Vr-rag: Open-vocabulary species recognition with rag-assisted large multi-modal models.
\newblock \emph{arXiv preprint arXiv:2505.05635}.

\bibitem[{Khattab and Zaharia(2020)}]{10.1145/3397271.3401075}
Omar Khattab and Matei Zaharia. 2020.
\newblock \href {https://doi.org/10.1145/3397271.3401075} {Colbert: Efficient and effective passage search via contextualized late interaction over bert}.
\newblock In \emph{Proceedings of the 43rd International ACM SIGIR Conference on Research and Development in Information Retrieval}, SIGIR '20, page 39–48, New York, NY, USA. Association for Computing Machinery.

\bibitem[{Kilgour et~al.(2019)Kilgour, Zuluaga, Roblek, and Sharifi}]{kilgour2019frechetaudiodistancemetric}
Kevin Kilgour, Mauricio Zuluaga, Dominik Roblek, and Matthew Sharifi. 2019.
\newblock \href {https://arxiv.org/abs/1812.08466} {Fr\'echet audio distance: A metric for evaluating music enhancement algorithms}.
\newblock \emph{Preprint}, arXiv:1812.08466.

\bibitem[{Kim et~al.(2019)Kim, Kim, Lee, and Kim}]{kim-etal-2019-audiocaps}
Chris~Dongjoo Kim, Byeongchang Kim, Hyunmin Lee, and Gunhee Kim. 2019.
\newblock \href {https://doi.org/10.18653/v1/N19-1011} {Audiocaps: Generating captions for audios in the wild}.
\newblock In \emph{Proceedings of the 2019 Conference of the North American Chapter of the Association for Computational Linguistics: Human Language Technologies, Volume 1 (Long and Short Papers)}, pages 119--132, Minneapolis, Minnesota. Association for Computational Linguistics.

\bibitem[{Kim et~al.(2018)Kim, Rohrbach, Darrell, Canny, and Akata}]{bddx}
Jinkyu Kim, Anna Rohrbach, Trevor Darrell, John Canny, and Zeynep Akata. 2018.
\newblock Textual explanations for self-driving vehicles.
\newblock \emph{Proceedings of the European Conference on Computer Vision (ECCV)}.

\bibitem[{Kim et~al.(2024)Kim, Kim, Moon, Choi, and Kim}]{kim2024you}
Minkuk Kim, Hyeon~Bae Kim, Jinyoung Moon, Jinwoo Choi, and Seong~Tae Kim. 2024.
\newblock Do you remember? dense video captioning with cross-modal memory retrieval.
\newblock In \emph{Proceedings of the IEEE/CVF Conference on Computer Vision and Pattern Recognition}, pages 13894--13904.

\bibitem[{Krishna et~al.(2017)Krishna, Hata, Ren, Fei-Fei, and Carlos~Niebles}]{Krishna_2017_ICCV}
Ranjay Krishna, Kenji Hata, Frederic Ren, Li~Fei-Fei, and Juan Carlos~Niebles. 2017.
\newblock Dense-captioning events in videos.
\newblock In \emph{Proceedings of the IEEE International Conference on Computer Vision (ICCV)}.

\bibitem[{Kumar and Marttinen(2024)}]{kumar2024improvingmedicalmultimodalcontrastive}
Yogesh Kumar and Pekka Marttinen. 2024.
\newblock \href {https://doi.org/10.1007/978-3-031-72661-3_27} {Improving medical multi-modal contrastive learning with expert annotations}.
\newblock In \emph{Computer Vision – ECCV 2024: 18th European Conference, Milan, Italy, September 29–October 4, 2024, Proceedings, Part XX}, page 468–486, Berlin, Heidelberg. Springer-Verlag.

\bibitem[{Kwiatkowski et~al.(2019)Kwiatkowski, Palomaki, Redfield, Collins, Parikh, Alberti, Epstein, Polosukhin, Devlin, Lee et~al.}]{kwiatkowski2019natural}
Tom Kwiatkowski, Jennimaria Palomaki, Olivia Redfield, Michael Collins, Ankur Parikh, Chris Alberti, Danielle Epstein, Illia Polosukhin, Jacob Devlin, Kenton Lee, et~al. 2019.
\newblock Natural questions: a benchmark for question answering research.
\newblock \emph{Transactions of the Association for Computational Linguistics}, 7:453--466.

\bibitem[{Lahiri and Hu(2024)}]{lahiri2024alzheimerragmultimodalretrievalaugmented}
Aritra~Kumar Lahiri and Qinmin~Vivian Hu. 2024.
\newblock \href {https://arxiv.org/abs/2412.16701} {Alzheimerrag: Multimodal retrieval augmented generation for pubmed articles}.
\newblock \emph{Preprint}, arXiv:2412.16701.

\bibitem[{Lan et~al.(2025)Lan, Niu, Meng, Zhou, and Su}]{lan2025llave}
Zhibin Lan, Liqiang Niu, Fandong Meng, Jie Zhou, and Jinsong Su. 2025.
\newblock Llave: Large language and vision embedding models with hardness-weighted contrastive learning.
\newblock \emph{arXiv preprint arXiv:2503.04812}.

\bibitem[{Lang et~al.(2025)Lang, Cheng, Zhong, and Zhou}]{lang2025retrieval}
Jian Lang, Zhangtao Cheng, Ting Zhong, and Fan Zhou. 2025.
\newblock Retrieval-augmented dynamic prompt tuning for incomplete multimodal learning.
\newblock \emph{arXiv preprint arXiv:2501.01120}.

\bibitem[{Lee et~al.(2024)Lee, An, and Kim}]{lee-etal-2024-planrag}
Myeonghwa Lee, Seonho An, and Min-Soo Kim. 2024.
\newblock \href {https://doi.org/10.18653/v1/2024.naacl-long.364} {{P}lan{RAG}: A plan-then-retrieval augmented generation for generative large language models as decision makers}.
\newblock In \emph{Proceedings of the 2024 Conference of the North American Chapter of the Association for Computational Linguistics: Human Language Technologies (Volume 1: Long Papers)}, pages 6537--6555, Mexico City, Mexico. Association for Computational Linguistics.

\bibitem[{Lerner et~al.(2022)Lerner, Ferret, Guinaudeau, Le~Borgne, Besançon, Moreno, and Lovón~Melgarejo}]{viquae}
Paul Lerner, Olivier Ferret, Camille Guinaudeau, Hervé Le~Borgne, Romaric Besançon, Jose~G Moreno, and Jesús Lovón~Melgarejo. 2022.
\newblock \href {https://doi.org/10.1145/3477495.3531753} {{ViQuAE}, a dataset for knowledge-based visual question answering about named entities}.
\newblock In \emph{Proceedings of The 45th International ACM SIGIR Conference on Research and Development in Information Retrieval}, SIGIR’22, New York, NY, USA. Association for Computing Machinery.

\bibitem[{Lewis et~al.(2020)Lewis, Perez, Piktus, Petroni, Karpukhin, Goyal, K{\"u}ttler, Lewis, Yih, Rockt{\"a}schel et~al.}]{lewis2020retrieval}
Patrick Lewis, Ethan Perez, Aleksandra Piktus, Fabio Petroni, Vladimir Karpukhin, Naman Goyal, Heinrich K{\"u}ttler, Mike Lewis, Wen-tau Yih, Tim Rockt{\"a}schel, et~al. 2020.
\newblock Retrieval-augmented generation for knowledge-intensive nlp tasks.
\newblock \emph{Advances in Neural Information Processing Systems}, 33:9459--9474.

\bibitem[{Lewis et~al.(2021)Lewis, Wu, Liu, Minervini, K{\"u}ttler, Piktus, Stenetorp, and Riedel}]{lewis2021paq}
Patrick Lewis, Yuxiang Wu, Linqing Liu, Pasquale Minervini, Heinrich K{\"u}ttler, Aleksandra Piktus, Pontus Stenetorp, and Sebastian Riedel. 2021.
\newblock Paq: 65 million probably-asked questions and what you can do with them.
\newblock \emph{Transactions of the Association for Computational Linguistics}, 9:1098--1115.

\bibitem[{Li et~al.(2025{\natexlab{a}})Li, Zhang, Chen, Wang, Pu, Cahyono, Yang, Li, and Liu}]{li2025otter}
Bo~Li, Yuanhan Zhang, Liangyu Chen, Jinghao Wang, Fanyi Pu, Joshua~Adrian Cahyono, Jingkang Yang, Chunyuan Li, and Ziwei Liu. 2025{\natexlab{a}}.
\newblock Otter: A multi-modal model with in-context instruction tuning.
\newblock \emph{IEEE Transactions on Pattern Analysis and Machine Intelligence}.

\bibitem[{Li et~al.(2023{\natexlab{a}})Li, Wang, Wang, Ge, Ge, and Shan}]{li2023seedbenchbenchmarkingmultimodalllms}
Bohao Li, Rui Wang, Guangzhi Wang, Yuying Ge, Yixiao Ge, and Ying Shan. 2023{\natexlab{a}}.
\newblock \href {https://arxiv.org/abs/2307.16125} {Seed-bench: Benchmarking multimodal llms with generative comprehension}.
\newblock \emph{Preprint}, arXiv:2307.16125.

\bibitem[{Li et~al.(2018)Li, Wu, Liu, and Lee}]{li2018spoken}
Chia-Hsuan Li, Szu-Lin Wu, Chi-Liang Liu, and Hung-yi Lee. 2018.
\newblock Spoken squad: A study of mitigating the impact of speech recognition errors on listening comprehension.
\newblock \emph{arXiv preprint arXiv:1804.00320}.

\bibitem[{Li et~al.(2023{\natexlab{b}})Li, Li, Savarese, and Hoi}]{10.5555/3618408.3619222}
Junnan Li, Dongxu Li, Silvio Savarese, and Steven Hoi. 2023{\natexlab{b}}.
\newblock Blip-2: bootstrapping language-image pre-training with frozen image encoders and large language models.
\newblock In \emph{Proceedings of the 40th International Conference on Machine Learning}, ICML'23. JMLR.org.

\bibitem[{Li et~al.(2022)Li, Li, Xiong, and Hoi}]{li2022blip}
Junnan Li, Dongxu Li, Caiming Xiong, and Steven Hoi. 2022.
\newblock Blip: Bootstrapping language-image pre-training for unified vision-language understanding and generation.
\newblock In \emph{International conference on machine learning}, pages 12888--12900. PMLR.

\bibitem[{Li et~al.(2024{\natexlab{a}})Li, Zhang, Dong, Xie, and Qin}]{li2024adaptivedatasetquantization}
Muquan Li, Dongyang Zhang, Qiang Dong, Xiurui Xie, and Ke~Qin. 2024{\natexlab{a}}.
\newblock \href {https://arxiv.org/abs/2412.16895} {Adaptive dataset quantization}.
\newblock \emph{Preprint}, arXiv:2412.16895.

\bibitem[{Li et~al.(2025{\natexlab{b}})Li, Chen, Wang, Wang, Ye, Jin, Chen, He, Gao, Cui et~al.}]{li2025omnicorpus}
Qingyun Li, Zhe Chen, Weiyun Wang, Wenhai Wang, Shenglong Ye, Zhenjiang Jin, Guanzhou Chen, Yinan He, Zhangwei Gao, Erfei Cui, et~al. 2025{\natexlab{b}}.
\newblock Omnicorpus: A unified multimodal corpus of 10 billion-level images interleaved with text.
\newblock In \emph{ICLR}.

\bibitem[{Li et~al.(2024{\natexlab{b}})Li, Shang, Wei, Guo, Li, He, Zhang, and Yang}]{li2024laragenhancingllmbasedasraccuracy}
Shaojun Li, Hengchao Shang, Daimeng Wei, Jiaxin Guo, Zongyao Li, Xianghui He, Min Zhang, and Hao Yang. 2024{\natexlab{b}}.
\newblock \href {https://arxiv.org/abs/2409.08597} {La-rag:enhancing llm-based asr accuracy with retrieval-augmented generation}.
\newblock \emph{Preprint}, arXiv:2409.08597.

\bibitem[{Li et~al.(2025{\natexlab{c}})Li, Chen, Ma, Xu, Liang, Zheng, Kong, and Chen}]{10890325}
Xiquan Li, Wenxi Chen, Ziyang Ma, Xuenan Xu, Yuzhe Liang, Zhisheng Zheng, Qiuqiang Kong, and Xie Chen. 2025{\natexlab{c}}.
\newblock \href {https://doi.org/10.1109/ICASSP49660.2025.10890325} {Drcap: Decoding clap latents with retrieval-augmented generation for zero-shot audio captioning}.
\newblock In \emph{ICASSP 2025 - 2025 IEEE International Conference on Acoustics, Speech and Signal Processing (ICASSP)}, pages 1--5.

\bibitem[{Li et~al.(2019)Li, Xu, Wang, Lan, Jia, Yang, and Xu}]{cococn}
Xirong Li, Chaoxi Xu, Xiaoxu Wang, Weiyu Lan, Zhengxiong Jia, Gang Yang, and Jieping Xu. 2019.
\newblock \href {https://arxiv.org/abs/1805.08661} {Coco-cn for cross-lingual image tagging, captioning and retrieval}.

\bibitem[{Li et~al.(2024{\natexlab{c}})Li, Zhang, Yang, Fu, Cheng, Chen, Chen, and Wei}]{li2024appagentv2advancedagent}
Yanda Li, Chi Zhang, Wanqi Yang, Bin Fu, Pei Cheng, Xin Chen, Ling Chen, and Yunchao Wei. 2024{\natexlab{c}}.
\newblock \href {https://arxiv.org/abs/2408.11824} {Appagent v2: Advanced agent for flexible mobile interactions}.
\newblock \emph{Preprint}, arXiv:2408.11824.

\bibitem[{Li et~al.(2024{\natexlab{d}})Li, Li, Wang, Jiang, Zhang, Zheng, Wang, Zheng, Yu, Huang et~al.}]{li2024benchmarking}
Yangning Li, Yinghui Li, Xingyu Wang, Yong Jiang, Zhen Zhang, Xinran Zheng, Hui Wang, Hai-Tao Zheng, Philip~S Yu, Fei Huang, et~al. 2024{\natexlab{d}}.
\newblock Benchmarking multimodal retrieval augmented generation with dynamic vqa dataset and self-adaptive planning agent.
\newblock \emph{arXiv preprint arXiv:2411.02937}.

\bibitem[{Li et~al.(2023{\natexlab{c}})Li, Zhang, Zhang, Long, Xie, and Zhang}]{li2023generaltextembeddingsmultistage}
Zehan Li, Xin Zhang, Yanzhao Zhang, Dingkun Long, Pengjun Xie, and Meishan Zhang. 2023{\natexlab{c}}.
\newblock \href {https://arxiv.org/abs/2308.03281} {Towards general text embeddings with multi-stage contrastive learning}.
\newblock \emph{Preprint}, arXiv:2308.03281.

\bibitem[{Lin(2004)}]{lin-2004-rouge}
Chin-Yew Lin. 2004.
\newblock \href {https://aclanthology.org/W04-1013/} {{ROUGE}: A package for automatic evaluation of summaries}.
\newblock In \emph{Text Summarization Branches Out}, pages 74--81, Barcelona, Spain. Association for Computational Linguistics.

\bibitem[{Lin et~al.(2024{\natexlab{a}})Lin, Lee, Shoeybi, Lin, Catanzaro, and Ping}]{lin2024mmembeduniversalmultimodalretrieval}
Sheng-Chieh Lin, Chankyu Lee, Mohammad Shoeybi, Jimmy Lin, Bryan Catanzaro, and Wei Ping. 2024{\natexlab{a}}.
\newblock \href {https://arxiv.org/abs/2411.02571} {Mm-embed: Universal multimodal retrieval with multimodal llms}.
\newblock \emph{Preprint}, arXiv:2411.02571.

\bibitem[{Lin et~al.(2014)Lin, Maire, Belongie, Hays, Perona, Ramanan, Doll{\'a}r, and Zitnick}]{lin2014microsoft}
Tsung-Yi Lin, Michael Maire, Serge Belongie, James Hays, Pietro Perona, Deva Ramanan, Piotr Doll{\'a}r, and C~Lawrence Zitnick. 2014.
\newblock Microsoft coco: Common objects in context.
\newblock \emph{European Conference on Computer Vision}, pages 740--755.

\bibitem[{Lin et~al.(2024{\natexlab{b}})Lin, Mei, Chen, and Byrne}]{lin-etal-2024-preflmr}
Weizhe Lin, Jingbiao Mei, Jinghong Chen, and Bill Byrne. 2024{\natexlab{b}}.
\newblock \href {https://doi.org/10.18653/v1/2024.acl-long.289} {{P}re{FLMR}: Scaling up fine-grained late-interaction multi-modal retrievers}.
\newblock In \emph{Proceedings of the 62nd Annual Meeting of the Association for Computational Linguistics (Volume 1: Long Papers)}, pages 5294--5316, Bangkok, Thailand. Association for Computational Linguistics.

\bibitem[{Liu et~al.(2024{\natexlab{a}})Liu, Chen, Lu, Jiang, Han, Zhang, Chen, Zhang, Ding, Zhang, Chen, Yang, Yang, and Qiu}]{liu2024retrievalattentionacceleratinglongcontextllm}
Di~Liu, Meng Chen, Baotong Lu, Huiqiang Jiang, Zhenhua Han, Qianxi Zhang, Qi~Chen, Chengruidong Zhang, Bailu Ding, Kai Zhang, Chen Chen, Fan Yang, Yuqing Yang, and Lili Qiu. 2024{\natexlab{a}}.
\newblock \href {https://arxiv.org/abs/2409.10516} {Retrievalattention: Accelerating long-context llm inference via vector retrieval}.
\newblock \emph{Preprint}, arXiv:2409.10516.

\bibitem[{Liu et~al.(2023{\natexlab{a}})Liu, Li, Wu, and Lee}]{liu2023llava}
Haotian Liu, Chunyuan Li, Qingyang Wu, and Yong~Jae Lee. 2023{\natexlab{a}}.
\newblock Visual instruction tuning.
\newblock \emph{Advances in neural information processing systems}, 36:34892--34916.

\bibitem[{Liu et~al.(2025{\natexlab{a}})Liu, Liu, Yao, Liu, Meng, Wang, and Ma}]{liu2025hmraghierarchicalmultiagentmultimodal}
Pei Liu, Xin Liu, Ruoyu Yao, Junming Liu, Siyuan Meng, Ding Wang, and Jun Ma. 2025{\natexlab{a}}.
\newblock \href {https://arxiv.org/abs/2504.12330} {Hm-rag: Hierarchical multi-agent multimodal retrieval augmented generation}.
\newblock \emph{Preprint}, arXiv:2504.12330.

\bibitem[{Liu et~al.(2017)Liu, Zhu, Ye, Guadarrama, and Murphy}]{liu2017improved}
Siqi Liu, Zhenhai Zhu, Ning Ye, Sergio Guadarrama, and Kevin Murphy. 2017.
\newblock Improved image captioning via policy gradient optimization of spider.
\newblock In \emph{Proceedings of the IEEE international conference on computer vision}, pages 873--881.

\bibitem[{Liu et~al.(2024{\natexlab{b}})Liu, Wang, Deng, Peng, Liu, Nong, Williams, and Li}]{Liu_2024}
Xingzu Liu, Mingbang Wang, Songhang Deng, Xinyue Peng, Yanming Liu, Ruilin Nong, David Williams, and Jiyuan Li. 2024{\natexlab{b}}.
\newblock \href {https://doi.org/10.36227/techrxiv.172840252.24352951/v1} {Iterative retrieval augmentation for multi-modal knowledge integration and generation}.

\bibitem[{Liu et~al.(2024{\natexlab{c}})Liu, Peng, Zhang, Liu, Yin, Cao, and Du}]{liu-etal-2024-ra}
Yanming Liu, Xinyue Peng, Xuhong Zhang, Weihao Liu, Jianwei Yin, Jiannan Cao, and Tianyu Du. 2024{\natexlab{c}}.
\newblock \href {https://doi.org/10.18653/v1/2024.findings-acl.281} {{RA}-{ISF}: Learning to answer and understand from retrieval augmentation via iterative self-feedback}.
\newblock In \emph{Findings of the Association for Computational Linguistics: ACL 2024}, pages 4730--4749, Bangkok, Thailand. Association for Computational Linguistics.

\bibitem[{Liu et~al.(2025{\natexlab{b}})Liu, Yuan, Tie, Shi, Zhou, Sun, and Gong}]{liu2025poisoned}
Yinuo Liu, Zenghui Yuan, Guiyao Tie, Jiawen Shi, Pan Zhou, Lichao Sun, and Neil~Zhenqiang Gong. 2025{\natexlab{b}}.
\newblock Poisoned-mrag: Knowledge poisoning attacks to multimodal retrieval augmented generation.
\newblock \emph{arXiv preprint arXiv:2503.06254}.

\bibitem[{Liu et~al.(2025{\natexlab{c}})Liu, Duan, Zhang, Li, Zhang, Zhao, Yuan, Wang, He, Liu et~al.}]{liu2025mmbench}
Yuan Liu, Haodong Duan, Yuanhan Zhang, Bo~Li, Songyang Zhang, Wangbo Zhao, Yike Yuan, Jiaqi Wang, Conghui He, Ziwei Liu, et~al. 2025{\natexlab{c}}.
\newblock Mmbench: Is your multi-modal model an all-around player?
\newblock In \emph{European conference on computer vision}, pages 216--233. Springer.

\bibitem[{Liu et~al.(2023{\natexlab{b}})Liu, Xiong, Lv, Liu, and Yu}]{liu2023universal}
Zhenghao Liu, Chenyan Xiong, Yuanhuiyi Lv, Zhiyuan Liu, and Ge~Yu. 2023{\natexlab{b}}.
\newblock Universal vision-language dense retrieval: Learning a unified representation space for multi-modal retrieval.
\newblock In \emph{Proceedings of ICLR}.

\bibitem[{Liu et~al.(2025{\natexlab{d}})Liu, Zhu, Zhou, Zhang, Yi, Yan, Gu, Yu, and Sun}]{liu2025benchmarking}
Zhenghao Liu, Xingsheng Zhu, Tianshuo Zhou, Xinyi Zhang, Xiaoyuan Yi, Yukun Yan, Yu~Gu, Ge~Yu, and Maosong Sun. 2025{\natexlab{d}}.
\newblock Benchmarking retrieval-augmented generation in multi-modal contexts.
\newblock \emph{arXiv preprint arXiv:2502.17297}.

\bibitem[{Liu et~al.(2021)Liu, Rodriguez-Opazo, Teney, and Gould}]{Liu_2021_ICCV}
Zheyuan Liu, Cristian Rodriguez-Opazo, Damien Teney, and Stephen Gould. 2021.
\newblock Image retrieval on real-life images with pre-trained vision-and-language models.
\newblock In \emph{Proceedings of the IEEE/CVF International Conference on Computer Vision (ICCV)}, pages 2125--2134.

\bibitem[{Liu et~al.(2016)Liu, Luo, Qiu, Wang, and Tang}]{Liu_2016_CVPR}
Ziwei Liu, Ping Luo, Shi Qiu, Xiaogang Wang, and Xiaoou Tang. 2016.
\newblock Deepfashion: Powering robust clothes recognition and retrieval with rich annotations.
\newblock In \emph{Proceedings of the IEEE Conference on Computer Vision and Pattern Recognition (CVPR)}.

\bibitem[{Lu et~al.(2022)Lu, Mishra, Xia, Qiu, Chang, Zhu, Tafjord, Clark, and Kalyan}]{lu2022learn}
Pan Lu, Swaroop Mishra, Tony Xia, Liang Qiu, Kai-Wei Chang, Song-Chun Zhu, Oyvind Tafjord, Peter Clark, and Ashwin Kalyan. 2022.
\newblock Learn to explain: Multimodal reasoning via thought chains for science question answering.
\newblock In \emph{The 36th Conference on Neural Information Processing Systems (NeurIPS)}.

\bibitem[{Lu et~al.(2024)Lu, Li, Chen, Xu, Luo, Zhang, and Ye}]{lu2024ovisstructuralembeddingalignment}
Shiyin Lu, Yang Li, Qing-Guo Chen, Zhao Xu, Weihua Luo, Kaifu Zhang, and Han-Jia Ye. 2024.
\newblock \href {https://arxiv.org/abs/2405.20797} {Ovis: Structural embedding alignment for multimodal large language model}.
\newblock \emph{Preprint}, arXiv:2405.20797.

\bibitem[{Luo et~al.(2024{\natexlab{a}})Luo, Zheng, Zhu, and You}]{luo2024doestextualinformationaffect}
Yang Luo, Zangwei Zheng, Zirui Zhu, and Yang You. 2024{\natexlab{a}}.
\newblock \href {https://doi.org/10.18653/v1/2024.emnlp-main.305} {How does the textual information affect the retrieval of multimodal in-context learning?}
\newblock In \emph{Proceedings of the 2024 Conference on Empirical Methods in Natural Language Processing}, pages 5321--5335, Miami, Florida, USA. Association for Computational Linguistics.

\bibitem[{Luo et~al.(2024{\natexlab{b}})Luo, Zheng, Yang, Li, Lin, Huang, Ji, Chao, Luo, and Ji}]{luo2024videoragvisuallyalignedretrievalaugmentedlong}
Yongdong Luo, Xiawu Zheng, Xiao Yang, Guilin Li, Haojia Lin, Jinfa Huang, Jiayi Ji, Fei Chao, Jiebo Luo, and Rongrong Ji. 2024{\natexlab{b}}.
\newblock \href {https://arxiv.org/abs/2411.13093} {Video-rag: Visually-aligned retrieval-augmented long video comprehension}.
\newblock \emph{Preprint}, arXiv:2411.13093.

\bibitem[{Ma et~al.(2024{\natexlab{a}})Ma, Lin, Li, Chen, and Lin}]{ma2024unifyingmultimodalretrievaldocument}
Xueguang Ma, Sheng-Chieh Lin, Minghan Li, Wenhu Chen, and Jimmy Lin. 2024{\natexlab{a}}.
\newblock \href {https://arxiv.org/abs/2406.11251} {Unifying multimodal retrieval via document screenshot embedding}.
\newblock \emph{Preprint}, arXiv:2406.11251.

\bibitem[{Ma et~al.(2024{\natexlab{b}})Ma, Zhuang, Koopman, Zuccon, Chen, and Lin}]{ma2024visaretrievalaugmentedgeneration}
Xueguang Ma, Shengyao Zhuang, Bevan Koopman, Guido Zuccon, Wenhu Chen, and Jimmy Lin. 2024{\natexlab{b}}.
\newblock \href {https://arxiv.org/abs/2412.14457} {Visa: Retrieval augmented generation with visual source attribution}.
\newblock \emph{Preprint}, arXiv:2412.14457.

\bibitem[{Ma et~al.(2024{\natexlab{c}})Ma, Zang, Chen, Chen, Jiao, Li, Lu, Liu, Ma, Dong, Zhang, Pan, Jiang, Wang, Cao, and Sun}]{ma2024mmlongbenchdocbenchmarkinglongcontextdocument}
Yubo Ma, Yuhang Zang, Liangyu Chen, Meiqi Chen, Yizhu Jiao, Xinze Li, Xinyuan Lu, Ziyu Liu, Yan Ma, Xiaoyi Dong, Pan Zhang, Liangming Pan, Yu-Gang Jiang, Jiaqi Wang, Yixin Cao, and Aixin Sun. 2024{\natexlab{c}}.
\newblock \href {https://arxiv.org/abs/2407.01523} {Mmlongbench-doc: Benchmarking long-context document understanding with visualizations}.
\newblock \emph{Preprint}, arXiv:2407.01523.

\bibitem[{Ma et~al.(2024{\natexlab{d}})Ma, Lan, Tu, Hu, Huang, and Mao}]{ma2024multimodalretrievalaugmentedmultimodal}
Zi-Ao Ma, Tian Lan, Rong-Cheng Tu, Yong Hu, Heyan Huang, and Xian-Ling Mao. 2024{\natexlab{d}}.
\newblock \href {https://arxiv.org/abs/2411.16365} {Multi-modal retrieval augmented multi-modal generation: A benchmark, evaluate metrics and strong baselines}.
\newblock \emph{Preprint}, arXiv:2411.16365.

\bibitem[{Ma et~al.(2024{\natexlab{e}})Ma, Gou, Shi, Sun, Li, Rezatofighi, and Cai}]{ma2024drvideodocumentretrievalbased}
Ziyu Ma, Chenhui Gou, Hengcan Shi, Bin Sun, Shutao Li, Hamid Rezatofighi, and Jianfei Cai. 2024{\natexlab{e}}.
\newblock \href {https://arxiv.org/abs/2406.12846} {Drvideo: Document retrieval based long video understanding}.
\newblock \emph{Preprint}, arXiv:2406.12846.

\bibitem[{Mao et~al.(2024)Mao, Bai, Hou, Shang, Jiang, Liu, and Wong}]{mao-etal-2024-visually}
Zhiming Mao, Haoli Bai, Lu~Hou, Lifeng Shang, Xin Jiang, Qun Liu, and Kam-Fai Wong. 2024.
\newblock \href {https://doi.org/10.18653/v1/2024.naacl-long.264} {Visually guided generative text-layout pre-training for document intelligence}.
\newblock In \emph{Proceedings of the 2024 Conference of the North American Chapter of the Association for Computational Linguistics: Human Language Technologies (Volume 1: Long Papers)}, pages 4713--4730, Mexico City, Mexico. Association for Computational Linguistics.

\bibitem[{Marino et~al.(2019)Marino, Chen, Gupta, Rohrbach, and Parikh}]{marino2019okvqa}
Kenneth Marino, Xinlei Chen, Abhinav Gupta, Marcus Rohrbach, and Devi Parikh. 2019.
\newblock Ok-vqa: A visual question answering benchmark requiring external knowledge.
\newblock \emph{Proceedings of the IEEE/CVF Conference on Computer Vision and Pattern Recognition (CVPR)}, pages 3195--3204.

\bibitem[{Masry et~al.(2022)Masry, Long, Tan, Joty, and Hoque}]{masry2022chartqa}
Ahmed Masry, Do~Xuan Long, Jia~Qing Tan, Shafiq Joty, and Enamul Hoque. 2022.
\newblock Chartqa: A benchmark for question answering about charts with visual and logical reasoning.
\newblock \emph{arXiv preprint arXiv:2203.10244}.

\bibitem[{Mathew et~al.(2021)Mathew, Karatzas, and Jawahar}]{mathew2021docvqa}
Minesh Mathew, Dimosthenis Karatzas, and CV~Jawahar. 2021.
\newblock Docvqa: A dataset for vqa on document images.
\newblock In \emph{Proceedings of the IEEE/CVF winter conference on applications of computer vision}, pages 2200--2209.

\bibitem[{Mei et~al.(2024)Mei, Meng, Liu, Kong, Ko, Zhao, Plumbley, Zou, and Wang}]{mei2024wavcaps}
Xinhao Mei, Chutong Meng, Haohe Liu, Qiuqiang Kong, Tom Ko, Chengqi Zhao, Mark~D Plumbley, Yuexian Zou, and Wenwu Wang. 2024.
\newblock Wavcaps: A chatgpt-assisted weakly-labelled audio captioning dataset for audio-language multimodal research.
\newblock \emph{IEEE/ACM Transactions on Audio, Speech, and Language Processing}.

\bibitem[{Miech et~al.(2019)Miech, Zhukov, Alayrac, Tapaswi, Laptev, and Sivic}]{Miech_2019_ICCV}
Antoine Miech, Dimitri Zhukov, Jean-Baptiste Alayrac, Makarand Tapaswi, Ivan Laptev, and Josef Sivic. 2019.
\newblock Howto100m: Learning a text-video embedding by watching hundred million narrated video clips.
\newblock In \emph{Proceedings of the IEEE/CVF International Conference on Computer Vision (ICCV)}.

\bibitem[{Min et~al.(2025)Min, Mundnich, Lapastora, Soltanmohammadi, Ronanki, and Han}]{10888900}
Do~June Min, Karel Mundnich, Andy Lapastora, Erfan Soltanmohammadi, Srikanth Ronanki, and Kyu Han. 2025.
\newblock \href {https://doi.org/10.1109/ICASSP49660.2025.10888900} {Speech retrieval-augmented generation without automatic speech recognition}.
\newblock In \emph{ICASSP 2025 - 2025 IEEE International Conference on Acoustics, Speech and Signal Processing (ICASSP)}, pages 1--5.

\bibitem[{Mortaheb et~al.(2025{\natexlab{a}})Mortaheb, Khojastepour, Chakradhar, and Ulukus}]{mortaheb2025ragcheckevaluatingmultimodalretrieval}
Matin Mortaheb, Mohammad A.~Amir Khojastepour, Srimat~T. Chakradhar, and Sennur Ulukus. 2025{\natexlab{a}}.
\newblock \href {https://arxiv.org/abs/2501.03995} {Rag-check: Evaluating multimodal retrieval augmented generation performance}.
\newblock \emph{Preprint}, arXiv:2501.03995.

\bibitem[{Mortaheb et~al.(2025{\natexlab{b}})Mortaheb, Khojastepour, Chakradhar, and Ulukus}]{mortaheb2025rerankingcontextmultimodalretrieval}
Matin Mortaheb, Mohammad A.~Amir Khojastepour, Srimat~T. Chakradhar, and Sennur Ulukus. 2025{\natexlab{b}}.
\newblock \href {https://arxiv.org/abs/2501.04695} {Re-ranking the context for multimodal retrieval augmented generation}.
\newblock \emph{Preprint}, arXiv:2501.04695.

\bibitem[{Nan et~al.(2024{\natexlab{a}})Nan, Xie, Zhou, Fan, Yang, Chen, Li, Yang, and Tai}]{nan2024openvid}
Kepan Nan, Rui Xie, Penghao Zhou, Tiehan Fan, Zhenheng Yang, Zhijie Chen, Xiang Li, Jian Yang, and Ying Tai. 2024{\natexlab{a}}.
\newblock Openvid-1m: A large-scale high-quality dataset for text-to-video generation.
\newblock \emph{arXiv preprint arXiv:2407.02371}.

\bibitem[{Nan et~al.(2024{\natexlab{b}})Nan, Fang, Rasteh, Lahabi, Zou, Zhao, and Cohan}]{nan-etal-2024-omg}
Linyong Nan, Weining Fang, Aylin Rasteh, Pouya Lahabi, Weijin Zou, Yilun Zhao, and Arman Cohan. 2024{\natexlab{b}}.
\newblock \href {https://doi.org/10.18653/v1/2024.emnlp-industry.75} {{OMG}-{QA}: Building open-domain multi-modal generative question answering systems}.
\newblock In \emph{Proceedings of the 2024 Conference on Empirical Methods in Natural Language Processing: Industry Track}, pages 1001--1015, Miami, Florida, US. Association for Computational Linguistics.

\bibitem[{Nashid et~al.(2023)Nashid, Sintaha, and Mesbah}]{CEDAR}
Noor Nashid, Mifta Sintaha, and Ali Mesbah. 2023.
\newblock \href {https://doi.org/10.1109/ICSE48619.2023.00205} {Retrieval-based prompt selection for code-related few-shot learning}.
\newblock In \emph{2023 IEEE/ACM 45th International Conference on Software Engineering (ICSE)}, pages 2450--2462.

\bibitem[{Naveed et~al.(2024)Naveed, Khan, Qiu, Saqib, Anwar, Usman, Akhtar, Barnes, and Mian}]{naveed2024comprehensiveoverviewlargelanguage}
Humza Naveed, Asad~Ullah Khan, Shi Qiu, Muhammad Saqib, Saeed Anwar, Muhammad Usman, Naveed Akhtar, Nick Barnes, and Ajmal Mian. 2024.
\newblock \href {https://arxiv.org/abs/2307.06435} {A comprehensive overview of large language models}.
\newblock \emph{Preprint}, arXiv:2307.06435.

\bibitem[{Nazar et~al.(2024)Nazar, Celik, Selim, Abdallah, Qiao, and Eltawil}]{nazar2024enwar}
Ahmad~M Nazar, Abdulkadir Celik, Mohamed~Y Selim, Asmaa Abdallah, Daji Qiao, and Ahmed~M Eltawil. 2024.
\newblock Enwar: A rag-empowered multi-modal llm framework for wireless environment perception.
\newblock \emph{arXiv preprint arXiv:2410.18104}.

\bibitem[{Nguyen et~al.(2024)Nguyen, Hendriksen, Yates, and Rijke}]{nguyen2024multimodallearnedsparseretrieval}
Thong Nguyen, Mariya Hendriksen, Andrew Yates, and Maarten~de Rijke. 2024.
\newblock Multimodal learned sparse retrieval with probabilistic expansion control.
\newblock In \emph{Advances in Information Retrieval}, pages 448--464, Cham. Springer Nature Switzerland.

\bibitem[{Niu et~al.(2021)Niu, Tang, Zhang, Lu, Hua, and Wen}]{niu2021counterfactual}
Yulei Niu, Kaihua Tang, Hanwang Zhang, Zhiwu Lu, Xian-Sheng Hua, and Ji-Rong Wen. 2021.
\newblock Counterfactual vqa: A cause-effect look at language bias.
\newblock In \emph{Proceedings of the IEEE/CVF conference on computer vision and pattern recognition}, pages 12700--12710.

\bibitem[{OpenAI et~al.(2024)OpenAI, Achiam, Adler, Agarwal, Ahmad, Akkaya, Aleman, Almeida, Altenschmidt, Altman, Anadkat, Avila, Babuschkin, Balaji, Balcom, Baltescu, Bao, Bavarian, Belgum, Bello, Berdine, Bernadett-Shapiro, Berner, Bogdonoff, Boiko, Boyd, Brakman, Brockman, Brooks, Brundage, Button, Cai, Campbell, Cann, Carey, Carlson, Carmichael, Chan, Chang, Chantzis, Chen, Chen, Chen, Chen, Chen, Chess, Cho, Chu, Chung, Cummings, Currier, Dai, Decareaux, Degry, and et~al.}]{openai2024gpt4technicalreport}
OpenAI, Josh Achiam, Steven Adler, Sandhini Agarwal, Lama Ahmad, Ilge Akkaya, Florencia~Leoni Aleman, Diogo Almeida, Janko Altenschmidt, Sam Altman, Shyamal Anadkat, Red Avila, Igor Babuschkin, Suchir Balaji, Valerie Balcom, Paul Baltescu, Haiming Bao, Mohammad Bavarian, Jeff Belgum, Irwan Bello, Jake Berdine, Gabriel Bernadett-Shapiro, Christopher Berner, Lenny Bogdonoff, Oleg Boiko, Madelaine Boyd, Anna-Luisa Brakman, Greg Brockman, Tim Brooks, Miles Brundage, Kevin Button, Trevor Cai, Rosie Campbell, Andrew Cann, Brittany Carey, Chelsea Carlson, Rory Carmichael, Brooke Chan, Che Chang, Fotis Chantzis, Derek Chen, Sully Chen, Ruby Chen, Jason Chen, Mark Chen, Ben Chess, Chester Cho, Casey Chu, Hyung~Won Chung, Dave Cummings, Jeremiah Currier, Yunxing Dai, Cory Decareaux, Thomas Degry, and et~al. 2024.
\newblock \href {https://arxiv.org/abs/2303.08774} {Gpt-4 technical report}.
\newblock \emph{Preprint}, arXiv:2303.08774.

\bibitem[{Oquab et~al.(2023)Oquab, Darcet, Moutakanni, Vo, Szafraniec, Khalidov, Fernandez, Haziza, Massa, El-Nouby, Howes, Huang, Xu, Sharma, Li, Galuba, Rabbat, Assran, Ballas, Synnaeve, Misra, Jegou, Mairal, Labatut, Joulin, and Bojanowski}]{oquab2023dinov2}
Maxime Oquab, Timothée Darcet, Theo Moutakanni, Huy~V. Vo, Marc Szafraniec, Vasil Khalidov, Pierre Fernandez, Daniel Haziza, Francisco Massa, Alaaeldin El-Nouby, Russell Howes, Po-Yao Huang, Hu~Xu, Vasu Sharma, Shang-Wen Li, Wojciech Galuba, Mike Rabbat, Mido Assran, Nicolas Ballas, Gabriel Synnaeve, Ishan Misra, Herve Jegou, Julien Mairal, Patrick Labatut, Armand Joulin, and Piotr Bojanowski. 2023.
\newblock Dinov2: Learning robust visual features without supervision.

\bibitem[{Ou et~al.(2025)Ou, Chen, Liang, Gou, Xiong, Zhang, Lai, and Zhang}]{10.1007/s00530-024-01649-6}
Weihua Ou, Yingjie Chen, Linqing Liang, Jianping Gou, Jiahao Xiong, Jiacheng Zhang, Lingge Lai, and Lei Zhang. 2025.
\newblock \href {https://doi.org/10.1007/s00530-024-01649-6} {Cross-modal retrieval of chest x-ray images and diagnostic reports based on report entity graph and dual attention: Cross-modal retrieval of chest x-ray images and diagnostic reports...}
\newblock \emph{Multimedia Syst.}, 31(1).

\bibitem[{Ouyang et~al.(2024)Ouyang, Qu, Zhou, Zhu, Zhang, Lin, Wang, Zhao, Jiang, Zhao, Shi, Wu, Chu, Liu, Li, Xu, Zhang, Shi, Tu, and He}]{ouyang2024omnidocbenchbenchmarkingdiversepdf}
Linke Ouyang, Yuan Qu, Hongbin Zhou, Jiawei Zhu, Rui Zhang, Qunshu Lin, Bin Wang, Zhiyuan Zhao, Man Jiang, Xiaomeng Zhao, Jin Shi, Fan Wu, Pei Chu, Minghao Liu, Zhenxiang Li, Chao Xu, Bo~Zhang, Botian Shi, Zhongying Tu, and Conghui He. 2024.
\newblock \href {https://arxiv.org/abs/2412.07626} {Omnidocbench: Benchmarking diverse pdf document parsing with comprehensive annotations}.
\newblock \emph{Preprint}, arXiv:2412.07626.

\bibitem[{Ouyang et~al.(2022)Ouyang, Wu, Jiang, Almeida, Wainwright, Mishkin, Zhang, Agarwal, Slama, Ray, Schulman, Hilton, Kelton, Miller, Simens, Askell, Welinder, Christiano, Leike, and Lowe}]{NEURIPS2022_b1efde53}
Long Ouyang, Jeffrey Wu, Xu~Jiang, Diogo Almeida, Carroll Wainwright, Pamela Mishkin, Chong Zhang, Sandhini Agarwal, Katarina Slama, Alex Ray, John Schulman, Jacob Hilton, Fraser Kelton, Luke Miller, Maddie Simens, Amanda Askell, Peter Welinder, Paul~F Christiano, Jan Leike, and Ryan Lowe. 2022.
\newblock \href {https://proceedings.neurips.cc/paper_files/paper/2022/file/b1efde53be364a73914f58805a001731-Paper-Conference.pdf} {Training language models to follow instructions with human feedback}.
\newblock In \emph{Advances in Neural Information Processing Systems}, volume~35, pages 27730--27744. Curran Associates, Inc.

\bibitem[{Overwijk et~al.(2022)Overwijk, Xiong, and Callan}]{clueweb22}
Arnold Overwijk, Chenyan Xiong, and Jamie Callan. 2022.
\newblock \href {https://doi.org/10.1145/3477495.3536321} {Clueweb22: 10 billion web documents with rich information}.
\newblock In \emph{Proceedings of the 45th International ACM SIGIR Conference on Research and Development in Information Retrieval}, SIGIR '22, page 3360–3362, New York, NY, USA. Association for Computing Machinery.

\bibitem[{Panayotov et~al.(2015)Panayotov, Chen, Povey, and Khudanpur}]{panayotov2015librispeech}
Vassil Panayotov, Guoguo Chen, Daniel Povey, and Sanjeev Khudanpur. 2015.
\newblock Librispeech: An asr corpus based on public domain audio books.
\newblock \emph{Proceedings of the IEEE International Conference on Acoustics, Speech and Signal Processing (ICASSP)}, pages 5206--5210.

\bibitem[{Papineni et~al.(2002)Papineni, Roukos, Ward, and Zhu}]{papineni-etal-2002-bleu}
Kishore Papineni, Salim Roukos, Todd Ward, and Wei-Jing Zhu. 2002.
\newblock \href {https://doi.org/10.3115/1073083.1073135} {{B}leu: a method for automatic evaluation of machine translation}.
\newblock In \emph{Proceedings of the 40th Annual Meeting of the Association for Computational Linguistics}, pages 311--318, Philadelphia, Pennsylvania, USA. Association for Computational Linguistics.

\bibitem[{Parvez et~al.(2021)Parvez, Ahmad, Chakraborty, Ray, and Chang}]{parvez2021retrieval}
Md~Rizwan Parvez, Wasi~Uddin Ahmad, Saikat Chakraborty, Baishakhi Ray, and Kai-Wei Chang. 2021.
\newblock Retrieval augmented code generation and summarization.
\newblock In \emph{EMNLP-Findings}.

\bibitem[{Pavlopoulos et~al.(2019)Pavlopoulos, Kougia, and Androutsopoulos}]{pavlopoulos-etal-2019-survey}
John Pavlopoulos, Vasiliki Kougia, and Ion Androutsopoulos. 2019.
\newblock \href {https://doi.org/10.18653/v1/W19-1803} {A survey on biomedical image captioning}.
\newblock In \emph{Proceedings of the Second Workshop on Shortcomings in Vision and Language}, pages 26--36, Minneapolis, Minnesota. Association for Computational Linguistics.

\bibitem[{Penamakuri et~al.(2023)Penamakuri, Gupta, Gupta, and Mishra}]{retvqa}
Abhirama~Subramanyam Penamakuri, Manish Gupta, Mithun~Das Gupta, and Anand Mishra. 2023.
\newblock \href {https://doi.org/10.24963/ijcai.2023/146} {Answer mining from a pool of images: Towards retrieval-based visual question answering}.
\newblock In \emph{IJCAI}. ijcai.org.

\bibitem[{Pramanick et~al.(2023)Pramanick, Jing, Nag, Zhu, Shah, LeCun, and Chellappa}]{pramanick2023volta}
Shraman Pramanick, Li~Jing, Sayan Nag, Jiachen Zhu, Hardik Shah, Yann LeCun, and Rama Chellappa. 2023.
\newblock Volta: Vision-language transformer with weakly-supervised local-feature alignment.
\newblock \emph{TMLR}.

\bibitem[{Procko and Ochoa(2024)}]{peng2024graphragsurvey}
Tyler~Thomas Procko and Omar Ochoa. 2024.
\newblock \href {https://doi.org/10.1109/AIxSET62544.2024.00030} {Graph retrieval-augmented generation for large language models: A survey}.
\newblock In \emph{2024 Conference on AI, Science, Engineering, and Technology (AIxSET)}, pages 166--169.

\bibitem[{Qin et~al.(2024)Qin, Song, Hu, Yao, Cho, Wang, Wu, Liu, Liu, and Yu}]{qin-etal-2024-infobench}
Yiwei Qin, Kaiqiang Song, Yebowen Hu, Wenlin Yao, Sangwoo Cho, Xiaoyang Wang, Xuansheng Wu, Fei Liu, Pengfei Liu, and Dong Yu. 2024.
\newblock \href {https://doi.org/10.18653/v1/2024.findings-acl.772} {{I}n{F}o{B}ench: Evaluating instruction following ability in large language models}.
\newblock In \emph{Findings of the Association for Computational Linguistics: ACL 2024}, pages 13025--13048, Bangkok, Thailand. Association for Computational Linguistics.

\bibitem[{Qwen et~al.(2025)Qwen, :, Yang, Yang, Zhang, Hui, Zheng, Yu, Li, Liu, Huang, Wei, Lin, Yang, Tu, Zhang, Yang, Yang, Zhou, Lin, Dang, Lu, Bao, Yang, Yu, Li, Xue, Zhang, Zhu, Men, Lin, Li, Tang, Xia, Ren, Ren, Fan, Su, Zhang, Wan, Liu, Cui, Zhang, and Qiu}]{qwen2025qwen25technicalreport}
Qwen, :, An~Yang, Baosong Yang, Beichen Zhang, Binyuan Hui, Bo~Zheng, Bowen Yu, Chengyuan Li, Dayiheng Liu, Fei Huang, Haoran Wei, Huan Lin, Jian Yang, Jianhong Tu, Jianwei Zhang, Jianxin Yang, Jiaxi Yang, Jingren Zhou, Junyang Lin, Kai Dang, Keming Lu, Keqin Bao, Kexin Yang, Le~Yu, Mei Li, Mingfeng Xue, Pei Zhang, Qin Zhu, Rui Men, Runji Lin, Tianhao Li, Tianyi Tang, Tingyu Xia, Xingzhang Ren, Xuancheng Ren, Yang Fan, Yang Su, Yichang Zhang, Yu~Wan, Yuqiong Liu, Zeyu Cui, Zhenru Zhang, and Zihan Qiu. 2025.
\newblock \href {https://arxiv.org/abs/2412.15115} {Qwen2.5 technical report}.
\newblock \emph{Preprint}, arXiv:2412.15115.

\bibitem[{Radford et~al.(2021)Radford, Kim, Hallacy, Ramesh, Goh, Agarwal, Sastry, Askell, Mishkin, Clark et~al.}]{radford2021learning}
Alec Radford, Jong~Wook Kim, Chris Hallacy, Aditya Ramesh, Gabriel Goh, Sandhini Agarwal, Girish Sastry, Amanda Askell, Pamela Mishkin, Jack Clark, et~al. 2021.
\newblock Learning transferable visual models from natural language supervision.
\newblock In \emph{International conference on machine learning}, pages 8748--8763. PMLR.

\bibitem[{Rahimi et~al.(2025)Rahimi, Cattoni, Beghili, Abrini, Khoramshahi, Pino, and Chetouani}]{rahimi2025reasoningllmsuserawaremultimodal}
Hamed Rahimi, Jeanne Cattoni, Meriem Beghili, Mouad Abrini, Mahdi Khoramshahi, Maribel Pino, and Mohamed Chetouani. 2025.
\newblock \href {https://arxiv.org/abs/2504.01700} {Reasoning llms for user-aware multimodal conversational agents}.
\newblock \emph{Preprint}, arXiv:2504.01700.

\bibitem[{Ramaswamy et~al.(2023)Ramaswamy, Lin, Zhao, Adcock, van~der Maaten, Ghadiyaram, and Russakovsky}]{geode}
Vikram~V. Ramaswamy, Sing~Yu Lin, Dora Zhao, Aaron~B. Adcock, Laurens van~der Maaten, Deepti Ghadiyaram, and Olga Russakovsky. 2023.
\newblock \href {https://arxiv.org/abs/2301.02560} {Geode: a geographically diverse evaluation dataset for object recognition}.
\newblock \emph{Preprint}, arXiv:2301.02560.

\bibitem[{Rao et~al.(2024)Rao, Choudhary, Deshpande, Satzoda, and Appalaraju}]{rao2024ravenmultitaskretrievalaugmented}
Varun~Nagaraj Rao, Siddharth Choudhary, Aditya Deshpande, Ravi~Kumar Satzoda, and Srikar Appalaraju. 2024.
\newblock \href {https://arxiv.org/abs/2406.19150} {Raven: Multitask retrieval augmented vision-language learning}.
\newblock \emph{Preprint}, arXiv:2406.19150.

\bibitem[{Rau et~al.(2024)Rau, Wang, Déjean, and Clinchant}]{rau2024contextembeddingsefficientanswer}
David Rau, Shuai Wang, Hervé Déjean, and Stéphane Clinchant. 2024.
\newblock \href {https://arxiv.org/abs/2407.09252} {Context embeddings for efficient answer generation in rag}.
\newblock \emph{Preprint}, arXiv:2407.09252.

\bibitem[{Ren et~al.(2025)Ren, Xu, Xia, Wang, Yin, and Huang}]{ren2025videoragretrievalaugmentedgenerationextreme}
Xubin Ren, Lingrui Xu, Long Xia, Shuaiqiang Wang, Dawei Yin, and Chao Huang. 2025.
\newblock \href {https://arxiv.org/abs/2502.01549} {Videorag: Retrieval-augmented generation with extreme long-context videos}.
\newblock \emph{Preprint}, arXiv:2502.01549.

\bibitem[{Riedler and Langer(2024)}]{riedler2024textoptimizingragmultimodal}
Monica Riedler and Stefan Langer. 2024.
\newblock \href {https://arxiv.org/abs/2410.21943} {Beyond text: Optimizing rag with multimodal inputs for industrial applications}.
\newblock \emph{Preprint}, arXiv:2410.21943.

\bibitem[{Robertson and Zaragoza(2009)}]{INR-019}
Stephen Robertson and Hugo Zaragoza. 2009.
\newblock \href {https://doi.org/10.1561/1500000019} {The probabilistic relevance framework: Bm25 and beyond}.
\newblock \emph{Foundations and Trends® in Information Retrieval}, 3(4):333--389.

\bibitem[{Rohrbach et~al.(2015)Rohrbach, Rohrbach, Tandon, and Schiele}]{rohrbach2015lsmdc}
Anna Rohrbach, Marcus Rohrbach, Nihar Tandon, and Bernt Schiele. 2015.
\newblock A dataset for movie description.
\newblock \emph{Proceedings of the IEEE Conference on Computer Vision and Pattern Recognition (CVPR)}, pages 1--10.

\bibitem[{Rostamzadeh et~al.(2018)Rostamzadeh, Hosseini, Boquet, Stokowiec, Zhang, Jauvin, and Pal}]{fashiongen}
Negar Rostamzadeh, Seyedarian Hosseini, Thomas Boquet, Wojciech Stokowiec, Ying Zhang, Christian Jauvin, and Chris Pal. 2018.
\newblock Fashion-gen: The generative fashion dataset and challenge.
\newblock \emph{arXiv preprint arXiv:1806.08317}.

\bibitem[{Saito et~al.(2023)Saito, Sohn, Zhang, Li, Lee, Saenko, and Pfister}]{saito2023pic2wordmappingpictureswords}
Kuniaki Saito, Kihyuk Sohn, Xiang Zhang, Chun-Liang Li, Chen-Yu Lee, Kate Saenko, and Tomas Pfister. 2023.
\newblock Pic2word: Mapping pictures to words for zero-shot composed image retrieval.
\newblock In \emph{Proceedings of the IEEE/CVF Conference on Computer Vision and Pattern Recognition}, pages 19305--19314.

\bibitem[{Sanguigni et~al.(2025)Sanguigni, Morelli, Cornia, and Cucchiara}]{sanguigni2025fashion}
Fulvio Sanguigni, Davide Morelli, Marcella Cornia, and Rita Cucchiara. 2025.
\newblock Fashion-rag: Multimodal fashion image editing via retrieval-augmented generation.
\newblock \emph{arXiv preprint arXiv:2504.14011}.

\bibitem[{Schneider et~al.(2025)Schneider, Ahmadi, Ahmadi, Vogel, Semmann, and Biemann}]{schneider2025collexmultimodalagentic}
Florian Schneider, Narges~Baba Ahmadi, Niloufar~Baba Ahmadi, Iris Vogel, Martin Semmann, and Chris Biemann. 2025.
\newblock \href {https://arxiv.org/abs/2504.07643} {Collex -- a multimodal agentic rag system enabling interactive exploration of scientific collections}.
\newblock \emph{Preprint}, arXiv:2504.07643.

\bibitem[{Schuhmann et~al.(2022)Schuhmann, Beaumont, Vencu, Gordon, Wightman, Cherti, Coombes, Katta, Mullis, Wortsman et~al.}]{schuhmann2022laion}
Christoph Schuhmann, Romain Beaumont, Richard Vencu, Cade Gordon, Ross Wightman, Mehdi Cherti, Theo Coombes, Aarush Katta, Clayton Mullis, Mitchell Wortsman, et~al. 2022.
\newblock Laion-5b: An open large-scale dataset for training next generation image-text models.
\newblock \emph{Advances in Neural Information Processing Systems}, 35:25278--25294.

\bibitem[{Schuhmann et~al.(2021)Schuhmann, Vencu, Beaumont, Kaczmarczyk, Jitsev, Komatsuzaki et~al.}]{laion400m}
Christoph Schuhmann, Romain Vencu, Richard Beaumont, Robert Kaczmarczyk, Jenia Jitsev, Atsushi Komatsuzaki, et~al. 2021.
\newblock Laion-400m: Open dataset of clip-filtered 400 million image-text pairs.
\newblock \emph{arXiv preprint arXiv:2111.02114}.

\bibitem[{Schwenk et~al.(2022)Schwenk, Khandelwal, Clark, Marino, and Mottaghi}]{schwenk2022aokvqabenchmarkvisualquestion}
Dustin Schwenk, Apoorv Khandelwal, Christopher Clark, Kenneth Marino, and Roozbeh Mottaghi. 2022.
\newblock A-okvqa: A benchmark for visual question answering using world knowledge.
\newblock In \emph{European Conference on Computer Vision}, pages 146--162.

\bibitem[{Sharifymoghaddam et~al.(2024)Sharifymoghaddam, Upadhyay, Chen, and Lin}]{Sharifymoghaddam2024UniRAGUR}
Sahel Sharifymoghaddam, Shivani Upadhyay, Wenhu Chen, and Jimmy Lin. 2024.
\newblock \href {https://api.semanticscholar.org/CorpusID:269790936} {Unirag: Universal retrieval augmentation for multi-modal large language models}.
\newblock \emph{ArXiv}, abs/2405.10311.

\bibitem[{Sharma et~al.(2018)Sharma, Ding, Goodman, and Soricut}]{Sharma2018ConceptualCA}
Piyush Sharma, Nan Ding, Sebastian Goodman, and Radu Soricut. 2018.
\newblock \href {https://api.semanticscholar.org/CorpusID:51876975} {Conceptual captions: A cleaned, hypernymed, image alt-text dataset for automatic image captioning}.
\newblock In \emph{Annual Meeting of the Association for Computational Linguistics}.

\bibitem[{Shen et~al.(2024)Shen, Huang, Lan, and Zheng}]{ijcai2024p136}
Xiaobo Shen, Qianxin Huang, Long Lan, and Yuhui Zheng. 2024.
\newblock \href {https://doi.org/10.24963/ijcai.2024/136} {Contrastive transformer cross-modal hashing for video-text retrieval}.
\newblock In \emph{Proceedings of the Thirty-Third International Joint Conference on Artificial Intelligence, {IJCAI-24}}, pages 1227--1235. International Joint Conferences on Artificial Intelligence Organization.
\newblock Main Track.

\bibitem[{Shi et~al.(2022)Shi, Wang, Tao, Du, Zhang, Han, Zhang, and Sun}]{shi-etal-2022-race}
Ensheng Shi, Yanlin Wang, Wei Tao, Lun Du, Hongyu Zhang, Shi Han, Dongmei Zhang, and Hongbin Sun. 2022.
\newblock \href {https://doi.org/10.18653/v1/2022.emnlp-main.372} {{RACE}: Retrieval-augmented commit message generation}.
\newblock In \emph{Proceedings of the 2022 Conference on Empirical Methods in Natural Language Processing}, pages 5520--5530, Abu Dhabi, United Arab Emirates. Association for Computational Linguistics.

\bibitem[{Shohan et~al.(2024)Shohan, Nayeem, Islam, Akash, and Joty}]{shohan2024xlheadtagsleveragingmultimodalretrieval}
Faisal~Tareque Shohan, Mir~Tafseer Nayeem, Samsul Islam, Abu~Ubaida Akash, and Shafiq Joty. 2024.
\newblock \href {https://doi.org/10.18653/v1/2024.findings-acl.771} {{XL}-{H}ead{T}ags: Leveraging multimodal retrieval augmentation for the multilingual generation of news headlines and tags}.
\newblock In \emph{Findings of the Association for Computational Linguistics: ACL 2024}, pages 12991--13024, Bangkok, Thailand. Association for Computational Linguistics.

\bibitem[{Shuster et~al.(2021)Shuster, Poff, Chen, Kiela, and Weston}]{shuster-etal-2021-retrieval-augmentation}
Kurt Shuster, Spencer Poff, Moya Chen, Douwe Kiela, and Jason Weston. 2021.
\newblock \href {https://doi.org/10.18653/v1/2021.findings-emnlp.320} {Retrieval augmentation reduces hallucination in conversation}.
\newblock In \emph{Findings of the Association for Computational Linguistics: EMNLP 2021}, pages 3784--3803, Punta Cana, Dominican Republic. Association for Computational Linguistics.

\bibitem[{Sigurdsson et~al.(2018)Sigurdsson, Varol, Farinella et~al.}]{charadesego}
Gunnar~A Sigurdsson, Gul Varol, Giovanni~Maria Farinella, et~al. 2018.
\newblock Charadesego: A dataset for egocentric video understanding.
\newblock \emph{arXiv preprint arXiv:1804.09626}.

\bibitem[{Singh et~al.(2025)Singh, Ehtesham, Kumar, and Khoei}]{singh2025agenticretrievalaugmentedgenerationsurvey}
Aditi Singh, Abul Ehtesham, Saket Kumar, and Tala~Talaei Khoei. 2025.
\newblock \href {https://arxiv.org/abs/2501.09136} {Agentic retrieval-augmented generation: A survey on agentic rag}.
\newblock \emph{Preprint}, arXiv:2501.09136.

\bibitem[{Singh et~al.(2022)Singh, Hu, Goswami, Couairon, Galuba, Rohrbach, and Kiela}]{singh2022flava}
Amanpreet Singh, Ronghang Hu, Vedanuj Goswami, Guillaume Couairon, Wojciech Galuba, Marcus Rohrbach, and Douwe Kiela. 2022.
\newblock Flava: A foundational language and vision alignment model.
\newblock In \emph{Proceedings of the IEEE/CVF Conference on Computer Vision and Pattern Recognition}, pages 15638--15650.

\bibitem[{Song et~al.(2025)Song, Li, Li, Zhao, Yu, Ma, Mao, Zhang, and Wang}]{song2025bridge}
Shezheng Song, Xiaopeng Li, Shasha Li, Shan Zhao, Jie Yu, Jun Ma, Xiaoguang Mao, Weimin Zhang, and Meng Wang. 2025.
\newblock How to bridge the gap between modalities: Survey on multimodal large language model.
\newblock \emph{IEEE Transactions on Knowledge and Data Engineering}.

\bibitem[{Strand et~al.(2024)Strand, Gautam, Midoglu, and Halvorsen}]{strand2024soccerragmultimodalsoccerinformation}
Aleksander~Theo Strand, Sushant Gautam, Cise Midoglu, and Pål Halvorsen. 2024.
\newblock \href {https://arxiv.org/abs/2406.01273} {Soccerrag: Multimodal soccer information retrieval via natural queries}.
\newblock \emph{Preprint}, arXiv:2406.01273.

\bibitem[{Su et~al.(2024{\natexlab{a}})Su, Wen, Kang, Wang, Su, Pan, Zhong, and Hossain}]{su2024hybrid}
Cheng Su, Jinbo Wen, Jiawen Kang, Yonghua Wang, Yuanjia Su, Hudan Pan, Zishao Zhong, and M~Shamim Hossain. 2024{\natexlab{a}}.
\newblock Hybrid rag-empowered multi-modal llm for secure data management in internet of medical things: A diffusion-based contract approach.
\newblock \emph{IEEE Internet of Things Journal}.

\bibitem[{Su et~al.(2024{\natexlab{b}})Su, Luo, Pan, Chou, Lal, and Howard}]{su2024sk}
Xin Su, Man Luo, Kris~W Pan, Tien~Pei Chou, Vasudev Lal, and Phillip Howard. 2024{\natexlab{b}}.
\newblock Sk-vqa: Synthetic knowledge generation at scale for training context-augmented multimodal llms.
\newblock \emph{arXiv preprint arXiv:2406.19593}.

\bibitem[{Sun et~al.(2025)Sun, Liu, Cui, Qi, hao Zhang, Zhou, and Lu}]{sun2025sealspeechembeddingalignment}
Chunyu Sun, Bingyu Liu, Zhichao Cui, Anbin Qi, Tian hao Zhang, Dinghao Zhou, and Lewei Lu. 2025.
\newblock \href {https://arxiv.org/abs/2502.02603} {Seal: Speech embedding alignment learning for speech large language model with retrieval-augmented generation}.
\newblock \emph{Preprint}, arXiv:2502.02603.

\bibitem[{Sun et~al.(2024{\natexlab{a}})Sun, Zhang, Zhou, Su, Qu, and Cheng}]{sun2024surfteachinglargevisionlanguage}
Jiashuo Sun, Jihai Zhang, Yucheng Zhou, Zhaochen Su, Xiaoye Qu, and Yu~Cheng. 2024{\natexlab{a}}.
\newblock Surf: Teaching large vision-language models to selectively utilize retrieved information.
\newblock In \emph{Proceedings of the 2024 Conference on Empirical Methods in Natural Language Processing}, pages 7611--7629.

\bibitem[{Sun et~al.(2024{\natexlab{b}})Sun, Zhao, Han, and Xiong}]{sun2024factawaremultimodalretrievalaugmentation}
Liwen Sun, James Zhao, Megan Han, and Chenyan Xiong. 2024{\natexlab{b}}.
\newblock \href {https://arxiv.org/abs/2407.15268} {Fact-aware multimodal retrieval augmentation for accurate medical radiology report generation}.
\newblock \emph{Preprint}, arXiv:2407.15268.

\bibitem[{Suri et~al.(2024)Suri, Mathur, Dernoncourt, Goswami, Rossi, and Manocha}]{suri2024visdommultidocumentqavisually}
Manan Suri, Puneet Mathur, Franck Dernoncourt, Kanika Goswami, Ryan~A Rossi, and Dinesh Manocha. 2024.
\newblock Visdom: Multi-document qa with visually rich elements using multimodal retrieval-augmented generation.
\newblock \emph{arXiv preprint arXiv:2412.10704}.

\bibitem[{Szegedy et~al.(2016)Szegedy, Vanhoucke, Ioffe, Shlens, and Wojna}]{szegedy2016rethinking}
Christian Szegedy, Vincent Vanhoucke, Sergey Ioffe, Jon Shlens, and Zbigniew Wojna. 2016.
\newblock Rethinking the inception architecture for computer vision.
\newblock In \emph{Proceedings of the IEEE conference on computer vision and pattern recognition}, pages 2818--2826.

\bibitem[{Talmor et~al.(2021)Talmor, Yoran, Catav, Lahav, Wang, Asai, Ilharco, Hajishirzi, and Berant}]{talmor2021multimodalqa}
Alon Talmor, Ori Yoran, Amnon Catav, Dan Lahav, Yizhong Wang, Akari Asai, Gabriel Ilharco, Hannaneh Hajishirzi, and Jonathan Berant. 2021.
\newblock \href {https://openreview.net/forum?id=ee6W5UgQLa} {Multimodal{\{}qa{\}}: complex question answering over text, tables and images}.
\newblock In \emph{International Conference on Learning Representations}.

\bibitem[{Tan et~al.(2024)Tan, Wei, Sun, Gao, Li, Yu, Guo, and Li}]{tan2024retrievalmeetsreasoninghighschool}
Cheng Tan, Jingxuan Wei, Linzhuang Sun, Zhangyang Gao, Siyuan Li, Bihui Yu, Ruifeng Guo, and Stan~Z. Li. 2024.
\newblock \href {https://arxiv.org/abs/2405.20834} {Retrieval meets reasoning: Even high-school textbook knowledge benefits multimodal reasoning}.
\newblock \emph{Preprint}, arXiv:2405.20834.

\bibitem[{Tang et~al.(2019)Tang, Wang, Wang et~al.}]{coin}
Yansong Tang, Xiaohan Wang, Jingdong Wang, et~al. 2019.
\newblock Coin: A large-scale dataset for comprehensive instructional video analysis.
\newblock \emph{Proceedings of the IEEE Conference on Computer Vision and Pattern Recognition (CVPR)}, pages 1--10.

\bibitem[{Team et~al.(2024)Team, Anil, Borgeaud, Alayrac, Yu, Soricut, Schalkwyk, Dai, Hauth, Millican, Silver, Johnson, Antonoglou, Schrittwieser, Glaese, Chen, Pitler, Lillicrap, Lazaridou, Firat, Molloy, Isard, Barham, Hennigan, Lee, Viola, Reynolds, Xu, Doherty, Collins, Meyer, Rutherford, Moreira, Ayoub, Goel, Krawczyk, Du, Chi, Cheng, Ni, Shah, Kane, Chan, Faruqui, Severyn, Lin, Li, Cheng, Ittycheriah, Mahdieh, Chen, Sun, Tran, Bagri, Lakshminarayanan, and et~al.}]{geminiteam2024geminifamilyhighlycapable}
Gemini Team, Rohan Anil, Sebastian Borgeaud, Jean-Baptiste Alayrac, Jiahui Yu, Radu Soricut, Johan Schalkwyk, Andrew~M. Dai, Anja Hauth, Katie Millican, David Silver, Melvin Johnson, Ioannis Antonoglou, Julian Schrittwieser, Amelia Glaese, Jilin Chen, Emily Pitler, Timothy Lillicrap, Angeliki Lazaridou, Orhan Firat, James Molloy, Michael Isard, Paul~R. Barham, Tom Hennigan, Benjamin Lee, Fabio Viola, Malcolm Reynolds, Yuanzhong Xu, Ryan Doherty, Eli Collins, Clemens Meyer, Eliza Rutherford, Erica Moreira, Kareem Ayoub, Megha Goel, Jack Krawczyk, Cosmo Du, Ed~Chi, Heng-Tze Cheng, Eric Ni, Purvi Shah, Patrick Kane, Betty Chan, Manaal Faruqui, Aliaksei Severyn, Hanzhao Lin, YaGuang Li, Yong Cheng, Abe Ittycheriah, Mahdis Mahdieh, Mia Chen, Pei Sun, Dustin Tran, Sumit Bagri, Balaji Lakshminarayanan, and et~al. 2024.
\newblock \href {https://arxiv.org/abs/2312.11805} {Gemini: A family of highly capable multimodal models}.
\newblock \emph{Preprint}, arXiv:2312.11805.

\bibitem[{Tiwari et~al.(2024)Tiwari, Kang, Lee, Lee, Piech, Thrun, Shomorony, and Zhang}]{tiwari2024faster}
Mo~Tiwari, Ryan Kang, Jaeyong Lee, Donghyun Lee, Christopher~J Piech, Sebastian Thrun, Ilan Shomorony, and Martin~Jinye Zhang. 2024.
\newblock \href {https://openreview.net/forum?id=FKkkdyRdsD} {Faster maximum inner product search in high dimensions}.
\newblock In \emph{Forty-first International Conference on Machine Learning}.

\bibitem[{Tong et~al.(2022)Tong, Song, Wang, and Wang}]{tong2022videomaemaskedautoencodersdataefficient}
Zhan Tong, Yibing Song, Jue Wang, and Limin Wang. 2022.
\newblock Video{MAE}: Masked autoencoders are data-efficient learners for self-supervised video pre-training.
\newblock In \emph{Advances in Neural Information Processing Systems}.

\bibitem[{Tonmoy et~al.(2024)Tonmoy, Zaman, Jain, Rani, Rawte, Chadha, and Das}]{tonmoy2024comprehensive}
SM~Tonmoy, SM~Zaman, Vinija Jain, Anku Rani, Vipula Rawte, Aman Chadha, and Amitava Das. 2024.
\newblock A comprehensive survey of hallucination mitigation techniques in large language models.
\newblock \emph{arXiv preprint arXiv:2401.01313}.

\bibitem[{Touvron et~al.(2023)Touvron, Martin, Stone, Albert, Almahairi, Babaei, Bashlykov, Batra, Bhargava, Bhosale, Bikel, Blecher, Ferrer, Chen, Cucurull, Esiobu, Fernandes, Fu, Fu, Fuller, Gao, Goswami, Goyal, Hartshorn, Hosseini, Hou, Inan, Kardas, Kerkez, Khabsa, Kloumann, Korenev, Koura, Lachaux, Lavril, Lee, Liskovich, Lu, Mao, Martinet, Mihaylov, Mishra, Molybog, Nie, and et~al.}]{touvron2023llama2openfoundation}
Hugo Touvron, Louis Martin, Kevin Stone, Peter Albert, Amjad Almahairi, Yasmine Babaei, Nikolay Bashlykov, Soumya Batra, Prajjwal Bhargava, Shruti Bhosale, Dan Bikel, Lukas Blecher, Cristian~Canton Ferrer, Moya Chen, Guillem Cucurull, David Esiobu, Jude Fernandes, Jeremy Fu, Wenyin Fu, Brian Fuller, Cynthia Gao, Vedanuj Goswami, Naman Goyal, Anthony Hartshorn, Saghar Hosseini, Rui Hou, Hakan Inan, Marcin Kardas, Viktor Kerkez, Madian Khabsa, Isabel Kloumann, Artem Korenev, Punit~Singh Koura, Marie-Anne Lachaux, Thibaut Lavril, Jenya Lee, Diana Liskovich, Yinghai Lu, Yuning Mao, Xavier Martinet, Todor Mihaylov, Pushkar Mishra, Igor Molybog, Yixin Nie, and et~al. 2023.
\newblock \href {https://arxiv.org/abs/2307.09288} {Llama 2: Open foundation and fine-tuned chat models}.
\newblock \emph{Preprint}, arXiv:2307.09288.

\bibitem[{van~den Oord et~al.(2019)van~den Oord, Li, and Vinyals}]{oord2019representationlearningcontrastivepredictive}
Aaron van~den Oord, Yazhe Li, and Oriol Vinyals. 2019.
\newblock \href {https://arxiv.org/abs/1807.03748} {Representation learning with contrastive predictive coding}.
\newblock \emph{Preprint}, arXiv:1807.03748.

\bibitem[{Vaswani et~al.(2017)Vaswani, Shazeer, Parmar, Uszkoreit, Jones, Gomez, Kaiser, and Polosukhin}]{10.5555/3295222.3295349}
Ashish Vaswani, Noam Shazeer, Niki Parmar, Jakob Uszkoreit, Llion Jones, Aidan~N. Gomez, \L{}ukasz Kaiser, and Illia Polosukhin. 2017.
\newblock Attention is all you need.
\newblock In \emph{Proceedings of the 31st International Conference on Neural Information Processing Systems}, NIPS'17, page 6000–6010, Red Hook, NY, USA. Curran Associates Inc.

\bibitem[{Vedantam et~al.(2015)Vedantam, Lawrence~Zitnick, and Parikh}]{vedantam2015cider}
Ramakrishna Vedantam, C~Lawrence~Zitnick, and Devi Parikh. 2015.
\newblock Cider: Consensus-based image description evaluation.
\newblock In \emph{Proceedings of the IEEE conference on computer vision and pattern recognition}, pages 4566--4575.

\bibitem[{Wang et~al.(2024{\natexlab{a}})Wang, Raman, Sibue, Ma, Babkin, Kaur, Pei, Nourbakhsh, and Liu}]{wang-etal-2024-docllm}
Dongsheng Wang, Natraj Raman, Mathieu Sibue, Zhiqiang Ma, Petr Babkin, Simerjot Kaur, Yulong Pei, Armineh Nourbakhsh, and Xiaomo Liu. 2024{\natexlab{a}}.
\newblock \href {https://doi.org/10.18653/v1/2024.acl-long.463} {{D}oc{LLM}: A layout-aware generative language model for multimodal document understanding}.
\newblock In \emph{Proceedings of the 62nd Annual Meeting of the Association for Computational Linguistics (Volume 1: Long Papers)}, pages 8529--8548, Bangkok, Thailand. Association for Computational Linguistics.

\bibitem[{Wang et~al.(2024{\natexlab{b}})Wang, Sun, Wang, Liu, Dianat, Rabbani, Rao, and Tao}]{wang2024textmassmodelingstochastic}
Jiamian Wang, Guohao Sun, Pichao Wang, Dongfang Liu, Sohail Dianat, Majid Rabbani, Raghuveer Rao, and Zhiqiang Tao. 2024{\natexlab{b}}.
\newblock Text is mass: Modeling as stochastic embedding for text-video retrieval.
\newblock In \emph{Proceedings of the IEEE/CVF Conference on Computer Vision and Pattern Recognition (CVPR)}, pages 16551--16560.

\bibitem[{Wang et~al.(2024{\natexlab{c}})Wang, Ke, Xu, Chen, Gao, Huang, and Zhu}]{wang2023musteffectivescalableframework}
Mengzhao Wang, Xiangyu Ke, Xiaoliang Xu, Lu~Chen, Yunjun Gao, Pinpin Huang, and Runkai Zhu. 2024{\natexlab{c}}.
\newblock Must: An effective and scalable framework for multimodal search of target modality.
\newblock In \emph{2024 IEEE 40th International Conference on Data Engineering (ICDE)}, pages 4747--4759. IEEE.

\bibitem[{Wang et~al.(2024{\natexlab{d}})Wang, Bai, Tan, Wang, Fan, Bai, Chen, Liu, Wang, Ge, Fan, Dang, Du, Ren, Men, Liu, Zhou, Zhou, and Lin}]{Qwen2VL}
Peng Wang, Shuai Bai, Sinan Tan, Shijie Wang, Zhihao Fan, Jinze Bai, Keqin Chen, Xuejing Liu, Jialin Wang, Wenbin Ge, Yang Fan, Kai Dang, Mengfei Du, Xuancheng Ren, Rui Men, Dayiheng Liu, Chang Zhou, Jingren Zhou, and Junyang Lin. 2024{\natexlab{d}}.
\newblock Qwen2-vl: Enhancing vision-language model's perception of the world at any resolution.
\newblock \emph{arXiv preprint arXiv:2409.12191}.

\bibitem[{Wang et~al.(2023{\natexlab{a}})Wang, Wang, Lin, Bai, Zhou, Zhou, Wang, and Zhou}]{wang2023one}
Peng Wang, Shijie Wang, Junyang Lin, Shuai Bai, Xiaohuan Zhou, Jingren Zhou, Xinggang Wang, and Chang Zhou. 2023{\natexlab{a}}.
\newblock One-peace: Exploring one general representation model toward unlimited modalities.
\newblock \emph{arXiv preprint arXiv:2305.11172}.

\bibitem[{Wang et~al.(2020{\natexlab{a}})Wang, Wei, Dong, Bao, Yang, and Zhou}]{wang2020minilmdeepselfattentiondistillation}
Wenhui Wang, Furu Wei, Li~Dong, Hangbo Bao, Nan Yang, and Ming Zhou. 2020{\natexlab{a}}.
\newblock Minilm: Deep self-attention distillation for task-agnostic compression of pre-trained transformers.
\newblock \emph{Advances in Neural Information Processing Systems}, 33:5776--5788.

\bibitem[{Wang et~al.(2019)Wang, Wu, Chen et~al.}]{wang2019vatex}
Xin Wang, Jiawei Wu, Junkun Chen, et~al. 2019.
\newblock Vatex: A large-scale, high-quality multilingual dataset for video-and-language research.
\newblock \emph{Proceedings of the IEEE International Conference on Computer Vision (ICCV)}, pages 1--10.

\bibitem[{Wang et~al.(2023{\natexlab{b}})Wang, He, Li, Li, Yu, Ma, Li, Chen, Chen, Wang et~al.}]{wang2023internvid}
Yi~Wang, Yinan He, Yizhuo Li, Kunchang Li, Jiashuo Yu, Xin Ma, Xinhao Li, Guo Chen, Xinyuan Chen, Yaohui Wang, et~al. 2023{\natexlab{b}}.
\newblock Internvid: A large-scale video-text dataset for multimodal understanding and generation.
\newblock In \emph{The Twelfth International Conference on Learning Representations}.

\bibitem[{Wang et~al.(2022)Wang, Li, Li, He, Huang, Zhao, Zhang, Xu, Liu, Wang, Xing, Chen, Pan, Yu, Wang, Wang, and Qiao}]{wang2022internvideogeneralvideofoundation}
Yi~Wang, Kunchang Li, Yizhuo Li, Yinan He, Bingkun Huang, Zhiyu Zhao, Hongjie Zhang, Jilan Xu, Yi~Liu, Zun Wang, Sen Xing, Guo Chen, Junting Pan, Jiashuo Yu, Yali Wang, Limin Wang, and Yu~Qiao. 2022.
\newblock \href {https://arxiv.org/abs/2212.03191} {Internvideo: General video foundation models via generative and discriminative learning}.
\newblock \emph{Preprint}, arXiv:2212.03191.

\bibitem[{Wang et~al.(2020{\natexlab{b}})Wang, Chen, and Hoi}]{wang2020deeplearningimagesuperresolution}
Zhihao Wang, Jian Chen, and Steven C.~H. Hoi. 2020{\natexlab{b}}.
\newblock \href {https://arxiv.org/abs/1902.06068} {Deep learning for image super-resolution: A survey}.
\newblock \emph{Preprint}, arXiv:1902.06068.

\bibitem[{Wei et~al.(2024{\natexlab{a}})Wei, Chen, Chen, Hu, Zhang, Fu, Ritter, and Chen}]{wei2023uniirtrainingbenchmarkinguniversal}
Cong Wei, Yang Chen, Haonan Chen, Hexiang Hu, Ge~Zhang, Jie Fu, Alan Ritter, and Wenhu Chen. 2024{\natexlab{a}}.
\newblock Uniir: Training and benchmarking universal multimodal information retrievers.
\newblock In \emph{European Conference on Computer Vision}, pages 387--404. Springer.

\bibitem[{Wei et~al.(2024{\natexlab{b}})Wei, Wang, Schuurmans, Bosma, Ichter, Xia, Chi, Le, and Zhou}]{10.5555/3600270.3602070}
Jason Wei, Xuezhi Wang, Dale Schuurmans, Maarten Bosma, Brian Ichter, Fei Xia, Ed~H. Chi, Quoc~V. Le, and Denny Zhou. 2024{\natexlab{b}}.
\newblock Chain-of-thought prompting elicits reasoning in large language models.
\newblock In \emph{Proceedings of the 36th International Conference on Neural Information Processing Systems}, NIPS '22, Red Hook, NY, USA. Curran Associates Inc.

\bibitem[{Wen et~al.(2024)Wen, Song, Chen, Wei, Nie, and Chua}]{Wen_2024}
Haokun Wen, Xuemeng Song, Xiaolin Chen, Yinwei Wei, Liqiang Nie, and Tat-Seng Chua. 2024.
\newblock \href {https://doi.org/10.1145/3626772.3657727} {Simple but effective raw-data level multimodal fusion for composed image retrieval}.
\newblock In \emph{Proceedings of the 47th International ACM SIGIR Conference on Research and Development in Information Retrieval}, SIGIR 2024, page 229–239. ACM.

\bibitem[{Winterbottom et~al.(2020)Winterbottom, Xiao, McLean, and Moubayed}]{winterbottom2020modalitybiastvqadataset}
Thomas Winterbottom, Sarah Xiao, Alistair McLean, and Noura~Al Moubayed. 2020.
\newblock \href {https://arxiv.org/abs/2012.10210} {On modality bias in the tvqa dataset}.
\newblock \emph{Preprint}, arXiv:2012.10210.

\bibitem[{Wu et~al.(2021)Wu, Gao, Guo, Al-Halah, Rennie, Grauman, and Feris}]{wu2019fashioniq}
Hui Wu, Yupeng Gao, Xiaoxiao Guo, Ziad Al-Halah, Steven Rennie, Kristen Grauman, and Rogerio Feris. 2021.
\newblock The fashion iq dataset: Retrieving images by combining side information and relative natural language feedback.
\newblock \emph{CVPR}.

\bibitem[{Wu et~al.(2024{\natexlab{a}})Wu, Jayanthi, Viswanathan, Rosenberg, Pakazad, Wu, and Neubig}]{wu-etal-2024-synthetic}
Ian Wu, Sravan Jayanthi, Vijay Viswanathan, Simon Rosenberg, Sina~Khoshfetrat Pakazad, Tongshuang Wu, and Graham Neubig. 2024{\natexlab{a}}.
\newblock \href {https://doi.org/10.18653/v1/2024.findings-emnlp.759} {Synthetic multimodal question generation}.
\newblock In \emph{Findings of the Association for Computational Linguistics: EMNLP 2024}, pages 12960--12993, Miami, Florida, USA. Association for Computational Linguistics.

\bibitem[{Wu et~al.(2024{\natexlab{b}})Wu, Fei, Qu, Ji, and Chua}]{wu24next}
Shengqiong Wu, Hao Fei, Leigang Qu, Wei Ji, and Tat-Seng Chua. 2024{\natexlab{b}}.
\newblock Next-gpt: Any-to-any multimodal llm.
\newblock In \emph{Proceedings of the International Conference on Machine Learning}, pages 53366--53397.

\bibitem[{Wu et~al.(2023)Wu, Chen, Zhang, Hui, Berg-Kirkpatrick, and Dubnov}]{10095969}
Yusong Wu, Ke~Chen, Tianyu Zhang, Yuchen Hui, Taylor Berg-Kirkpatrick, and Shlomo Dubnov. 2023.
\newblock \href {https://doi.org/10.1109/ICASSP49357.2023.10095969} {Large-scale contrastive language-audio pretraining with feature fusion and keyword-to-caption augmentation}.
\newblock In \emph{ICASSP 2023 - 2023 IEEE International Conference on Acoustics, Speech and Signal Processing (ICASSP)}, pages 1--5.

\bibitem[{Xia et~al.(2024{\natexlab{a}})Xia, Zhu, Li, Wang, Shi, Wang, Zhang, Zou, and Yao}]{xia2024mmedragversatilemultimodalrag}
Peng Xia, Kangyu Zhu, Haoran Li, Tianze Wang, Weijia Shi, Sheng Wang, Linjun Zhang, James Zou, and Huaxiu Yao. 2024{\natexlab{a}}.
\newblock \href {https://arxiv.org/abs/2410.13085} {Mmed-rag: Versatile multimodal rag system for medical vision language models}.
\newblock \emph{Preprint}, arXiv:2410.13085.

\bibitem[{Xia et~al.(2024{\natexlab{b}})Xia, Zhu, Li, Zhu, Li, Li, Zhang, and Yao}]{xia-etal-2024-rule}
Peng Xia, Kangyu Zhu, Haoran Li, Hongtu Zhu, Yun Li, Gang Li, Linjun Zhang, and Huaxiu Yao. 2024{\natexlab{b}}.
\newblock \href {https://doi.org/10.18653/v1/2024.emnlp-main.62} {{RULE}: Reliable multimodal {RAG} for factuality in medical vision language models}.
\newblock In \emph{Proceedings of the 2024 Conference on Empirical Methods in Natural Language Processing}, pages 1081--1093, Miami, Florida, USA. Association for Computational Linguistics.

\bibitem[{Xiao et~al.(2025)Xiao, Hou, Garcia-Romero, and Han}]{10890057}
Cihan Xiao, Zejiang Hou, Daniel Garcia-Romero, and Kyu~J Han. 2025.
\newblock \href {https://doi.org/10.1109/ICASSP49660.2025.10890057} {Contextual asr with retrieval augmented large language model}.
\newblock In \emph{ICASSP 2025 - 2025 IEEE International Conference on Acoustics, Speech and Signal Processing (ICASSP)}, pages 1--5.

\bibitem[{Xu et~al.(2017)Xu, Zhao, Xiao, Wu, Zhang, He, and Zhuang}]{Xu2017VideoQA}
D.~Xu, Zhou Zhao, Jun Xiao, Fei Wu, Hanwang Zhang, Xiangnan He, and Yueting Zhuang. 2017.
\newblock \href {https://api.semanticscholar.org/CorpusID:3864050} {Video question answering via gradually refined attention over appearance and motion}.
\newblock \emph{Proceedings of the 25th ACM international conference on Multimedia}.

\bibitem[{Xu et~al.(2023)Xu, Ye, Wu, Yan, Miao, Ye, Xu, Hu, Shi, Xu, Li, Qian, Que, Zhang, Zeng, and Huang}]{xu2023youkumplug10millionlargescale}
Haiyang Xu, Qinghao Ye, Xuan Wu, Ming Yan, Yuan Miao, Jiabo Ye, Guohai Xu, Anwen Hu, Yaya Shi, Guangwei Xu, Chenliang Li, Qi~Qian, Maofei Que, Ji~Zhang, Xiao Zeng, and Fei Huang. 2023.
\newblock \href {https://arxiv.org/abs/2306.04362} {Youku-mplug: A 10 million large-scale chinese video-language dataset for pre-training and benchmarks}.
\newblock \emph{Preprint}, arXiv:2306.04362.

\bibitem[{Xu et~al.(2024{\natexlab{a}})Xu, Huang, Hou, Chen, Zhang, Feng, and Xie}]{xu2024retrievalaugmentedegocentricvideocaptioning}
Jilan Xu, Yifei Huang, Junlin Hou, Guo Chen, Yuejie Zhang, Rui Feng, and Weidi Xie. 2024{\natexlab{a}}.
\newblock Retrieval-augmented egocentric video captioning.
\newblock In \emph{Proceedings of the IEEE/CVF Conference on Computer Vision and Pattern Recognition}, pages 13525--13536.

\bibitem[{Xu et~al.(2024{\natexlab{b}})Xu, Jain, and Kankanhalli}]{xu2024hallucinationinevitableinnatelimitation}
Ziwei Xu, Sanjay Jain, and Mohan Kankanhalli. 2024{\natexlab{b}}.
\newblock \href {https://arxiv.org/abs/2401.11817} {Hallucination is inevitable: An innate limitation of large language models}.
\newblock \emph{Preprint}, arXiv:2401.11817.

\bibitem[{Xue et~al.(2024{\natexlab{a}})Xue, Deng, Gao, and Li}]{xue2024retrievalaugmentedgenerationpromptbased}
Jinlong Xue, Yayue Deng, Yingming Gao, and Ya~Li. 2024{\natexlab{a}}.
\newblock \href {https://arxiv.org/abs/2406.03714} {Retrieval augmented generation in prompt-based text-to-speech synthesis with context-aware contrastive language-audio pretraining}.
\newblock \emph{Preprint}, arXiv:2406.03714.

\bibitem[{Xue et~al.(2024{\natexlab{b}})Xue, Deng, Yu, Wang, Wang, and Li}]{xue2024enhancedmultimodalragllmaccurate}
Junxiao Xue, Quan Deng, Fei Yu, Yanhao Wang, Jun Wang, and Yuehua Li. 2024{\natexlab{b}}.
\newblock \href {https://arxiv.org/abs/2412.20927} {Enhanced multimodal rag-llm for accurate visual question answering}.
\newblock \emph{Preprint}, arXiv:2412.20927.

\bibitem[{Yan et~al.(2024)Yan, Gu, Zhu, and Ling}]{yan2024corrective}
Shi-Qi Yan, Jia-Chen Gu, Yun Zhu, and Zhen-Hua Ling. 2024.
\newblock \href {https://openreview.net/forum?id=JnWJbrnaUE} {Corrective retrieval augmented generation}.

\bibitem[{Yan and Xie(2024)}]{Yan_2024}
Yibin Yan and Weidi Xie. 2024.
\newblock \href {https://doi.org/10.18653/v1/2024.findings-emnlp.83} {Echosight: Advancing visual-language models with wiki knowledge}.
\newblock In \emph{Findings of the Association for Computational Linguistics: EMNLP 2024}, pages 1538--1551, Miami, Florida, USA. Association for Computational Linguistics.

\bibitem[{Yang et~al.(2024{\natexlab{a}})Yang, Shi, Le, Hsu, and Tjandra}]{yang2024audioboxttaragimprovingzeroshot}
Mu~Yang, Bowen Shi, Matthew Le, Wei-Ning Hsu, and Andros Tjandra. 2024{\natexlab{a}}.
\newblock \href {https://arxiv.org/abs/2411.05141} {Audiobox tta-rag: Improving zero-shot and few-shot text-to-audio with retrieval-augmented generation}.
\newblock \emph{Preprint}, arXiv:2411.05141.

\bibitem[{Yang et~al.(2023)Yang, Chen, Wang, Hu, and Zhang}]{SKURG}
Qian Yang, Qian Chen, Wen Wang, Baotian Hu, and Min Zhang. 2023.
\newblock \href {https://doi.org/10.1145/3581783.3611964} {Enhancing multi-modal multi-hop question answering via structured knowledge and unified retrieval-generation}.
\newblock In \emph{Proceedings of the 31st ACM International Conference on Multimedia}, MM '23, page 5223–5234, New York, NY, USA. Association for Computing Machinery.

\bibitem[{Yang et~al.(2025)Yang, Fu, Wang, Wang, Song, and Bian}]{yang2025omgm}
Wei Yang, Jingjing Fu, Rui Wang, Jinyu Wang, Lei Song, and Jiang Bian. 2025.
\newblock Omgm: Orchestrate multiple granularities and modalities for efficient multimodal retrieval.
\newblock \emph{arXiv preprint arXiv:2505.07879}.

\bibitem[{Yang et~al.(2024{\natexlab{b}})Yang, Xue, Qian, Dong, and Xu}]{10.1145/3626772.3657740}
Zhenyu Yang, Dizhan Xue, Shengsheng Qian, Weiming Dong, and Changsheng Xu. 2024{\natexlab{b}}.
\newblock \href {https://doi.org/10.1145/3626772.3657740} {Ldre: Llm-based divergent reasoning and ensemble for zero-shot composed image retrieval}.
\newblock In \emph{Proceedings of the 47th International ACM SIGIR Conference on Research and Development in Information Retrieval}, SIGIR '24, page 80–90, New York, NY, USA. Association for Computing Machinery.

\bibitem[{Yao et~al.(2023)Yao, Shah, Sun, Cho, and Huang}]{yao2023end}
Barry~Menglong Yao, Aditya Shah, Lichao Sun, Jin-Hee Cho, and Lifu Huang. 2023.
\newblock End-to-end multimodal fact-checking and explanation generation: A challenging dataset and models.
\newblock In \emph{Proceedings of the 46th International ACM SIGIR Conference on Research and Development in Information Retrieval}, pages 2733--2743.

\bibitem[{Yasunaga et~al.(2023)Yasunaga, Aghajanyan, Shi, James, Leskovec, Liang, Lewis, Zettlemoyer, and Yih}]{yasunaga2023retrievalaugmentedmultimodallanguagemodeling}
Michihiro Yasunaga, Armen Aghajanyan, Weijia Shi, Richard James, Jure Leskovec, Percy Liang, Mike Lewis, Luke Zettlemoyer, and Wen-Tau Yih. 2023.
\newblock Retrieval-augmented multimodal language modeling.
\newblock In \emph{International Conference on Machine Learning}, pages 39755--39769. PMLR.

\bibitem[{Ye et~al.(2019)Ye, Rochan, Liu, and Wang}]{ye2019cross}
Linwei Ye, Mrigank Rochan, Zhi Liu, and Yang Wang. 2019.
\newblock Cross-modal self-attention network for referring image segmentation.
\newblock In \emph{Proceedings of the IEEE/CVF conference on computer vision and pattern recognition}, pages 10502--10511.

\bibitem[{Yeo et~al.(2025)Yeo, Kim, Jeong, Baek, and Hwang}]{yeo2025universalrag}
Woongyeong Yeo, Kangsan Kim, Soyeong Jeong, Jinheon Baek, and Sung~Ju Hwang. 2025.
\newblock Universalrag: Retrieval-augmented generation over multiple corpora with diverse modalities and granularities.
\newblock \emph{arXiv preprint arXiv:2504.20734}.

\bibitem[{Yi et~al.(2025)Yi, Xiao, and Albert}]{yi2025multimodalmultiagentframeworkradiology}
Ziruo Yi, Ting Xiao, and Mark~V. Albert. 2025.
\newblock \href {https://arxiv.org/abs/2505.09787} {A multimodal multi-agent framework for radiology report generation}.
\newblock \emph{Preprint}, arXiv:2505.09787.

\bibitem[{Young et~al.(2014)Young, Lai, Hodosh, and Hockenmaier}]{young2014image}
Peter Young, Alice Lai, Micah Hodosh, and Julia Hockenmaier. 2014.
\newblock From image descriptions to visual denotations: New similarity metrics for semantic inference over event descriptions.
\newblock \emph{Transactions of the Association for Computational Linguistics}, 2:67--78.

\bibitem[{Yu et~al.(2024)Yu, Tang, Xu, Cui, Ran, Yan, Liu, Wang, Han, Liu, and Sun}]{yu2024visragvisionbasedretrievalaugmentedgeneration}
Shi Yu, Chaoyue Tang, Bokai Xu, Junbo Cui, Junhao Ran, Yukun Yan, Zhenghao Liu, Shuo Wang, Xu~Han, Zhiyuan Liu, and Maosong Sun. 2024.
\newblock \href {https://arxiv.org/abs/2410.10594} {Visrag: Vision-based retrieval-augmented generation on multi-modality documents}.
\newblock \emph{Preprint}, arXiv:2410.10594.

\bibitem[{Yu et~al.(2025)Yu, Yang, and Chen}]{yu2025unveilingpotentialmultimodalretrieval}
Xiaohan Yu, Zhihan Yang, and Chong Chen. 2025.
\newblock \href {https://arxiv.org/abs/2501.15470} {Unveiling the potential of multimodal retrieval augmented generation with planning}.
\newblock \emph{Preprint}, arXiv:2501.15470.

\bibitem[{Yu et~al.(2019)Yu, Xu, Yu, Yu, Zhao, Zhuang, and Tao}]{yu2019activitynet}
Zhou Yu, Dejing Xu, Jun Yu, Ting Yu, Zhou Zhao, Yueting Zhuang, and Dacheng Tao. 2019.
\newblock Activitynet-qa: A dataset for understanding complex web videos via question answering.
\newblock In \emph{Proceedings of the AAAI Conference on Artificial Intelligence}, volume~33, pages 9127--9134.

\bibitem[{Yuan et~al.(2024)Yuan, Sun, Omeiza, Zhao, Newman, Kunze, and Gadd}]{yuan2024ragdrivergeneralisabledrivingexplanations}
Jianhao Yuan, Shuyang Sun, Daniel Omeiza, Bo~Zhao, Paul Newman, Lars Kunze, and Matthew Gadd. 2024.
\newblock \href {https://arxiv.org/abs/2402.10828} {Rag-driver: Generalisable driving explanations with retrieval-augmented in-context learning in multi-modal large language model}.
\newblock \emph{Preprint}, arXiv:2402.10828.

\bibitem[{Yuan et~al.(2023)Yuan, Jin, Tan, Zhao, Yuan, Huang, and Huang}]{yuan2023rammretrievalaugmentedbiomedicalvisual}
Zheng Yuan, Qiao Jin, Chuanqi Tan, Zhengyun Zhao, Hongyi Yuan, Fei Huang, and Songfang Huang. 2023.
\newblock Ramm: Retrieval-augmented biomedical visual question answering with multi-modal pre-training.
\newblock In \emph{Proceedings of the 31st ACM International Conference on Multimedia}, pages 547--556.

\bibitem[{Zhai et~al.(2023)Zhai, Gong, Wang, Sun, Yan, Li, and Liu}]{10.1145/3580305.3599897}
Jiaqi Zhai, Zhaojie Gong, Yueming Wang, Xiao Sun, Zheng Yan, Fu~Li, and Xing Liu. 2023.
\newblock \href {https://doi.org/10.1145/3580305.3599897} {Revisiting neural retrieval on accelerators}.
\newblock In \emph{Proceedings of the 29th ACM SIGKDD Conference on Knowledge Discovery and Data Mining}, KDD '23, page 5520–5531, New York, NY, USA. Association for Computing Machinery.

\bibitem[{Zhai(2024)}]{zhai2024selfadaptivemultimodalretrievalaugmentedgeneration}
Wenjia Zhai. 2024.
\newblock \href {https://arxiv.org/abs/2410.11321} {Self-adaptive multimodal retrieval-augmented generation}.
\newblock \emph{Preprint}, arXiv:2410.11321.

\bibitem[{Zhang et~al.(2024{\natexlab{a}})Zhang, Liu, Zhu, Zeng, Sheng, Yang, Dai, and Wang}]{zhang2024efficienteffectiveretrievaldensesparse}
Haoyu Zhang, Jun Liu, Zhenhua Zhu, Shulin Zeng, Maojia Sheng, Tao Yang, Guohao Dai, and Yu~Wang. 2024{\natexlab{a}}.
\newblock \href {https://arxiv.org/abs/2410.20381} {Efficient and effective retrieval of dense-sparse hybrid vectors using graph-based approximate nearest neighbor search}.
\newblock \emph{Preprint}, arXiv:2410.20381.

\bibitem[{Zhang et~al.(2023{\natexlab{a}})Zhang, Lian, Zhang, Wang, and Chen}]{Zhang_Lian_Zhang_Wang_Chen_2023}
Jin Zhang, Defu Lian, Haodi Zhang, Baoyun Wang, and Enhong Chen. 2023{\natexlab{a}}.
\newblock \href {https://doi.org/10.1609/aaai.v37i4.25613} {Query-aware quantization for maximum inner product search}.
\newblock \emph{Proceedings of the AAAI Conference on Artificial Intelligence}, 37(4):4875--4883.

\bibitem[{Zhang et~al.(2024{\natexlab{b}})Zhang, Yu, and Zhang}]{10.1145/3664647.3680750}
Jinxu Zhang, Yongqi Yu, and Yu~Zhang. 2024{\natexlab{b}}.
\newblock \href {https://doi.org/10.1145/3664647.3680750} {Cream: Coarse-to-fine retrieval and multi-modal efficient tuning for document vqa}.
\newblock In \emph{Proceedings of the 32nd ACM International Conference on Multimedia}, MM '24, page 925–934, New York, NY, USA. Association for Computing Machinery.

\bibitem[{Zhang et~al.(2024{\natexlab{c}})Zhang, Liu, Tai, and Tang}]{zhang2024c3net}
Juntao Zhang, Yuehuai Liu, Yu-Wing Tai, and Chi-Keung Tang. 2024{\natexlab{c}}.
\newblock C3net: Compound conditioned controlnet for multimodal content generation.
\newblock In \emph{Proceedings of the IEEE/CVF Conference on Computer Vision and Pattern Recognition}, pages 26886--26895.

\bibitem[{Zhang et~al.(2024{\natexlab{d}})Zhang, Zhang, Wang, Ouyang, Wen, Li, Chow, He, and Zhang}]{zhang2024ocr}
Junyuan Zhang, Qintong Zhang, Bin Wang, Linke Ouyang, Zichen Wen, Ying Li, Ka-Ho Chow, Conghui He, and Wentao Zhang. 2024{\natexlab{d}}.
\newblock Ocr hinders rag: Evaluating the cascading impact of ocr on retrieval-augmented generation.
\newblock \emph{arXiv preprint arXiv:2412.02592}.

\bibitem[{Zhang et~al.(2024{\natexlab{e}})Zhang, Zhao, Ying, Ma, and Lee}]{zhang-etal-2024-omagent}
Lu~Zhang, Tiancheng Zhao, Heting Ying, Yibo Ma, and Kyusong Lee. 2024{\natexlab{e}}.
\newblock \href {https://doi.org/10.18653/v1/2024.emnlp-main.559} {{O}m{A}gent: A multi-modal agent framework for complex video understanding with task divide-and-conquer}.
\newblock In \emph{Proceedings of the 2024 Conference on Empirical Methods in Natural Language Processing}, pages 10031--10045, Miami, Florida, USA. Association for Computational Linguistics.

\bibitem[{Zhang et~al.(2024{\natexlab{f}})Zhang, Zhang, Ma, Chen, Qi, Yuan, Li, Pu, Zhao, Xie, Ma, Shan, and Hu}]{Zhang2024mR2AGMR}
Tao Zhang, Ziqi Zhang, Zongyang Ma, Yuxin Chen, Zhongang Qi, Chunfen Yuan, Bing Li, Junfu Pu, Yuxuan Zhao, Zehua Xie, Jin Ma, Ying Shan, and Weiming Hu. 2024{\natexlab{f}}.
\newblock \href {https://api.semanticscholar.org/CorpusID:274192536} {mr2ag: Multimodal retrieval-reflection-augmented generation for knowledge-based vqa}.
\newblock \emph{ArXiv}, abs/2411.15041.

\bibitem[{Zhang et~al.(2024{\natexlab{g}})Zhang, Zhang, Ma, Chen, Qi, Yuan, Li, Pu, Zhao, Xie, Ma, Shan, and Hu}]{zhang2024mr2agmultimodalretrievalreflectionaugmentedgeneration}
Tao Zhang, Ziqi Zhang, Zongyang Ma, Yuxin Chen, Zhongang Qi, Chunfeng Yuan, Bing Li, Junfu Pu, Yuxuan Zhao, Zehua Xie, Jin Ma, Ying Shan, and Weiming Hu. 2024{\natexlab{g}}.
\newblock \href {https://arxiv.org/abs/2411.15041} {mr$^2$ag: Multimodal retrieval-reflection-augmented generation for knowledge-based vqa}.
\newblock \emph{Preprint}, arXiv:2411.15041.

\bibitem[{Zhang et~al.(2024{\natexlab{h}})Zhang, Patil, Jain, Shen, Zaharia, Stoica, and Gonzalez}]{zhang2024raft}
Tianjun Zhang, Shishir~G Patil, Naman Jain, Sheng Shen, Matei Zaharia, Ion Stoica, and Joseph~E. Gonzalez. 2024{\natexlab{h}}.
\newblock \href {https://openreview.net/forum?id=rzQGHXNReU} {{RAFT}: Adapting language model to domain specific {RAG}}.
\newblock In \emph{First Conference on Language Modeling}.

\bibitem[{Zhang et~al.(2020)Zhang, Kishore, Wu, Weinberger, and Artzi}]{zhang2020bertscoreevaluatingtextgeneration}
Tianyi Zhang, Varsha Kishore, Felix Wu, Kilian~Q. Weinberger, and Yoav Artzi. 2020.
\newblock \href {https://arxiv.org/abs/1904.09675} {Bertscore: Evaluating text generation with bert}.
\newblock \emph{Preprint}, arXiv:1904.09675.

\bibitem[{Zhang et~al.(2024{\natexlab{i}})Zhang, Zhang, Xie, Li, Dai, Long, Xie, Zhang, Li, and Zhang}]{zhang2024gmeimprovinguniversalmultimodal}
Xin Zhang, Yanzhao Zhang, Wen Xie, Mingxin Li, Ziqi Dai, Dingkun Long, Pengjun Xie, Meishan Zhang, Wenjie Li, and Min Zhang. 2024{\natexlab{i}}.
\newblock \href {https://arxiv.org/abs/2412.16855} {Gme: Improving universal multimodal retrieval by multimodal llms}.
\newblock \emph{Preprint}, arXiv:2412.16855.

\bibitem[{Zhang et~al.(2023{\natexlab{b}})Zhang, Zhang, Li, Zhao, Karypis, and Smola}]{zhang2023multicot}
Zhuosheng Zhang, Aston Zhang, Mu~Li, Hai Zhao, George Karypis, and Alex Smola. 2023{\natexlab{b}}.
\newblock Multimodal chain-of-thought reasoning in language models.
\newblock \emph{arXiv preprint arXiv:2302.00923}.

\bibitem[{Zhao et~al.(2023{\natexlab{a}})Zhao, Chen, Wang, Jiao, Do, Qin, Ding, Guo, Li, Li, and Joty}]{zhao-etal-2023-retrieving}
Ruochen Zhao, Hailin Chen, Weishi Wang, Fangkai Jiao, Xuan~Long Do, Chengwei Qin, Bosheng Ding, Xiaobao Guo, Minzhi Li, Xingxuan Li, and Shafiq Joty. 2023{\natexlab{a}}.
\newblock \href {https://doi.org/10.18653/v1/2023.findings-emnlp.314} {Retrieving multimodal information for augmented generation: A survey}.
\newblock In \emph{Findings of the Association for Computational Linguistics: EMNLP 2023}, pages 4736--4756, Singapore. Association for Computational Linguistics.

\bibitem[{Zhao et~al.(2023{\natexlab{b}})Zhao, Zheng, Yi, Luan, Xie, Zhou, and Jensen}]{10.14778/3579075.3579084}
Xi~Zhao, Bolong Zheng, Xiaomeng Yi, Xiaofan Luan, Charles Xie, Xiaofang Zhou, and Christian~S. Jensen. 2023{\natexlab{b}}.
\newblock \href {https://doi.org/10.14778/3579075.3579084} {Fargo: Fast maximum inner product search via global multi-probing}.
\newblock \emph{Proc. VLDB Endow.}, 16(5):1100–1112.

\bibitem[{Zhao et~al.(2024)Zhao, Zhang, Zhang, and Wu}]{zhao-etal-2024-unifashion}
Xiangyu Zhao, Yuehan Zhang, Wenlong Zhang, and Xiao-Ming Wu. 2024.
\newblock \href {https://doi.org/10.18653/v1/2024.emnlp-main.89} {Unifashion: A unified vision-language model for multimodal fashion retrieval and generation}.
\newblock In \emph{Proceedings of the 2024 Conference on Empirical Methods in Natural Language Processing}, pages 1490--1507, Miami, Florida, USA. Association for Computational Linguistics.

\bibitem[{Zheng et~al.(2023)Zheng, Lyu, Gao, Dai, and Song}]{10204116}
Chaofan Zheng, Xinyu Lyu, Lianli Gao, Bo~Dai, and Jingkuan Song. 2023.
\newblock \href {https://doi.org/10.1109/CVPR52729.2023.02182} {{ Prototype-Based Embedding Network for Scene Graph Generation }}.
\newblock In \emph{2023 IEEE/CVF Conference on Computer Vision and Pattern Recognition (CVPR)}, pages 22783--22792, Los Alamitos, CA, USA. IEEE Computer Society.

\bibitem[{Zhi~Lim et~al.(2024)Zhi~Lim, Poo~Lee, Ming~Lim, and Kamsani~Samingan}]{10535103}
Qi~Zhi~Lim, Chin Poo~Lee, Kian Ming~Lim, and Ahmad Kamsani~Samingan. 2024.
\newblock \href {https://doi.org/10.1109/ACCESS.2024.3403101} {Unirag: Unification, retrieval, and generation for multimodal question answering with pre-trained language models}.
\newblock \emph{IEEE Access}, 12:71505--71519.

\bibitem[{Zhong et~al.(2024)Zhong, Lang, Zhang, Cheng, Zhang, and Zhou}]{10.1145/3626772.3657929}
Ting Zhong, Jian Lang, Yifan Zhang, Zhangtao Cheng, Kunpeng Zhang, and Fan Zhou. 2024.
\newblock \href {https://doi.org/10.1145/3626772.3657929} {Predicting micro-video popularity via multi-modal retrieval augmentation}.
\newblock In \emph{Proceedings of the 47th International ACM SIGIR Conference on Research and Development in Information Retrieval}, SIGIR '24, page 2579–2583, New York, NY, USA. Association for Computing Machinery.

\bibitem[{Zhong et~al.(2019)Zhong, Tang, and Yepes}]{zhong2019publaynetlargestdatasetdocument}
Xu~Zhong, Jianbin Tang, and Antonio~Jimeno Yepes. 2019.
\newblock Publaynet: largest dataset ever for document layout analysis.
\newblock In \emph{2019 International conference on document analysis and recognition (ICDAR)}, pages 1015--1022. IEEE.

\bibitem[{Zhou et~al.(2024{\natexlab{a}})Zhou, Liu, Liu, Xiao, Wang, Zhao, Zhang, Lian, and Xiong}]{zhou2024megapairsmassivedatasynthesis}
Junjie Zhou, Zheng Liu, Ze~Liu, Shitao Xiao, Yueze Wang, Bo~Zhao, Chen~Jason Zhang, Defu Lian, and Yongping Xiong. 2024{\natexlab{a}}.
\newblock \href {https://arxiv.org/abs/2412.14475} {Megapairs: Massive data synthesis for universal multimodal retrieval}.
\newblock \emph{Preprint}, arXiv:2412.14475.

\bibitem[{Zhou et~al.(2024{\natexlab{b}})Zhou, Liu, Xiao, Zhao, and Xiong}]{zhou-etal-2024-vista}
Junjie Zhou, Zheng Liu, Shitao Xiao, Bo~Zhao, and Yongping Xiong. 2024{\natexlab{b}}.
\newblock \href {https://doi.org/10.18653/v1/2024.acl-long.175} {{VISTA}: Visualized text embedding for universal multi-modal retrieval}.
\newblock In \emph{Proceedings of the 62nd Annual Meeting of the Association for Computational Linguistics (Volume 1: Long Papers)}, pages 3185--3200, Bangkok, Thailand. Association for Computational Linguistics.

\bibitem[{Zhou et~al.(2018)Zhou, Xu, and Corso}]{zhou2018youcook2}
Luowei Zhou, Chenliang Xu, and Jason~J Corso. 2018.
\newblock Towards automatic learning of procedures from web instructional videos.
\newblock \emph{Proceedings of the AAAI Conference on Artificial Intelligence}, 32(1):1--10.

\bibitem[{Zhou(2024)}]{Zhou_2024}
Ren Zhou. 2024.
\newblock \href {https://doi.org/10.54097/h8wf8vah} {Advanced embedding techniques in multimodal retrieval augmented generation a comprehensive study on cross modal ai applications}.
\newblock \emph{Journal of Computing and Electronic Information Management}, 13(3):16–22.

\bibitem[{Zhou et~al.(2023)Zhou, Alon, Xu, Jiang, and Neubig}]{zhou2023docprompting}
Shuyan Zhou, Uri Alon, Frank~F. Xu, Zhengbao Jiang, and Graham Neubig. 2023.
\newblock \href {https://openreview.net/forum?id=ZTCxT2t2Ru} {Docprompting: Generating code by retrieving the docs}.
\newblock In \emph{The Eleventh International Conference on Learning Representations}.

\bibitem[{Zhou et~al.(2024{\natexlab{c}})Zhou, Mei, Li, Liu, Xiong, Liu, Gu, and Yu}]{zhou2024marvelunlockingmultimodalcapability}
Tianshuo Zhou, Sen Mei, Xinze Li, Zhenghao Liu, Chenyan Xiong, Zhiyuan Liu, Yu~Gu, and Ge~Yu. 2024{\natexlab{c}}.
\newblock Marvel: unlocking the multi-modal capability of dense retrieval via visual module plugin.
\newblock In \emph{Proceedings of the 62nd Annual Meeting of the Association for Computational Linguistics (Volume 1: Long Papers)}, pages 14608--14624.

\bibitem[{Zhou et~al.(2024{\natexlab{d}})Zhou, Liu, Li, Jin, Qian, Liu, Li, Dou, Ho, and Yu}]{zhou2024TrustworthyRAG}
Yujia Zhou, Yan Liu, Xiaoxi Li, Jiajie Jin, Hongjin Qian, Zheng Liu, Chaozhuo Li, Zhicheng Dou, Tsung-Yi Ho, and Philip~S. Yu. 2024{\natexlab{d}}.
\newblock Trustworthiness in retrieval-augmented generation systems: A survey.
\newblock volume abs/2409.10102.

\bibitem[{Zhou et~al.(2024{\natexlab{e}})Zhou, Zhang, Guan, Hu, Lao, Mu, Li, and Mai}]{zhou2024img2loc}
Zhongliang Zhou, Jielu Zhang, Zihan Guan, Mengxuan Hu, Ni~Lao, Lan Mu, Sheng Li, and Gengchen Mai. 2024{\natexlab{e}}.
\newblock Img2loc: Revisiting image geolocalization using multi-modality foundation models and image-based retrieval-augmented generation.
\newblock In \emph{Proceedings of the 47th International ACM SIGIR Conference on Research and Development in Information Retrieval}, pages 2749--2754.

\bibitem[{Zhu et~al.(2024{\natexlab{a}})Zhu, Liu, Dong, Xu, Huang, Kong, Chen, and Li}]{zhu-etal-2024-multilingual}
Wenhao Zhu, Hongyi Liu, Qingxiu Dong, Jingjing Xu, Shujian Huang, Lingpeng Kong, Jiajun Chen, and Lei Li. 2024{\natexlab{a}}.
\newblock \href {https://doi.org/10.18653/v1/2024.findings-naacl.176} {Multilingual machine translation with large language models: Empirical results and analysis}.
\newblock In \emph{Findings of the Association for Computational Linguistics: NAACL 2024}, pages 2765--2781, Mexico City, Mexico. Association for Computational Linguistics.

\bibitem[{Zhu et~al.(2024{\natexlab{b}})Zhu, Ren, Wang, Zheng, Xie, Feng, Zhu, Li, Ma, and Pan}]{zhu2024emergeintegratingragimproved}
Yinghao Zhu, Changyu Ren, Zixiang Wang, Xiaochen Zheng, Shiyun Xie, Junlan Feng, Xi~Zhu, Zhoujun Li, Liantao Ma, and Chengwei Pan. 2024{\natexlab{b}}.
\newblock \href {https://doi.org/10.1145/3627673.3679582} {Emerge: Enhancing multimodal electronic health records predictive modeling with retrieval-augmented generation}.
\newblock In \emph{Proceedings of the 33rd ACM International Conference on Information and Knowledge Management}, CIKM '24, page 3549–3559, New York, NY, USA. Association for Computing Machinery.

\bibitem[{Zhu et~al.(2024{\natexlab{c}})Zhu, Ren, Xie, Liu, Ji, Wang, Sun, He, Li, Zhu, and Pan}]{zhu2024realmragdrivenenhancementmultimodal}
Yinghao Zhu, Changyu Ren, Shiyun Xie, Shukai Liu, Hangyuan Ji, Zixiang Wang, Tao Sun, Long He, Zhoujun Li, Xi~Zhu, and Chengwei Pan. 2024{\natexlab{c}}.
\newblock \href {https://arxiv.org/abs/2402.07016} {Realm: Rag-driven enhancement of multimodal electronic health records analysis via large language models}.
\newblock \emph{Preprint}, arXiv:2402.07016.

\bibitem[{Zhu et~al.(2025)Zhu, Lee, Zhang, Harsha, Feujio, Maharaj, and Li}]{zhu2024murarsimpleeffectivemultimodal}
Zhengyuan Zhu, Daniel Lee, Hong Zhang, Sai~Sree Harsha, Loic Feujio, Akash Maharaj, and Yunyao Li. 2025.
\newblock Murar: A simple and effective multimodal retrieval and answer refinement framework for multimodal question answering.
\newblock In \emph{Proceedings of the 31st International Conference on Computational Linguistics: System Demonstrations}, pages 126--135.

\bibitem[{Zuo et~al.(2024)Zuo, Zhou, Nie, Zhang, Guo, Sang, Wang, and Gao}]{zuo2024ufinebench}
Jialong Zuo, Hanyu Zhou, Ying Nie, Feng Zhang, Tianyu Guo, Nong Sang, Yunhe Wang, and Changxin Gao. 2024.
\newblock Ufinebench: Towards text-based person retrieval with ultra-fine granularity.
\newblock In \emph{Proceedings of the IEEE/CVF Conference on Computer Vision and Pattern Recognition}, pages 22010--22019.

\end{thebibliography}

\newpage
\appendix

\section{Taxonomy}
\label{sec:taxonomy_details}
\noindent
In this section, we provide more details regarding the taxonomy of multimodal RAG systems, previously mentioned in \autoref{fig:taxonomy_full}. Additionally, we present a classification of multimodal RAG application domains in \autoref{fig:app_taxonomy}.

\noindent
\autoref{fig:taxonomy_full} provides an overview of recent advances in multimodal RAG systems. The taxonomy is organized into several key categories.
\begin{itemize} \item \textbf{Retrieval strategies} cover efficient search and similarity retrieval methods (including maximum inner product search (MIPS) variants and different multimodal encoders) and modality-centric techniques that distinguish between text-, vision-, audio-, and video-centric as well as document retrieval models. Re-ranking strategies further refine these methods via optimized example selection, relevance scoring, and filtering.
\item \textbf{Fusion mechanisms} cover score fusion and alignment techniques, including CLIP score fusion and prototype-based embeddings that unify multimodal representations, attention-based methods such as cross-attention and co-attention transformers that dynamically weight cross-modal interactions, and unified frameworks and projections like hierarchical fusion and dense-to-sparse projections that consolidate multimodal inputs. \item \textbf{Augmentation techniques} address context enrichment as well as adaptive and iterative retrieval. \item \textbf{Generation methods} include in-context learning, reasoning, instruction tuning, source attribution, and agentic frameworks. \item \textbf{training strategies} are characterized by approaches to alignment and robustness.
\end{itemize}
Detailed discussions of these categories are provided in the corresponding sections.

\noindent
\autoref{fig:app_taxonomy} presents the taxonomy of application domains for multimodal RAG systems. The identified domains include \emph{healthcare and medicine}, \emph{software engineering}, \emph{fashion and e-commerce}, \emph{entertainment and social computing}, and \emph{emerging applications}. This classification offers a concise overview of the diverse applications and serves as a framework for the more detailed analyses that follow.

\section{Dataset and Benchmark}
\label{sec:app_dataset}
\noindent
Multimodal RAG research employs diverse datasets and benchmarks to evaluate retrieval, integration, and generation across heterogeneous sources. Image–text tasks, including captioning and retrieval, commonly use MS-COCO \cite{lin2014microsoft}, Flickr30K \cite{young2014image}, and LAION-400M \cite{laion400m}, while visual question answering (QA) with external knowledge is supported by OK-VQA \cite{marino2019okvqa} and WebQA \cite{Chang_2022_CVPR}. For complex multimodal reasoning, MultimodalQA \cite{talmor2021multimodalqa} integrates text, images, and tables, whereas video-text tasks leverage ActivityNet \cite{caba2015activitynet} and YouCook2 \cite{zhou2018youcook2}. In the medical domain, MIMIC-CXR \cite{johnson2019mimic} and CheXpert \cite{irvin2019chexpert} facilitate tasks such as medical report generation. It should be noted that a number of these datasets are unimodal (e.g., solely text-based or image-based). Unimodal datasets are frequently employed to represent a specific modality and are subsequently integrated with complementary datasets from other modalities. This modular approach allows each dataset to contribute its domain-specific strengths, thereby enhancing the overall performance of the multimodal retrieval and generation processes. 

\noindent
Benchmarks assess multimodal RAG systems on visual reasoning, external knowledge integration, and dynamic retrieval. The $M^{2}RAG$ \cite{ma2024multimodalretrievalaugmentedmultimodal} benchmark provides a unified evaluation framework that combines fine-grained text-modal and multimodal metrics to jointly assess both the quality of generated language and the effective integration of visual elements. In addition, \cite{liu2025benchmarking} introduce another specialized benchmark for multimodal RAG that evaluates performance across image captioning, multi-modal question answering, fact verification, and image reranking in an open-domain retrieval setting. Vision-focused evaluations, including MRAG-Bench \cite{hu2024mragbench}, VQAv2 \cite{balanced_vqa_v2} and VisDoMBench \cite{suri2024visdommultidocumentqavisually}, test models on complex visual tasks. Dyn-VQA \cite{li2024benchmarking}, MMBench \cite{liu2025mmbench}, and ScienceQA \cite{lu2022learn} evaluate dynamic retrieval and multi-hop reasoning across textual, visual, and diagrammatic inputs. Knowledge-intensive benchmarks, such as TriviaQA \cite{joshi2017triviaqa} and Natural Questions \cite{kwiatkowski2019natural}, together with document-oriented evaluations such as OmniDocBench \cite{ouyang2024omnidocbenchbenchmarkingdiversepdf}, measure integration of unstructured and structured data. Advanced retrieval benchmarks such as RAG-Check \cite{mortaheb2025ragcheckevaluatingmultimodalretrieval} evaluate retrieval relevance and system reliability, while specialized assessments like Counterfactual VQA \cite{niu2021counterfactual} test robustness against adversarial inputs. Additionally, OCR impact studies such as OHRBench \cite{zhang2024ocr} examine the cascading effects of errors on RAG systems.

\noindent
The choice of dataset significantly influences the evaluation focus, ranging from foundational pre-training on large-scale image-text corpora like LAION-5B \cite{schuhmann2022laion} (5.85 billion pairs) or MINT-1T \cite{awadalla2024mint} (3.4 billion images with 1 trillion text tokens), to more specialized tasks such as video understanding with HowTo100M \cite{Miech_2019_ICCV} (136 million video clips) or medical report generation using MIMIC-CXR \cite{johnson2019mimic} (125,417 image-report pairs).

\noindent
Datasets are often tailored for specific downstream tasks. For visual question answering, VQA \cite{antol2015vqa} and A-OKVQA \cite{schwenk2022aokvqabenchmarkvisualquestion} specifically require external knowledge, making them suitable for evaluating RAG systems' ability to retrieve and reason over such knowledge. For document understanding, datasets such as DocVQA \cite{mathew2021docvqa} and M3DocVQA \cite{cho2024m3docragmultimodalretrievalneed} are essential. As discussed in the benchmarks overview above, unified evaluation frameworks like $M^{2}RAG$ \cite{ma2024multimodalretrievalaugmentedmultimodal} provide a comprehensive assessment across multiple tasks, including image captioning, visual question answering, and fact verification.

\noindent
Evaluating complex reasoning capabilities in multimodal RAG systems has become increasingly important. Datasets such as MultimodalQA \cite{talmor2021multimodalqa}, WebQA \cite{Chang_2022_CVPR}, and ScienceQA \cite{lu2022learn} are specifically designed to benchmark multi-hop reasoning abilities crucial for advanced RAG systems, with Dyn-VQA \cite{li2024benchmarking} additionally focusing on robustness to changing information.

\paragraph{Comparative Analysis of Datasets} 
Understanding the strategic trade-offs in dataset design is crucial for multimodal RAG development, as different dataset characteristics serve distinct purposes across the model development pipeline.

\noindent\textbf{(i) Scale and Diversity vs. Curation:} Large-scale datasets such as LAION-5B \cite{schuhmann2022laion} and Conceptual Captions \cite{Sharma2018ConceptualCA} provide substantial scale essential for pre-training, enabling models to learn generalizable representations across diverse domains. However, their reliance on web-crawled data introduces inherent noise that can compromise training quality. Conversely, smaller, meticulously curated datasets like Flickr30K \cite{young2014image} (31,000 images with human annotations) and domain-specific collections such as Fashionpedia \cite{fashionpedia} (48,000 images with segmentation masks) prioritize annotation quality over scale, making them essential for fine-tuning models and assessing specialized performance.

\noindent\textbf{(ii) Modality Focus and Combination:} While many systems aggregate unimodal datasets to construct multimodal contexts, datasets explicitly designed for multimodal tasks demonstrate superior alignment between modalities. Foundational datasets like MS-COCO \cite{lin2014microsoft} and VQA \cite{antol2015vqa} establish benchmarks for image-text understanding, while specialized collections such as AudioSet \cite{gemmeke2017audioset} (2 million audio clips) and AudioCaps \cite{kim-etal-2019-audiocaps} (46,000 audio clips with captions) address audio-language integration. Emerging modalities like 3D (e.g., ShapeNet \cite{chang2015shapenet}) remain underrepresented, yet are essential for expanding RAG applications into spatial reasoning domains.

\noindent
\autoref{tab:categorized_datasets} and ~\autoref{tab:benchmark_datasets} present a comprehensive overview of datasets and benchmarks commonly employed in multimodal RAG research. The table is organized into five columns:
\begin{itemize} \item \textbf{Category:} This column categorizes each dataset or benchmark based on its primary domain or modality. The datasets are grouped into eight categories: \emph{Image–Text General}, \emph{Video–Text}, \emph{Audio–Text}, \emph{Medical}, \emph{Fashion}, \emph{3D}, \emph{Knowledge \& QA}, and \emph{Other}. The benchmarks are grouped into two categories: \emph{Cross-Modal Understanding} and \emph{Text-Focused}.
This classification facilitates a clearer understanding of each dataset or benchmark’s role within a multimodal framework. \item \textbf{Name:} The official name of the dataset or benchmarks is provided along with a citation for reference. \item \textbf{Statistics and Description:} This column summarizes key details such as dataset size, the nature of the content (e.g., image–text pairs, video captions, QA pairs), and the specific tasks or applications for which the dataset or benchmarks are used. These descriptions are intended to convey the dataset’s scope and its relevance to various multimodal RAG tasks. \item \textbf{Modalities:} The modalities covered by each dataset or benchmark are indicated (e.g., Image, Text, Video, Audio, or 3D). Notably, several datasets are unimodal; however, within multimodal RAG systems, these are combined with others to represent distinct aspects of a broader multimodal context. \item \textbf{Link:} A hyperlink is provided to direct readers to the official repository or additional resources for the dataset or benchmark, thereby facilitating further exploration of its properties and applications. \end{itemize}

\paragraph{Limitations of Existing Datasets and Benchmarks} While the datasets and benchmarks discussed above have significantly advanced multimodal RAG research, several limitations persist that offer important avenues for future work:

\noindent\textbf{(i) Bias and Fairness:} Large datasets, especially those scraped from the web, can inherit societal biases related to gender, race, or culture. This can lead to skewed model behavior and unfair outcomes. Efforts to create more balanced datasets are crucial, but comprehensive bias auditing across modalities remains a challenge. 

\noindent\textbf{(ii) Annotation Quality and Noise:} The trade-off between dataset scale and annotation quality remains a persistent challenge. While large datasets facilitate broad learning, their often noisy or weakly supervised labels (e.g., alt-text for images) can hinder precise model training. As demonstrated by OHRBench \cite{zhang2024ocr}, OCR errors exemplify how noise in one modality can cascade and affect overall RAG system performance.

\noindent\textbf{(iii) Coverage and Generalization Gaps:} Many datasets are domain-specific, which can limit the generalization of models to out-of-domain scenarios. There is a need for more datasets covering a wider array of real-world contexts and less common modalities. 

\noindent\textbf{(iv) Real-World Complexity and Long-Context Understanding:} Current datasets inadequately capture real-world multimodal information complexity. Challenges include efficient sampling of relevant video frames, handling multi-page documents with numerous images, and processing dynamic information environments; benchmarks like Dyn-VQA \cite{li2024benchmarking} are, however, beginning to address this latter challenge.

\noindent\textbf{(v) Lack of Adversarial and Robustness Testing:} While benchmarks like Counterfactual VQA \cite{niu2021counterfactual} specifically test robustness against certain perturbations, there is a general scarcity of datasets containing multimodal adversarial examples or structured negative instances. Such datasets are vital for developing more robust and reliable RAG systems that can handle out-of-distribution inputs or misleading information. 

\noindent\textbf{(vi) Retrieval-Generation Integration:} Many benchmarks evaluate retrieval and generation components separately rather than assessing their synergistic interplay. More holistic evaluation frameworks are needed that jointly measure retrieval accuracy, relevance of retrieved multimodal context, and final output quality, as aimed by benchmarks like MRAG-Bench \cite{hu2024mragbench} for visual integration and RAG-Check \cite{mortaheb2025ragcheckevaluatingmultimodalretrieval} for retrieval relevance. 

\noindent\textbf{(vii) Limited Support for "Any-to-Any" Modalities:} While current research primarily focuses on text, image, video, and audio, future RAG systems are envisioned to support any-to-any modality interactions. Existing datasets offer limited support for such comprehensive multimodality.

\begin{table*}[h]
\centering
\caption{Overview of Popular Datasets in Multimodal RAG Research.}
\label{tab:categorized_datasets}
\small
\renewcommand{\arraystretch}{1.5}
\resizebox{\textwidth}{!}{%
\begin{tabular}{c c l c l}
\toprule
\rotatebox[origin=c]{90}{\textbf{Category}} & \textbf{Name} & \multicolumn{1}{c}{\textbf{Statistics and Description}} & \textbf{Modalities} & \multicolumn{1}{c}{\textbf{Link}} \\ \midrule
\multirow{13}{*}{\rotatebox[origin=c]{90}{Image-Text General}} 
 & LAION-400M \cite{laion400m} & 400M image–text pairs; used for pre-training multimodal models. & Image, Text & \href{https://laion.ai/blog/laion-400-open-dataset/}{LAION-400M} \\
 & Conceptual-Captions (CC) \cite{Sharma2018ConceptualCA} & More than 3M image–caption pairs; multilingual English–German image descriptions. & Image, Text & \href{https://github.com/google-research-datasets/conceptual-captions}{Conceptual Captions} \\
 & CIRR \cite{Liu_2021_ICCV} & 36,554 triplets from 21,552 images; focuses on natural image relationships. & Image, Text & \href{https://github.com/Cuberick-Orion/CIRR}{CIRR} \\
 & MS-COCO \cite{lin2014microsoft} & 330K images with captions; used for caption–to–image and image–to–caption generation. & Image, Text & \href{https://cocodataset.org/}{MS-COCO} \\
 & Flickr30K \cite{young2014image} & 31K images annotated with five English captions per image. & Image, Text & \href{https://shannon.cs.illinois.edu/DenotationGraph/}{Flickr30K} \\
 & Multi30K \cite{elliott2016multi30k} & 30k German captions from native speakers and human–translated captions. & Image, Text & \href{https://github.com/multi30k/dataset}{Multi30K} \\
 & NoCaps \cite{nocaps} & For zero–shot image captioning evaluation; 15K images. & Image, Text & \href{https://nocaps.org/}{NoCaps} \\
 & Laion-5B \cite{schuhmann2022laion} & 5.85B image–text pairs used as external memory for retrieval. & Image, Text & \href{https://laion.ai/blog/laion-5b/}{LAION-5B} \\
 & COCO-CN \cite{cococn} & 20,341 images for cross-lingual tagging and captioning with Chinese sentences. & Image, Text & \href{https://github.com/li-xirong/coco-cn}{COCO-CN} \\
 & CIRCO \cite{circo} & 1,020 queries with an average of 4.53 ground truths per query; for composed image retrieval. & Image, Text & \href{https://github.com/miccunifi/CIRCO}{CIRCO} \\
 & MINT-1T \cite{awadalla2024mint} & 1T text tokens and 3.4B images; 10x larger than existing open-source datasets. & Image, Text & \href{https://huggingface.co/datasets/mlfoundations/MINT-1T-HTML}{MINT-1T} \\
 & ShareGPT4V \cite{chen2024sharegpt4v} & 1.2M images with GPT-4-generated captions, including spatial and factual details. & Image, Text & \href{https://sharegpt4v.github.io}{ShareGPT4V} \\
 & OmniCorpus \cite{li2025omnicorpus} & 8.6B images and 1.7T tokens across 2.2B web documents; interleaved text-image layout. & Image, Text & \href{https://github.com/OpenGVLab/OmniCorpus}{OmniCorpus} \\ \midrule
\multirow{20}{*}{\rotatebox[origin=c]{90}{Video-Text}} 
 & BDD-X \cite{bddx} & 77 hours of driving videos with expert textual explanations; for explainable driving behavior. & Video, Text & \href{https://github.com/JinkyuKimUCB/BDD-X-dataset}{BDD-X} \\
 & YouCook2 \cite{zhou2018youcook2} & 2,000 cooking videos with aligned descriptions; focused on video–text tasks. & Video, Text & \href{https://youcook2.eecs.umich.edu/}{YouCook2} \\
 & ActivityNet \cite{caba2015activitynet} & 20,000 videos with multiple captions; used for video understanding and captioning. & Video, Text & \href{http://activity-net.org/}{ActivityNet} \\
 & SoccerNet \cite{giancola2018soccernet} & Videos and metadata for 550 soccer games; includes transcribed commentary and key event annotations. & Video, Text & \href{https://www.soccer-net.org/}{SoccerNet} \\
 & MSVD \cite{chen2011collecting} & 1,970 videos with approximately 40 captions per video. & Video, Text & \href{https://www.cs.utexas.edu/~ml/clamp/videoDescription/}{MSVD} \\
 & LSMDC \cite{rohrbach2015lsmdc} & 118,081 video–text pairs from 202 movies; a movie description dataset. & Video, Text & \href{https://paperswithcode.com/dataset/lsmdc}{LSMDC} \\
 & DiDemo \cite{Hendricks_2017_ICCV} & 10,000 videos with four concatenated captions per video; with temporal localization of events. & Video, Text & \href{https://github.com/LisaAnne/TemporalLanguageRelease}{DiDemo} \\
 & COIN \cite{coin} & 11,827 instructional YouTube videos across 180 tasks; for comprehensive instructional video analysis. & Video, Text & \href{https://coin-dataset.github.io/}{COIN} \\
 & MSRVTT-QA \cite{Xu2017VideoQA} & Video question answering benchmark. & Video, Text & \href{https://github.com/xudejing/video-question-answering}{MSRVTT-QA} \\
 & ActivityNet-QA \cite{yu2019activitynet} & 58,000 human–annotated QA pairs on 5,800 videos; benchmark for video QA models. & Video, Text & \href{https://github.com/MILVLG/activitynet-qa}{ActivityNet-QA} \\
 & EpicKitchens-100 \cite{dima2020rescaling} & 700 videos (100 hours of cooking activities) for online action prediction; egocentric vision dataset. & Video, Text & \href{https://epic-kitchens.github.io/2025}{EPIC-KITCHENS-100} \\
 & Ego4D \cite{grauman2022ego4d} & 4.3M video–text pairs for egocentric videos; massive–scale egocentric video dataset. & Video, Text & \href{https://ego4d-data.org/}{Ego4D} \\
 & HowTo100M \cite{Miech_2019_ICCV} & 136M video clips with captions from 1.2M YouTube videos; for learning text–video embeddings. & Video, Text & \href{https://www.di.ens.fr/willow/research/howto100m/}{HowTo100M} \\
 & CharadesEgo \cite{charadesego} & 68,536 activity instances from ego–exo videos; used for evaluation. & Video, Text & \href{https://prior.allenai.org/projects/charades-ego}{Charades-Ego} \\
 & ActivityNet Captions \cite{Krishna_2017_ICCV} & 20K videos with 3.7 temporally localized sentences per video; dense–captioning events in videos. & Video, Text & \href{https://cs.stanford.edu/people/ranjaykrishna/densevid/}{ActivityNet Captions} \\
 & VATEX \cite{wang2019vatex} & 34,991 videos, each with multiple captions; a multilingual video–and–language dataset. & Video, Text & \href{https://eric-xw.github.io/vatex-website/}{VATEX} \\
 & WebVid \cite{bain2021frozen} & 10M video–text pairs (refined to WebVid-Refined-1M). & Video, Text & \href{https://github.com/m-bain/webvid}{WebVid} \\
 & InternVid \cite{wang2023internvid} & 7M YouTube videos (760K hours), 234M clips, 4.1B words; used for video-text pretraining and representation learning. & Video, Text & \href{https://github.com/OpenGVLab/InternVideo/tree/main/Data/InternVid}{InternVid} \\
 & OpenVid-1M \cite{nan2024openvid} & 1 million video-text pairs for multimodal learning. & Video, Text & \href{https://github.com/NJU-PCALab/OpenVid-1M}{OpenVid-1M} \\
 & Youku-mPLUG \cite{xu2023youkumplug10millionlargescale} & Chinese dataset with 10M video–text pairs (refined to Youku-Refined-1M). & Video, Text & \href{https://github.com/X-PLUG/Youku-mPLUG}{Youku-mPLUG} \\ \midrule
\multirow{8}{*}{\rotatebox[origin=c]{90}{Audio-Text}} 
 & LibriSpeech \cite{panayotov2015librispeech} & 1,000 hours of read English speech with corresponding text; ASR corpus based on audiobooks. & Audio, Text & \href{https://www.openslr.org/12}{LibriSpeech} \\
 & SpeechBrown \cite{abootorabi2024claspcontrastivelanguagespeechpretraining} & 55K paired speech-text samples; 15 categories covering diverse topics from religion to fiction. & Audio, Text & \href{https://huggingface.co/datasets/llm-lab/SpeechBrown}{SpeechBrown} \\
 & AudioCaps \cite{kim-etal-2019-audiocaps} & 46K audio clips paired with human-written text captions. & Audio, Text & \href{https://audiocaps.github.io/}{AudioCaps} \\
 & MusicCaps \cite{agostinelli2023musiclm} & It is composed of 5.5k music-text pairs, with rich text descriptions provided by human experts. & Audio, Text & \href{https://www.kaggle.com/datasets/googleai/musiccaps}{MusicCaps} \\
 & Clotho \cite{drossos2020clotho} & Audio captioning dataset with diverse soundscapes. & Audio, Text & \href{https://zenodo.org/record/3490684}{Clotho} \\
 & WavCaps \cite{mei2024wavcaps} & Large-scale weakly-labeled audio-text dataset, comprising approximately 400k audio clips with paired captions. & Audio, Text & \href{https://github.com/XinhaoMei/WavCaps}{WavCaps} \\
 & Spoken SQuAD \cite{li2018spoken} & Audio version of the SQuAD dataset for spoken question answering, focusing on the listening comprehension task. & Audio, Text & \href{https://github.com/chiahsuan156/Spoken-SQuAD}{Spoken SQuAD} \\
 & AudioSet \cite{gemmeke2017audioset} & 2,084,320 human–labeled 10–second sound clips from YouTube; 632 audio event classes. & Audio, Text & \href{https://research.google.com/audioset/}{AudioSet} \\ \midrule
\multirow{6}{*}{\rotatebox[origin=c]{90}{Medical}} 
 & MIMIC-CXR \cite{johnson2019mimic} & 125,417 training pairs of chest X–rays and reports. & Image, Text & \href{https://physionet.org/content/mimic-cxr/2.0.0/}{MIMIC-CXR} \\
 & CheXpert \cite{irvin2019chexpert} & 224,316 chest radiographs of 65,240 patients; focused on medical analysis. & Image, Text & \href{https://stanfordmlgroup.github.io/competitions/chexpert/}{CheXpert} \\
 & MIMIC-III \cite{johnson2016mimic} & Health-related data from over 40K patients (text data). & Text & \href{https://mimic.physionet.org/}{MIMIC-III} \\
 & IU-Xray \cite{pavlopoulos-etal-2019-survey} & 7,470 pairs of chest X–rays and corresponding diagnostic reports. & Image, Text & \href{https://www.kaggle.com/datasets/raddar/chest-xrays-indiana-university}{IU X-ray} \\
 & PubLayNet \cite{zhong2019publaynetlargestdatasetdocument} & 100,000 training samples and 2,160 test samples built from PubLayNet (tailored for the medical domain). & Image, Text & \href{https://github.com/ibm-aur-nlp/PubLayNet}{PubLayNet} \\
 & Quilt-1M \cite{ikezogwo2023quilt} & 438K medical images with 768K text pairs; includes microscopic images and UMLS entities. & Image, Text & \href{https://quilt1m.github.io/}{Quilt-1M} \\ \midrule
\multirow{5}{*}{\rotatebox[origin=c]{90}{Fashion}} 
 & Fashion-IQ \cite{wu2019fashioniq} & 77,684 images across three categories; evaluated with Recall@10 and Recall@50. & Image, Text & \href{https://github.com/XiaoxiaoGuo/fashion-iq}{Fashion IQ} \\
 & FashionGen \cite{fashiongen} & 260.5K image–text pairs of fashion images and item descriptions. & Image, Text & \href{https://paperswithcode.com/dataset/fashion-gen}{Fashion-Gen} \\
 & VITON-HD \cite{vitonhd} & 83K images for virtual try–on; high–resolution clothing items. & Image, Text & \href{https://github.com/shadow2496/VITON-HD}{VITON-HD} \\
 & Fashionpedia \cite{fashionpedia} & 48,000 fashion images annotated with segmentation masks and fine-grained attributes. & Image, Text & \href{https://paperswithcode.com/dataset/fashionpedia}{Fashionpedia} \\
 & DeepFashion \cite{Liu_2016_CVPR} & Approximately 800K diverse fashion images for pseudo triplet generation. & Image, Text & \href{https://paperswithcode.com/dataset/deepfashion}{DeepFashion} \\ \midrule
\multirow{1}{*}{\rotatebox[origin=c]{90}{3D}} 
 & ShapeNet \cite{chang2015shapenet} & Covering 55 common object categories with ~51,300 unique 3D models. & Text, 3D & \href{https://shapenet.org/}{ShapeNet} \\ \midrule
\multirow{20}{*}{\rotatebox[origin=c]{90}{Knowledge \& QA}} 
 & VQA \cite{antol2015vqa} & 400K QA pairs with images for visual question answering. & Image, Text & \href{https://visualqa.org/}{VQA} \\
 & PAQ \cite{lewis2021paq} & 65M text–based QA pairs; a large–scale dataset. & Text & \href{https://github.com/facebookresearch/PAQ}{PAQ} \\
 & ELI5 \cite{fan2019eli5} & 270K complex and diverse questions augmented with web pages and images. & Text & \href{https://facebookresearch.github.io/ELI5/}{ELI5} \\& MultimodalQA \cite{talmor2021multimodalqa} & 29,918 questions requiring multi-modal multi-hop reasoning over text, tables, and images. & Image, Text, Table & \href{https://allenai.github.io/multimodalqa/}{MultimodalQA} \\
 & ViQuAE \cite{viquae} & 11.8M passages from Wikipedia covering 2,397 unique entities; knowledge–intensive QA. & Text & \href{https://github.com/PaulLerner/ViQuAE}{ViQuAE} \\
 & OK-VQA \cite{marino2019okvqa} & 14K questions requiring external knowledge for VQA. & Image, Text & \href{https://okvqa.allenai.org/}{OK-VQA} \\
 & WebQA \cite{Chang_2022_CVPR} & 46K queries that require reasoning across text and images. & Text, Image & \href{https://webqna.github.io/}{WebQA} \\
 & Infoseek \cite{infoseek} & Fine-grained visual knowledge retrieval using a Wikipedia–based knowledge base (~6M passages). & Image, Text & \href{https://open-vision-language.github.io/infoseek/}{Infoseek} \\
 & ClueWeb22 \cite{clueweb22} & 10 billion web pages organized into three subsets; a large–scale web corpus. & Text & \href{https://lemurproject.org/clueweb22/}{ClueWeb22} \\
 & MOCHEG \cite{yao2023end} & 15,601 claims annotated with truthfulness labels and accompanied by textual and image evidence. & Text, Image & \href{https://github.com/VT-NLP/Mocheg}{MOCHEG} \\
 & VQA v2 \cite{goyal2017making} & 1.1M questions (augmented with VG–QA questions) for fine-tuning VQA models. & Image, Text & \href{https://visualqa.org/}{VQA v2} \\
 & A-OKVQA \cite{schwenk2022aokvqabenchmarkvisualquestion} & Benchmark for visual question answering using world knowledge; around 25K questions. & Image, Text & \href{https://github.com/allenai/aokvqa}{A-OKVQA} \\
 & XL-HeadTags \cite{shohan2024xlheadtagsleveragingmultimodalretrieval} & 415K news headline-article pairs consist of 20 languages across six diverse language families. & Text & \href{https://huggingface.co/datasets/faisaltareque/XL-HeadTags}{XL-HeadTags} \\
 & DocVQA \cite{mathew2021docvqa} & 12,767 diverse document images with 50K QA pairs, categorized by reasoning type to evaluate DocVQA methods. & Image, Text & \href{https://www.docvqa.org/}{DocVQA} \\
 & ChartQA \cite{masry2022chartqa} & 9.6K human-written QA pairs + 23.1K generated from chart summaries. & Image, Text & \href{https://github.com/vis-nlp/ChartQA}{ChartQA} \\
 & DVQA \cite{kafle2018dvqa} & 3.5M QA pairs on 300K diagrams, evaluating structure, data retrieval, and reasoning. & Image, Text & \href{https://github.com/kushalkafle/DVQA_dataset}{DVQA} \\
 & RETVQA \cite{retvqa} & 416,000 QA samples where retrieval from a large image set is needed to answer questions; emphasizes RAG pipeline. & Image, Text & \href{https://vl2g.github.io/projects/retvqa/}{RETVQA} \\
 & SEED-Bench \cite{li2023seedbenchbenchmarkingmultimodalllms} & 19K multiple–choice questions with accurate human annotations across 12 evaluation dimensions. & Text & \href{https://github.com/AILab-CVC/SEED-Bench}{SEED-Bench} \\
 & M3DocVQA \cite{cho2024m3docragmultimodalretrievalneed} & 2,441 multi-hop questions across 3,368 PDF documents; evaluates open-domain DocVQA. & Image, Text & \href{https://m3docrag.github.io/}{M3DocVQA} \\
 & MMLongBench-Doc \cite{ma2024mmlongbenchdocbenchmarkinglongcontextdocument} & 135 lengthy PDFs with 1,091 questions; focuses on multi-hop reasoning in single documents. & Image, Text & \href{https://mayubo2333.github.io/MMLongBench-Doc/}{MMLongBench-Doc} \\ \midrule
\multirow{4}{*}{\rotatebox[origin=c]{90}{Other}} 
 & GeoDE \cite{geode} & 61,940 images from 40 classes across 6 world regions; emphasizes geographic diversity in object recognition. & Image & \href{https://github.com/AliRamazani/GeoDE}{GeoDE} \\
 & RU-AI \cite{huang2025ru} & 1.47M samples of real vs AI-generated content for fake detection robustness. & Image, Text, Audio & \href{https://github.com/ZhihaoZhang97/RU-AI}{RU-AI} \\
 & MIMIC-IT \cite{li2025otter} & 2.8M multimodal instruction-response pairs for model alignment. & Image, Video, Text & \href{https://github.com/Luodian/Otter}{MIMIC-IT} \\
 & MMVQA \cite{ding2024mmvqa} & 262K question-answer pairs across 3,146 multipage research PDFs for robust multimodal information retrieval. & Image, Text & \href{https://github.com/adlnlp/mmvqa}{MMVQA} \\ \midrule
\end{tabular}%
}
\end{table*}

\begin{table*}[t]
\centering
\caption{Overview of Popular Benchmarks in Multimodal RAG Research.}
\label{tab:benchmark_datasets}
\large
\renewcommand{\arraystretch}{3.8}
\resizebox{\textwidth}{!}{%
\begin{tabular}{c c l c l}
\toprule
\rotatebox[origin=c]{90}{\textbf{Category}} & \textbf{Name} & \multicolumn{1}{c}{\textbf{Statistics and Description}} & \textbf{Modalities} & \multicolumn{1}{c}{\textbf{Link}} \\ \midrule
\multirow{6}{*}{\rotatebox[origin=c]{90}{Cross-Modal Understanding}} 
 & MRAG-Bench \cite{hu2024mragbench} & Evaluates visual retrieval, integration, and robustness to irrelevant visual information. & Images & \href{https://mragbench.github.io/}{MRAG-Bench} \\
 & $M^{2}RAG$ \cite{ma2024multimodalretrievalaugmentedmultimodal} & Benchmarks multimodal RAG; evaluates retrieval, multi-hop reasoning, and integration. & Images + Text & \href{https://arxiv.org/abs/2411.16365}{$M^{2}RAG$} \\ 
 & Dyn-VQA \cite{li2024benchmarking} & Focuses on dynamic retrieval, multi-hop reasoning, and robustness to changing information. & Images + Text & \href{https://openreview.net/forum?id=VvDEuyVXkG}{Dyn-VQA} \\ 
 & MMBench \cite{liu2025mmbench} & Covers VQA, captioning, retrieval; evaluates cross-modal understanding across vision, text, and audio. & Images + Text + Audio & \href{https://github.com/open-compass/MMBench}{MMBench} \\ 
 & ScienceQA \cite{lu2022learn} & Contains 21,208 questions; tests scientific reasoning with text, diagrams, and images. & Images + Diagrams + Text & \href{https://scienceqa.github.io/}{ScienceQA} \\ 
 & SK-VQA \cite{su2024sk} & Offers 2 million question-answer pairs; focuses on synthetic knowledge, multimodal reasoning, and external knowledge integration. & Images + Text & \href{https://arxiv.org/abs/2406.19593}{SK-VQA} \\ 
 & SMMQG \cite{wu-etal-2024-synthetic} & Includes 1,024 question-answer pairs; focuses on synthetic multimodal data and controlled question generation. & Images + Text & \href{https://arxiv.org/abs/2407.02233}{SMMQG} \\ \midrule
\multirow{2}{*}{\rotatebox[origin=c]{90}{Text-Focused}} 
 & TriviaQA \cite{joshi2017triviaqa} & Provides 650K question-answer pairs; reading comprehension dataset, adaptable for multimodal RAG. & Text & \href{https://nlp.cs.washington.edu/triviaqa/}{TriviaQA} \\
 & Natural Questions \cite{kwiatkowski2019natural} & Contains 307,373 training examples; real-world search queries, adaptable with visual contexts. & Text & \href{https://paperswithcode.com/dataset/natural-questions}{Natural Questions} \\ \midrule
\end{tabular}%
}
\end{table*}


\begin{figure*}[t]
    \centering
    \resizebox{\textwidth}{!}
    {
        \begin{forest}
            forked edges,
            for tree={
                child anchor=west,
                parent anchor=east,
                grow'=east,
                anchor=west,
                base=left,
                font=\normalsize,
                rectangle,
                draw=hidden-black,
                rounded corners,
                minimum height=2em,
                minimum width=4em,
                edge+={darkgray, line width=1pt},
                s sep=3pt,
                inner xsep=0.4em,
                inner ysep=0.6em,
                line width=0.8pt,
                text width=8.5em,
                ver/.style={
                    rotate=90,
                    child anchor=north,
                    parent anchor=south,
                    anchor=center,
                    text width=17em, 
                    font=\normalsize, 
                    minimum height=3em, 
                    minimum width=5em, 
                },
                leaf/.style={
                    fill=green!20, 
                    text= black,
                    text width=44.5em,
                    font=\normalsize,
                    inner xsep=0.4em,
                    inner ysep=0.6em,
                    draw,
                }, 
                intermediate/.style={
                    font=\normalsize, 
                    minimum height=2.5em, 
                    minimum width=18em, 
                    text width=18em, 
                    draw, 
                }
            },
            [
                Multimodal RAG Application Domains, ver , fill=white!20
                [
                    Healthcare and Medicine~(\S\ref{sec_healthcare})  , fill=yellow!20, intermediate
                    [
                        MMED-RAG~\cite{xia2024mmedragversatilemultimodalrag}{,}
                        RULE~\cite{xia-etal-2024-rule}{,}
                        AsthmaBot~\cite{bahaj2024asthmabot}{,}
                        Realm~\cite{zhu2024realmragdrivenenhancementmultimodal}{,}
                        ~\citet{su2024hybrid}{,}
                        FactMM-RAG~\cite{sun2024factawaremultimodalretrievalaugmentation}{,}
                        RA-RRG~\cite{choi2025leveraging}
                        , leaf
                    ]
                ]
                [
                    Software Engineering~(\S\ref{sec_Software_Engineering})  , fill=yellow!20, intermediate
                    [
                        DocPrompting~\cite{zhou2023docprompting}{,}
                        RACE~\cite{shi-etal-2022-race}{,}
                        CEDAR~\cite{CEDAR}{,}
                        REDCODER~\cite{parvez2021retrieval}
                        , leaf
                    ]
                ]
                [
                    Fashion and E-Commerce~(\S\ref{sec_Fashion})  , fill=yellow!20, intermediate
                    [
                        Unifashion~\cite{zhao-etal-2024-unifashion}{,}
                        ~\citet{dang2024multi}{,}
                        Fashion-RAG \cite{sanguigni2025fashion}{,}
                        LLM4DESIGN~\cite{chen2024llm4design}
                        , leaf
                    ]
                ]
                [
                    Entertainment and Social Computing~(\S\ref{sec_Entertainment})  , fill=yellow!20, intermediate
                    [
                        SoccerRAG~\cite{strand2024soccerragmultimodalsoccerinformation}{,}
                        MMRA~\cite{10.1145/3626772.3657929}
                        , leaf
                    ]
                ]
                [
                    Emerging Applications~(\S\ref{sec_Emerging_Applications})  , fill=yellow!20, intermediate
                    [                        
                        RAG-Driver~\cite{yuan2024ragdrivergeneralisabledrivingexplanations}{,}
                        ENWAR~\cite{nazar2024enwar}{,}
                        ~\citet{riedler2024textoptimizingragmultimodal}{,}
                        Img2Loc~\cite{zhou2024img2loc}
                        , leaf
                    ]
                ]
            ]
        \end{forest}   
    }
    \caption{Taxonomy of application domains for Multimodal Retrieval-Augmented Generation systems.}
    \label{fig:app_taxonomy}
\end{figure*}

\section{Evaluation and Metrics}
\label{sec:evalmetrics}
\noindent

\noindent
Evaluating multimodal RAG models is complex due to their varied input types and complex structure. The evaluation combines metrics from VLMs, generative AI, and retrieval systems to assess capabilities like text/image generation and information retrieval. Our review found about 60 different metrics used in the field. In the following paragraphs, we will examine the most important and widely used metrics for evaluating multimodal RAG.

\noindent
\paragraph{Retrieval Evaluation} Retrieval performance is measured through accuracy, recall, and precision metrics, with an F1 score combining recall and precision.
Accuracy is typically defined as the ratio of correctly predicted instances to the total instances.
In retrieval-based tasks, Top-K Accuracy is defined as:

\begin{align}
\text{Top-K Accuracy}(y, \hat{f}) &= \frac{1}{n} \sum_{i=0}^{n-1} \sum_{j=1}^{k} \mathbb{1}(\hat{f}_{i,j} = y_i)
\end{align}

\noindent
Recall@K, which examines relevant items in top K results, is preferred over standard recall. Mean Reciprocal Rank (MRR) serves as another key metric for evaluation, which is utilized by \cite{omar-etal-2024-multi, nguyen2024multimodallearnedsparseretrieval}.
MRR measures the rank position of the first relevant result in the returned list. The formula for calculating MRR is:

\begin{align}
\text{MRR} = \frac{1}{Q} \sum_{q=1}^{Q} \frac{1}{\text{rank}_q}
\end{align}

\noindent
where $Q$ is the total number of queries.
$rank_q$ is the rank of the first relevant result for query $q$.

\paragraph{Modality Evaluation} Modality-based evaluations primarily focus on text and image, assessing their alignment, text fluency, and image caption quality. For text evaluation, metrics include Exact Match (EM), BLEU \cite{papineni-etal-2002-bleu}, ROUGE \cite{lin-2004-rouge}, and METEOR \cite{banerjee-lavie-2005-meteor}. 
The ROUGE metric is commonly used to evaluate text summarization and generation. 
ROUGE-N measures the overlap of N-grams between the generated and reference text. The formula for ROUGE-N is:

\begin{align}
\text{ROUGE-N} &= \frac{\sum_{\text{gram}_N \in \text{Ref}} \text{Count}_{\text{match}}(\text{gram}_N)}{\sum_{\text{gram}_N \in \text{Ref}} \text{Count}(\text{gram}_N)}
\end{align}

\noindent
ROUGE-L measures the longest common subsequence (LCS) between generated and reference text. The formula for ROUGE-L is:
\begin{align}
\text{ROUGE-L} &= \frac{LCS(X, Y)}{|Y|}
\end{align}

\noindent
BLEU is another metric used for assessing text generation. The formula for calculating BLEU is:

\begin{align}
\text{BLEU}(p_n, \text{BP}) = \text{BP} \cdot \exp \left( \sum_{n=1}^{N} w_n \log p_n \right)
\end{align}

\noindent
Here, $p_n$ represents the precision of n-grams, $w_n$ denotes the weight assigned to the n-gram precision, and the Brevity Penalty (BP) is defined as:

\begin{align}
\text{BP} = 
\begin{cases}
1 & \text{length} > \text{$rl$} \\
\exp\left( 1 - \frac{\text{$rl$}}{\text{$cl$}} \right) & \text{length} \leq \text{$rl$}
\end{cases}
\end{align}

\noindent
Here, $rl$ represents the reference length and $cl$ represents the candidate length.
\\

\noindent
MultiRAGen \cite{shohan2024xlheadtagsleveragingmultimodalretrieval} uses Multilingual ROUGE for multilingual settings. 

\noindent
For image captioning, CIDEr (Consensus-Based Image Description Evaluation) \cite{vedantam2015cider} measures caption quality using TF-IDF and cosine similarity \cite{yasunaga2023retrievalaugmentedmultimodallanguagemodeling, zhao-etal-2024-unifashion, luo2024doestextualinformationaffect, yuan2024ragdrivergeneralisabledrivingexplanations, Sharifymoghaddam2024UniRAGUR, Hu_2023_CVPR, rao2024ravenmultitaskretrievalaugmented,xu2024retrievalaugmentedegocentricvideocaptioning,kim2024you,zhang2024c3net}, while SPICE (Semantic Propositional Image Caption Evaluation) \cite{anderson2016spicesemanticpropositionalimage} focuses on semantics. SPIDEr \cite{liu2017improved}, used in \cite{zhang2024c3net}, combines both metrics.

\noindent
For semantic alignment, BERTScore \cite{zhang2020bertscoreevaluatingtextgeneration} compares BERT embeddings \cite{sun2024factawaremultimodalretrievalaugmentation, shohan2024xlheadtagsleveragingmultimodalretrieval}, and evaluates fluency \cite{chen-etal-2022-murag,10535103,ma2024multimodalretrievalaugmentedmultimodal}. 

\noindent
CLIP Score \cite{hessel-etal-2021-clipscore}, used in \cite{Sharifymoghaddam2024UniRAGUR, zhang2024c3net}, measures image-text similarity using CLIP \cite{radford2021learning}.
The formula for calculating CLIPScore is:

\begin{align}
\text{CLIPScore} = \frac{\mathbf{t}. \mathbf{i}}{\| \mathbf{t} \|  \| \mathbf{i} \|}
\end{align}

\noindent
where t and i are text and image embedding, respectively. \\

\noindent
For image quality, FID (Fréchet Inception Distance) \cite{heusel2017gans} compares feature distributions \cite{yasunaga2023retrievalaugmentedmultimodallanguagemodeling, zhao-etal-2024-unifashion, Sharifymoghaddam2024UniRAGUR, zhang2024c3net}, while KID (Kernel Inception Distance) \cite{bińkowski2018demystifying} provides an unbiased alternative. 
The formula for FID is:
\begin{align}
\text{FID} &= \|\mu_r - \mu_g\|^2 + tr(\Sigma_r + \Sigma_g - 2\sqrt{\Sigma_r \Sigma_g}) 
\end{align}
\noindent

\noindent
where $\mu_r$ and $\Sigma_r$ are the mean and covariance of real images' feature representations, respectively. $\mu_g$ and $\Sigma_g$ are the mean and covariance of generated images' feature representations, respectively. To extract features, InceptionV3 \cite{szegedy2016rethinking} is typically used.

\noindent
Inception Score (IS) evaluates image diversity and quality through classification probabilities \cite{10535103}.
For audio evaluation, \citet{zhang2024c3net} use human annotators to assess sound quality (OVL) and text relevance (REL), while also employing Fréchet Audio Distance (FAD) \cite{kilgour2019frechetaudiodistancemetric}, an audio-specific variant of FID.

\noindent
System efficiency is measured through FLOPs, execution time, response time, and retrieval time per query \cite{nguyen2024multimodallearnedsparseretrieval,strand2024soccerragmultimodalsoccerinformation,dang2024multi, Zhou_2024}. Domain-specific metrics include geodesic distance for geographical accuracy \cite{zhou2024img2loc}, and Clinical Relevance for medical applications \cite{lahiri2024alzheimerragmultimodalretrievalaugmented}.

\section{Robustness Advancements and Loss Functions}
\label{sec:training}
\subsection{Robustness and Noise Management}
\label{sec:robustness}
\noindent
Multimodal training faces challenges such as noise and modality-specific biases \cite{buettner-kovashka-2024-quantifying}. Managing noisy retrieval inputs is critical for maintaining model performance. MORE \cite{cui2024moremultimodalretrievalaugmented} injects irrelevant results during training to enhance focus on relevant inputs. AlzheimerRAG \cite{lahiri2024alzheimerragmultimodalretrievalaugmented} uses progressive knowledge distillation to reduce noise while maintaining multimodal alignment. RAGTrans \cite{10.1145/3637528.3672041} leverages hypergraph-based knowledge aggregation to refine multimodal representations, ensuring more effective propagation of relevant information. RA-BLIP \cite{ding2024rablipmultimodaladaptiveretrievalaugmented} introduces the Adaptive Selection Knowledge Generation (ASKG) strategy, which leverages the implicit capabilities of LLMs to filter relevant knowledge for generation through a denoising-enhanced loss term, eliminating the need for fine-tuning. This approach achieves strong performance compared to baselines while significantly reducing computational overhead by minimizing trainable parameters. RagVL \cite{chen2024mllm} improves robustness through noise-injected training by adding hard negative samples at the data level and applying Gaussian noise with loss reweighting at the token level, enhancing the model’s resilience to multimodal noise. 
Finally, RA-CM3 \cite{yasunaga2023retrievalaugmentedmultimodallanguagemodeling} enhances generalization using Query Dropout, which randomly removes query tokens during retrieval, serving as a regularization method that improves generator performance.

\subsection{Loss Function}
\noindent
\textbf{InfoNCE (Information Noise Contrastive Estimation)}: The InfoNCE loss is commonly used in self-supervised learning, especially in contrastive learning methods. The formula for InfoNCE loss is:
\begin{align}
\mathcal{L}_{\text{InfoNCE}} = -\log \frac{\exp(\text{sim}(z_i, z_j)/\tau)}{\sum_{k=1}^{K} \exp(\text{sim}(z_i, z_k)/\tau)}
\end{align}
where $z_i$ and $z_j$ are the embeddings of a positive pair and $\tau$ is a temperature parameter. \\

\noindent
\textbf{GAN (Generative Adversarial Network)}: The GAN loss consists of two parts: the discriminator loss and the generator loss.
The discriminator loss formula is:
\begin{align}
\scriptstyle \mathcal{L}_D = - \mathbb{E}_{x \sim p_{\text{data}}(x)}[\log D(x)] - \mathbb{E}_{z \sim p_z(z)}[\log(1 - D(G(z)))]
\end{align}
where $x$ is a real sample from the data distribution.
$G(z)$ is the generated sample from the generator, where 
$z$ is a noise vector.
$D(x)$ is the probability that $x$ is real.

\noindent
The Generator loss formula is:
\begin{align}
\mathcal{L}_G = \mathbb{E}_{z \sim p_z(z)}[\log(1 - D(G(z)))]
\end{align}

\noindent
\textbf{Triplet Loss}: Triplet Loss is used in metric learning to ensure that similar data points are closer together while dissimilar ones are farther apart in the embedding space. The Triplet loss formula is:

\begin{align}
\scriptstyle \mathcal{L} = \sum_{i=1}^{N} \max(0, \| f(x_a^i) - f(x_p^i) \|^2 - \| f(x_a^i) - f(x_n^i) \|^2 + \alpha)
\end{align}

\noindent
where $x_a^i$ is the anchor sample. $x_p^i$ and $x_n^i$ are the positive and negative samples, respectively.
$f(x)$ is the neural network.

\section{Applications and Relevant Tasks}
\label{sec_app}
\noindent
Multimodal RAG extends traditional RAG beyond unimodal settings to cross-modal tasks. In content generation, it enhances image captioning \cite{10535103, Hu_2023_CVPR, rao2024ravenmultitaskretrievalaugmented} and text-to-image synthesis \cite{yasunaga2023retrievalaugmentedmultimodallanguagemodeling, chen2022reimagenretrievalaugmentedtexttoimagegenerator} by retrieving relevant contextual information. It also improves coherence in visual storytelling and factual alignment in multimodal summarization \cite{tonmoy2024comprehensive}. In knowledge-intensive applications, multimodal RAG supports open-domain QA \cite{chen2024mllm, ding2024rablipmultimodaladaptiveretrievalaugmented, yuan2023rammretrievalaugmentedbiomedicalvisual}, video-based QA \cite{luo2024videoragvisuallyalignedretrievalaugmentedlong}, fact verification \cite{khaliq-etal-2024-ragar}, and zero-shot image–text retrieval \cite{10.1145/3626772.3657740}, grounding responses in retrieved knowledge and thereby mitigating hallucinations. 

\noindent
Additionally, the incorporation of chain-of-thought reasoning \cite{zhai2024selfadaptivemultimodalretrievalaugmentedgeneration, khaliq-etal-2024-ragar} further enhances complex problem-solving and inference. Finally, their integration into AI assistants such as Gemini \cite{geminiteam2024geminifamilyhighlycapable} enables natural language-driven visual search, document understanding, and multimodal reasoning.

\noindent
Multimodal RAGs are increasingly applied across diverse domains, including healthcare, software engineering, and creative industries (e.g., fashion and design automation). The taxonomy of application domains can be seen in ~\autoref{fig:app_taxonomy}.
The following sections explore domain-specific adaptations of these techniques in greater depth.

\noindent
\paragraph{Healthcare and Medicine} \label{sec_healthcare}
Multimodal RAG enhances clinical decision-making through integrated analysis of medical imaging, electronic health records, and biomedical literature. Systems like MMED-RAG \cite{xia2024mmedragversatilemultimodalrag} address diagnostic uncertainty in medical visual question answering by aligning radiology images with contextual patient data. 
In automated report generation, RULE \cite{xia-etal-2024-rule} mitigates hallucinations through dynamic retrieval of clinically similar cases. Similarly, RA-RRG \cite{choi2025leveraging} first leverages an LLM to extract key textual phrases from a report corpus, then employs a multimodal retriever to link the visual features to these relevant phrases. The coherent report is generated after being retrieved by another LLM without fine-tuning, thereby reducing hallucinations.
FactMM-RAG \cite{sun2024factawaremultimodalretrievalaugmentation} further automates radiology report drafting by retrieving biomarker correlations from medical ontologies, exemplifying the paradigm’s capacity to operationalize expert knowledge at scale.
AsthmaBot \cite{bahaj2024asthmabot} introduces a multimodal RAG-based approach for supporting asthma patients across multiple languages, enabling structured, language-specific semantic searches. Predictive frameworks such as Realm \cite{zhu2024realmragdrivenenhancementmultimodal} demonstrate robust risk assessment by fusing heterogeneous patient data streams, while \citet{su2024hybrid} advance privacy-preserving architectures for federated clinical data integration.

\noindent
\paragraph{Software Engineering}\label{sec_Software_Engineering}
Code generation systems leverage multimodal RAG to synthesize context-aware solutions from technical documentation and version histories. DocPrompting \cite{zhou2023docprompting} improves semantic coherence in code completion by retrieving API specifications and debugging patterns. Commit message generation models like RACE \cite{shi-etal-2022-race} contextualize code diffs against historical repository activity, while CEDAR \cite{CEDAR} optimizes few-shot learning through retrieval-based prompt engineering. REDCODER \cite{parvez2021retrieval} enhances code summarization via semantic search across open-source repositories, preserving syntactic conventions across programming paradigms.

\noindent
\paragraph{Fashion and E-Commerce} \label{sec_Fashion}
Cross-modal alignment drives advancements in product discovery and design automation. UniFashion \cite{zhao-etal-2024-unifashion} enables style-aware retrieval by jointly embedding garment images and textual descriptors, while \citet{dang2024multi} reduces search friction through multimodal query expansion. 
For fashion image editing, Fashion-RAG \cite{sanguigni2025fashion} employs a retrieval-augmented approach, retrieving garments that match textual descriptions and integrating their attributes into image generation via textual inversion techniques within diffusion models, ensuring personalized and contextually relevant outputs.
LLM4DESIGN \cite{chen2024llm4design} demonstrates architectural design automation by retrieving compliance constraints and environmental impact assessments, underscoring RAG’s adaptability to creative domains.

\noindent
\paragraph{Entertainment and Social Computing} \label{sec_Entertainment} Multimedia analytics benefit from RAG’s capacity to correlate heterogeneous signals. SoccerRAG \cite{strand2024soccerragmultimodalsoccerinformation} derives tactical insights by linking match footage with player statistics. MMRA \cite{10.1145/3626772.3657929} predicts content virality through joint modeling of visual aesthetics and linguistic engagement patterns.

\noindent
\paragraph{Emerging Applications}\label{sec_Emerging_Applications} Autonomous systems adopt multimodal RAG for explainable decision-making, as seen in RAG-Driver’s \cite{yuan2024ragdrivergeneralisabledrivingexplanations} real-time retrieval of traffic scenarios during navigation. ENWAR \cite{nazar2024enwar} enhances wireless network resilience through multi-sensor fusion, while \citet{riedler2024textoptimizingragmultimodal} streamline equipment maintenance by retrieving schematics during fault diagnosis. Geospatial systems such as Img2Loc \cite{zhou2024img2loc} advance image geolocalization through cross-modal landmark correlation.

\section{Additional Future Directions}
\label{sec:future}
\noindent
High computational costs in video frame sampling and memory bottlenecks in processing multi-page documents with images remain key challenges in long-context processing. Fixed extraction rates struggle to capture relevant frames, requiring adaptive selection based on content complexity and movement \cite{kandhare2024empiricalcomparisonvideoframe}. Additionally, retrieval speed-accuracy trade-offs in edge deployments and redundant computations in cross-modal fusion layers emphasize the need for efficient, scalable architectures. Personalization mechanisms, like user-specific retrieval (e.g., adapting to medical history), remain underexplored. As these systems evolve, ensuring privacy and preventing sensitive data leakage in multimodal outputs is critical. Lastly, the lack of datasets with complex reasoning tasks and multimodal adversarial examples limits robust evaluation.
\end{document}